\documentclass[doctor]{thesis} 

\usepackage[cmex10]{amsmath}
\usepackage{amsthm,amssymb}
\usepackage{tikz}
\usepackage{pgfplots}
\usepackage{cjhebrew}

\DeclareGraphicsExtensions{.pdf,.png,.jpg}

\usepackage[caption=false]{subfig}

\usepackage{booktabs}

\usepackage{url}
\usepackage{cite} 

\usepackage{tabularx}
\usepackage[ruled,vlined]{algorithm2e}

\newcommand{\cref}[1]{Chapter~\ref{#1}}  

\theoremstyle{definition}

\title{Neural Subnetwork Ensembles}

\author{Timothy J. Whitaker}

\email{timothy.whitaker@colostate.edu}

\department{Department of Computer Science}

\semester{Fall 2023}

\advisor{Darrell Whitley}
\committee{Charles Anderson}
\committee{Nikhil Krishnaswamy}
\committee{Michael Kirby}

\mycopyright{%
Copyright by Timothy J. Whitaker 2023 \\
All Rights Reserved
}

\abstract{%
Neural network ensembles have been effectively used to improve generalization by combining the predictions of multiple independently trained models. However, the growing scale and complexity of deep neural networks have led to these methods becoming prohibitively expensive and time consuming to implement. Low-cost ensemble methods have become increasingly important as they can alleviate the need to train multiple models from scratch while retaining the generalization benefits that traditional ensemble learning methods afford. This dissertation introduces and formalizes a low-cost framework for constructing Subnetwork Ensembles, where a collection of child networks are formed by sampling, perturbing, and optimizing subnetworks from a trained parent model. We explore several distinct methodologies for generating child networks and we evaluate their efficacy through a variety of ablation studies and established benchmarks. Our findings reveal that this approach can greatly improve training efficiency, parametric utilization, and generalization performance while minimizing computational cost. Subnetwork Ensembles offer a compelling framework for exploring how we can build better systems by leveraging the unrealized potential of deep neural networks.
}

\acknowledgements{%
I would like to start by thanking my advisor Dr.\ Darrell Whitley for his mentorship and guidance. It's been an honor and a privilege to work together for the last several years. Thank you for taking a chance on me and for pushing me to become a better research scientist. Thank you to the members of my dissertation committee: Dr.\ Chuck Anderson, Dr.\ Nikhil Krishnaswamy, and Dr.\ Michael Kirby. I am so grateful to have been able to learn from each of you. Your classes have been an integral part of this research and I appreciate the feedback, advice, and guidance that you've offered along the way. 
I am appreciative of financial support from The Computer Science Department at Colorado State University (Artificial Intelligence and Evolutionary Computation Fellowship) and The National Science Foundation (Grant IIS-1908866).

I am thankful for my family and friends who have stood by me throughout this journey. A special thanks to my mom and dad, Sandy and Brian, for their unwavering support and unconditional love. I am eternally grateful for the sacrifices you have made to give me the opportunity to be where I am today. You've set an example for what one can accomplish with hard work, determination and the courage to pursue your dreams. 
Shout out to my brother, Dylan, and my best friend, Kyler, for always having my back. Thank you for giving me perspective and focus when I needed it most.
Thank you to my extended family and in-laws for their encouragement and continued interest in my work. It has meant a lot to know I have so many loved ones rooting for me.

Last but not least, my deepest gratitude is reserved for my wife, Alyssa. I feel so blessed to have you in my life. You were the inspiration for taking on this endeavor and you were the foundation upon which it was all built. You've been there to celebrate my triumphs and you've been there to help me overcome insurmountable challenges. You've been my constant source of strength and my guiding light when all seemed dark. Thank you for always believing in me. I could not have done this without you.
}

\begin{document} 

\frontmatter 

\maketitle              
\makemycopyright        
\makeabstract           
\makeacknowledgements   

\prelimtocentry{Dedication} 
\begin{flatcenter} 

    DEDICATION

    \vfill 

    \noindent \textit{For Alyssa\\
    In this grand cosmic ensemble, it is my deepest joy to share this time and space with you.\\}
    \vfill 
\end{flatcenter}
\newpage

\tableofcontents    
\listoftables       
\listoffigures      

\mainmatter 

\chapter{Introduction}
\label{chap:introduction}
\section{Overview}

Ensemble Learning has long been an active area of research in machine learning  as it is one of the most reliable, powerful, and flexible techniques for improving the predictive power of machine learning systems.
\cite{hansen1990neural, krogh1994neural}.
Rather than training a single model to make predictions, ensemble methods instead train several models and use them in combination with one another in order to make predictions.
Combining the outputs of several models allows for ensembles to improve generalization by leveraging the complementary strengths and weaknesses of its members \cite{dietterich2000ensemble}.
The principles behind ensemble learning algorithms are general and have been used with many types of parametric models (e.g. decision trees, support vector machines, neural networks) \cite{breiman1996bagging, drucker1995boosting, breiman2001random}.
Recent research has shown that ensembles of deep neural networks in particular are especially effective, often being implemented as a part of winning solutions in high profile machine learning competitions \cite{melis2014higgs, chesler2023vesuvius, zhang2021janestreet, singer2019earthquake, ito2019icecube, silver2016mastering, ferrucci2010watson}.

Improvements in network architecture and training methodologies have led to a large increase in both model size and predictive power.
Networks with billions of parameters are becoming increasingly commonplace as overparameterization is shown to be effective for improved training dynamics.
However, the resource intensive nature of these large models brings about important challenges related to computational requirements, memory constraints, and energy efficiency.
As deep neural networks continue to grow in both scale and complexity, the cost associated with traditional ensemble learning techniques quickly becomes untenable.
The development of low-cost ensemble methods have become increasingly important as they offer the opportunity to significantly reduce computational costs while retaining the generalization benefits that full size ensembles afford.

Meanwhile, pruning has emerged as a critical avenue of research for reducing the size of deep neural networks.
Large numbers of parameters can be removed from previously trained models with little discernible loss in predictive power after a few epochs of continued training \cite{blalock2020state, frankle2019lottery}.
Saliency metrics are often used to select the most unnecessary parameters to remove in order to produce the most accurate subnetwork \cite{lecun1990optimal,tanaka2020pruning,frankle2019lottery}.
However, there is no single magic subnetwork that works best for every situation \cite{wolpert1997nofreelunch, chen2023no}.
A sufficiently overparameterized network contains vast numbers of equally capable subnetworks and even random sampling has been shown to result in highly accurate subnetworks at moderate levels of sparsity \cite{diffenderfer2021multiprize, liu2022unreasonable}.

In light of these observations, we introduce a novel low-cost ensemble learning framework that involves the dynamic generation of a collection of child networks spawned through the random sampling of subnetworks in a trained parent model.
This research introduces and formalizes this process of constructing \textit{Subnetwork Ensembles} with the goal of exploring how we can improve the training efficiency, parametric utilization, and generalization of deep ensembles while minimizing computational cost.

\section{Challenges}

The benefits of improved generalization have been well established in ensemble learning literature as performance has been shown to increase with larger numbers of both accurate and diverse members \cite{dietterich2000ensemble}.
Diversity is an important concept that pertains to the correlation of predictions made by individual models in an ensemble \cite{brown2005diversity, ueda1996generalization, kuncheva2003measures}.
If each model in the ensemble makes the same predictions or learns the same feature representations, then there is little benefit to be gained by combining their outputs.

Traditional ensemble methods encouraged diversity by training large collections of fast and simple classifiers on different subsets of the feature space \cite{breiman1996bagging, drucker1995boosting, ho1998random}.
However, these dataset splitting techniques do not work well with deep neural networks due to their data hungry nature.
Deep neural network ensembles are more commonly trained on the entire dataset, instead relying on random weight initialization, heterogeneous network architectures and stochastic ordering of training samples to encourage diversity.

Diversity is an especially important consideration for low-cost ensemble learning methods as they typically alleviate the need to train multiple deep neural networks from scratch by sharing network structure, gradient information, or learned parameters among ensemble members.
This sharing of information invariably increases correlation and reduces diversity, which can have a significant negative impact on generalization \cite{brown2005diversity, kuncheva2003measures, ueda1996generalization}. 

Low-Cost ensemble methods often struggle to balance this trade-off between member diversity, member accuracy and ensemble size.
Prioritizing one aspect often comes at the expense of another when working with a fixed compute budget.
Low-cost methods can also lack algorithmic flexibility as they typically require particular training procedures or specialized network architectures to construct ensemble members \cite{huang2017snapshot, garipov2018loss, lee2015m, wen2020batchensemble}.
This can make it challenging to adapt these ensembles to accommodate evolving problem spaces.

Subnetwork Ensembles alleviate many of the challenges that typically plague competing methods.
Our approach builds on the idea that we can train a single parent network and spawn diverse child networks by sampling, perturbing, and optimizing unique subnetworks from the parent.
The decoupled nature between the parent and child networks allow for highly flexible implementations.
The parent network can use any standard training procedure or network architecture without modification.
This offers an opportunity to leverage pre-trained models which is becoming more valuable as networks continue to scale in size and cost.
This also enables the ability to generate new child networks dynamically, which can lead to larger and more robust ensembles than are typically seen with deep neural networks.
Child networks can be generated and optimized in parallel which allows for wide scaling as independent members don't rely on the weights of other child networks in the ensemble.
Using sparse subnetworks allows for the potential to greatly reduce computational complexity while maintaining accuracy. 
The investigation of Subnetwork Ensembles could prove to be an important avenue of research for not only improving the efficiency and generalization of low-cost ensemble methods, but also for advancing the theoretical underpinnings of neural computation.



\section{Contributions}

This work introduces, generalizes and formalizes the process of constructing, training, and evaluating Subnetwork Ensembles. We explore several approaches for generating child networks through noise perturbations, network pruning, and stochastic masking.
We conduct a thorough and in depth evaluation of these methods across a number of benchmark tasks and with a wide variety of deep neural network architectures.

Chapter 2 introduces relevant background information and related work grounded in the foundations of modern deep learning. We explore how this work relates to the natural structures observed in biological neural networks. We discuss the state of deep neural network architectures and optimization methods. We also discuss traditional ensemble learning techniques and conduct a literature review of recent low-cost deep neural network ensemble methods.

Chapter 3 introduces the theoretical framework and formalizes Subnetwork Ensembling at a high level. We provide a mathematical overview of this approach and discuss various hyperparameters that govern how Subnetwork Ensembles could be formulated through different sampling, perturbation, and optimization methodologies. We introduce Neural Partitioning as a novel technique for encouraging diversity by dividing the parent network into independent partitions such that sets of child networks are formed with no inherited parameter overlap.

Chapter 4 introduces Noisy Subnetwork Ensembles, where child networks are generated by applying noise perturbations to subnetworks. 
We connect the ideas behind this approach to popular neuroevolutionary methods that use dense mutations and model selection strategies to generate populations of child networks.
We conduct several ablation studies exploring the impact of noisy subnetwork mutations on network behavior.
We introduce a grid search algorithm for finding good perturbation hyperparameters by using trust region constraints, which limits divergence in the output space as a result of perturbation \cite{schulman2017trust}. 
We evaluate Noisy Subnetwork Ensembles with many different deep neural network architectures on the difficult ImageNet dataset and we explore how this technique can be used on difficult reinforcement learning tasks with the ProcGen benchmark \cite{deng2009imagenet, cobbe2020leveraging}.

Chapter 5 introduces Sparse Subnetwork Ensembles, where we generate child networks by pruning randomly sampled subnetworks from the trained parent network. 
We explore the efficacy of Neural Partitioning and demonstrate how it can reduce variance and improve performance by creating sets of child networks with opposing topologies. 
We illustrate how sparse child networks can be tuned to encourage diversity by using a phasic learning rate schedule that uses large learning rates to move child networks further apart in weight space before converging to local optima with small learning rates \cite{smith2018superconvergence, fastai2017onecycle}.
We conduct several other ablations exploring the impact of pruning structure, pruning severity, ensemble size, and tuning procedure.
We benchmark Sparse Subnetwork Ensembles against several state of the art low-cost and traditional ensemble methods on both in-distribution and out-of-distribution datasets, where we find that our approach proves to be more accurate and more robust than traditional ensemble methods that require up to 4x more training compute, all while using fewer parameters.

Chapter 6 introduces Stochastic Subnetwork Ensembles, which extends work in the previous chapter with a novel approach to representing subnetworks with probabilistic masks. 
Rather than pruning all the parameters in one-shot, instead each parameter is assigned a probability score that determines whether the connection will be retained on any forward pass. 
These probabilities are then annealed over several epochs such that the target subnetwork slowly reveals itself over time. 
Diversity is encouraged by assigning different probability distributions to different child networks. 
The inclusion of extra dimensionality in the early phases of tuning can help to provide smooth optimization of the target subnetwork. 
This proves to be highly effective technique for extremely sparse (98\%) and moderately sparse (50\%) subnetworks, outperforming established one-shot and iterative pruning strategies.

Chapter 7 includes an in depth analysis of diversity in Sparse Subnetwork Ensembles.
We include a standard evaluation of pairwise output diversity metrics and compare those to other benchmark ensemble algorithms.
We additionally introduce a new approach to analyzing diversity by leveraging interpretability and visualization methods. 
We discuss how feature visualization and saliency maps can be used to qualitatively compare the differences between knowledge representations among sibling child networks in an ensemble. 
This allows us to meaningfully explore how and why diversity develops and contributes to robust performance. 
We introduce several perceptual hashing algorithms used to measure the distances between feature visualizations, improving on the typically noisy results seen with common image difference metrics like the squared error or cosine similarity.
We conduct a large scale evaluation of this interpretable approach with both Sparse Subnetwork Ensembles and a benchmark temporal ensemble method called Snapshot Ensembles. 
We find that Sparse Subnetwork Ensembles produce significantly more diversity in both the feature space and output space among child networks while maintaining better accuracy.

Chapter 8 wraps up with a summary of the ideas, experiments, and observations found through this research.
We discuss of how the Subnetwork Ensemble framework fits into the current landscape of ensemble methods for deep neural networks.
We include our thoughts on the broader impacts and potential limitations that this work may have in other domains, and we explore several promising avenues and ideas for furthering this research in the future.


\chapter{Background}
\label{chap:background}
\section{Neuroscience}

Biological neural networks are far larger and more complex than the artificial neural networks used in machine learning research.
Even the largest artificial neural networks pale in comparison to the human brain, which is a vast and complex web of interconnected subsystems containing over 86 billion neurons and 100 trillion synaptic connections \cite{drachman2005we}.
Individual neurons within this framework tend to be very noisy, which makes it difficult to understand the relation between neuronal activity and behavior.
A new conceptual paradigm is reframing how we view neuronal activity by instead exploring how groups of structurally connected neurons form the functional building blocks of the brain \cite{carrillo2020neuronal}.
The organizational structure of these interconnected subsystems appears to play an important role for both energy efficiency and the emergence of complex, resilient, and adaptable learning behavior.

These structures emerge from an early process in brain development that involves a massive proliferation of neurons and synaptic connections followed by a widespread pruning of billions and trillions of those neurons and connections respectively.
Neuronal proliferation creates an environment in which neurons and connections compete for resources. 
Neuronal apoptosis results in widespread neural cell death, resulting in a highly specialized structure consisting of many sparsely connected neuronal groups.
These neuronal groups form the primary repertoire from which the rest of brain development occurs \cite{edelman1987neuraldarwinism}.
This overgrowth and subsequent pruning maximizes memory performance under metabolic energy constraints \cite{chechik1998synaptic}.
Additionally, this connective structure allows for information to be encoded in a sparse and distributed manner which optimizes a tradeoff between energy expenditure and expressivity \cite{attwell2001energy, glorot2011relu}.

Several other important processes rely on the flexibility and resilience of these sparse subnetwork structures to encourage healthy brain function \cite{sporns2016networks}.
Neuroplasticity serves as a fundamental mechanism for learning and memory that relies on reorganizing synaptic connections in response to stimulus \cite{costandi2016neuroplasticity}.
Long-term depression and long-term potentiation facilitate this process by strengthening or weakening the synaptic connections between neurons \cite{ito1989long, teyler1987long}.
Sparse subnetwork structures also provide a protective framework for mitigating the consequences of cellular damage or neurodegeneration.
Synaptic Stripping is one such process where immunocompetent cells called microglia constantly scan the brain and selectively remove dysfunctional synapses from injured neurons \cite{kettenmann2013microglia, vsivskova2013microglia}.
This is critical for preventing cascading failures that could result from synaptic dysfunction.

\section{Artificial Neural Networks}

The origins of artificial neural network research can be traced back to the work of Warren McCulloch and Walter Pitts in 1943, who proposed a mathematical model of a biological neuron based on the all-or-nothing character of nervous activity \cite{mcculloch1943logical}.
This simplified computational model casts the neuron as a binary threshold unit that computes a weighted sum of its inputs to produce an output.
Frank Rosenblatt later expanded on this work by developing the first single layer artificial neural network, consisting of McCulloch and Pitts neurons, called the Perceptron \cite{rosenblatt1958perceptron}.
While the Perceptron was limited to modeling simple linearly separable functions, it was groundbreaking work that introduced adaptability to neural models through learning algorithms \cite{marvin1969perceptrons}.

The transition to multilayer perceptrons greatly expanded the learning capabilities of neural networks \cite{hornik1989multilayer}.
Networks would now be organized into several successive layers, each consisting of multiple neurons with weighted connections to neurons in adjacent layers.
The output of a neuron is computed by taking the weighted sum of all the inputs into the neuron and passing it through some non-linear activation function. Table 2.1 includes a small sample of some of the most popular activation functions used in neural networks over time.

\tikzset{basic/.style={draw,fill=none,
                       text badly centered,minimum width=3em}}
\tikzset{input/.style={basic,circle,minimum width=4em}}
\tikzset{weights/.style={basic,rectangle,minimum width=2em}}
\tikzset{functions/.style={basic,circle, minimum width=4em}}
\newcommand{\addaxes}{\draw[help lines, dashed] (0em,1em) -- (0em,-1em)
                            (-1em,0em) -- (1em,0em);}
\newcommand{\relu}{\draw[line width=1.5pt] (-1em,0) -- (0,0)
                                (0,0) -- (0.75em,0.75em);}
\newcommand{\stepfunc}{\draw[line width=1.5pt] (0.65em,0.65em) -- (0,0.65em) 
                                    -- (0,-0.65em) -- (-0.65em,-0.65em);}

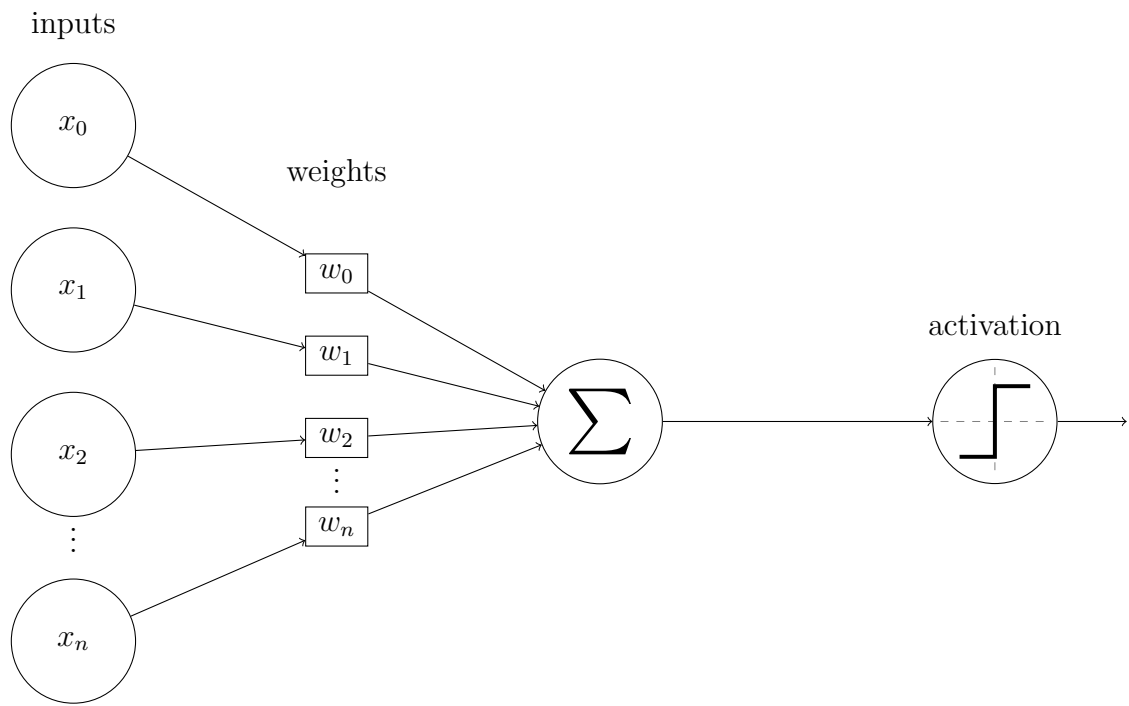
\begin{figure}
\begin{tikzpicture}[scale=1.75]
    \foreach \h [count=\hi ] in {$x_2$,$x_1$,$x_0$}{%
          \node[input] (f\hi) at (0,\hi*1.25cm-1.5cm) {\h};
        }
    \node[below=0.62cm] (idots) at (f1) {\vdots};
    \node[input, below=0.62cm] (last_input) at (idots) {$x_n$};
    \node[functions] (sum) at (4,0) {\huge$\sum$};
    \foreach \h [count=\hi ] in {$w_2$,$w_1$,$w_0$}{%
          \path (f\hi) -- node[weights] (w\hi) {\h} (sum);
          \draw[->] (f\hi) -- (w\hi);
          \draw[->] (w\hi) -- (sum);
        }
    \node[below=0.05cm] (wdots) at (w1) {\vdots};
    \node[weights, below=0.45cm] (last_weight) at (wdots) {$w_n$};
    \path[draw,->] (last_input) -- (last_weight);
    \path[draw,->] (last_weight) -- (sum);
    \node[functions] (activation) at (7,0) {};
    \begin{scope}[xshift=7cm,scale=1]
        \addaxes
        \stepfunc
    \end{scope}
    \draw[->] (sum) -- (activation);
    \draw[->] (activation) -- ++(1,0);
    \node[above=1cm]  at (f3) {inputs};
    \node[above=1cm] at (w3) {weights};
    \node[above=1cm] at (activation) {activation};
    \end{tikzpicture}
    \caption{A diagram of the McCulloch and Pitts computational model of a neuron used in the Perceptron. The neuron computes a weighted sum of the inputs and then passes the output through a binary step function for classification.}
\end{figure}

\begin{table}
\vspace{0.2in}
\caption{A collection of popular activation functions used in neural networks. Logistic Sigmoids and Hyperbolic Tangents have long been the standard for shallow multilayer perceptrons thanks to their smooth differentiable qualities. Deep neural networks have primarily moved towards rectifiers as their unbounded nature helps preserve gradient flow over many layers.}
\begin{tabularx}{\textwidth}{X >{\centering}X >{\centering}X c}
\toprule
Name & Function ($\sigma(x)$) & Derivative ($\sigma'(x)$) & Graph \\
\midrule
Logistic & $\displaystyle \frac{1}{1 + e^{-x}}$ & $\sigma(x)(1 - \sigma(x))$ &     \begin{tikzpicture}[baseline={(0,0.2)}]
     \draw[help lines, dashed] (-1,0) -- (1,0);
     \draw[help lines, dashed] (0,0) -- (0,1);
     \draw plot[domain=-1:1,variable=\x] ({\x},{1/(1+exp(-4*\x))});
    \end{tikzpicture} \\
\\
Hyperbolic Tangent & $\displaystyle \frac{e^x - e^{-x}}{e^x + e^{-x}}$ & $1 - \sigma(x)^2$ & 
\begin{tikzpicture}[baseline={(0,0)}]
     \draw[help lines, dashed] (-1,0) -- (1,0);
     \draw[help lines, dashed] (0,-0.6) -- (0,0.6);
     \draw plot[domain=-1:1,variable=\x] ({\x},{0.4* tanh(4*\x)});
    \end{tikzpicture} \\
\\
Binary Step &$\begin{cases}
    0 & ~\text{if}~ x<0 \\ 
    1 & ~\text{if}~x \geq 0
    \end{cases}$ & 0 &     
    \begin{tikzpicture}[baseline={(0,0.2)}]
     \draw[help lines, dashed] (-1,0) -- (1,0);
     \draw[help lines, dashed] (0,0) -- (0,1);
     \draw plot[domain=-1:0,variable=\x] ({\x},{0});
     \draw plot[domain=0:1,variable=\x] ({\x},{0.7});
\end{tikzpicture} \\
\\
    Rectified Linear Unit & $\begin{cases}
    0 & ~\text{if}~ x<0 \\ 
    x & ~\text{if}~x \geq 0
    \end{cases}$ & $\begin{cases}
    0 & ~\text{if}~ x<0 \\ 
    1 & ~\text{if}~x \geq 0
    \end{cases} $ & 
    \begin{tikzpicture}[baseline={(0,0.5)}]
     \draw[help lines, dashed] (-1,0) -- (1,0);
     \draw[help lines, dashed] (0,0) -- (0,1);
     \draw plot[domain=-1:1,variable=\x] ({\x},{ifthenelse(\x<0,0,\x)});
    \end{tikzpicture} \\[0.2in]
\bottomrule
\end{tabularx}
\end{table}

\subsection{Optimization}

Learning happens by adjusting the weights of the neural network to minimize error over a set of training samples. 
The error is described with a loss function that serves to measure the differences between model predictions and the true target values.
The loss function can vary according to the task and training procedure.
The mean squared error (MSE) and the mean absolute error (MAE) are both popular measures for regression tasks, where the goal is to predict a single scalar value.
\begin{align}
MSE &= \frac{1}{N} \sum_{i=1}^N (Y - \hat{Y})^2 \\
MAE &= \frac{1}{N} \sum_{i=1}^N |Y - \hat{Y}|
\end{align}

Different loss functions may result in different optimization behavior according to their mathematical properties. 
For example, the mean squared error will tend to magnify the penalty for outliers compared to the mean absolute error.
Regularization and penalty terms, such as weight decay, are often included in these loss functions to encourage optimization with respect to additional constraints \cite{zhang2018mechanisms}.

The cross entropy loss (CE) is commonly used for classification tasks.
Consider a dataset with samples that belong to one of $C$ potential classes.
The target label for each sample is encoded into a $C$-dimensional one-hot vector where each element in the vector has a value of 0 except for the correct class which has a value of 1.
Neural networks are configured to output a $C$-dimensional vector where each dimension corresponds to the likelihood of that particular class being predicted.
The outputs of the network are normalized and converted into a probability distribution by using the softmax function.
The error is then determined by measuring the log likelihood between the predicted probability distribution $\hat{Y}$ and the true distribution $Y$.
\begin{equation}
CE = - \sum_{i=1}^C Y_i \ log(\hat{Y}_i)
\end{equation}

Optimization is guided by algorithms that iteratively adjust the weights throughout training.
Supervised deep learning tasks overwhelmingly use variations of gradient descent \cite{ruder2016overview}.
Data is fed through the network, the error is calculated according to the loss function, and updates are made by adjusting the weights according to a learning rate $\eta$ with respect to the gradient of the loss $\nabla F(x)$. 
\begin{equation}
w_{t+1} = w_t - \eta \nabla F(x)
\end{equation}

The advent of the backpropagation algorithm provided an efficient method for calculating the gradient of neural networks by employing chain rule calculus layer by layer from the output to the input \cite{werbos1990backpropagation}.
Consider the simple network in Figure 2.1 that contains a single neuron and a set of weights $W$. The output of the neuron $O$, feeds into an activation function with an output $A$. The derivative of the error $E$ with respect to  particular weight $W_i$, can be described by a chain of partial derivatives backpropagating through the network.
\begin{equation}
\frac{\partial E}{\partial W_i} = \frac{\partial E}{\partial A} \frac{\partial A}{\partial O} \frac{\partial O}{\partial W_i}
\end{equation}




Classical gradient descent computes the gradient with respect to all samples in the training set for each weight update. While this approach leads directly to the optima of the loss function, it is computationally expensive for large datasets. Stochastic Gradient Descent instead updates the weights for every training sample individually. This approach is more efficient but adds variance as the gradient for each individual sample will not necessarily be in the direction of the true gradient of the training set. Minibatch Gradient Descent is most often used in deep learning, where the gradient is calculated over small subsets of the training data. This approach is more computationally efficient than full gradient descent while being less noisy than stochastic gradient descent \cite{ruder2016overview}.

Second order and derivative free methods have also been explored for optimizing neural networks.
Second order approaches like Newton's method and the conjugate gradient use curvature information through second order derivatives to achieve faster convergence \cite{battiti1992first}. 
However, the computational cost of these methods with deep neural networks can make them impractical for large-scale problems \cite{bottou2007tradeoffs}.
Derivative free methods like evolutionary strategies and particle swarm optimization eschew gradient information and require only some objective fitness measure. These approaches excel on tasks where the gradient is non-differentiable, unavailable, or unreliable. 
However, these methods tend to be sample inefficient on supervised learning tasks compared to gradient methods \cite{wright2006numerical}.

Adaptive optimizers like Adagrad, Adadelta, and Adam use running statistics of previous gradients to scale the learning rates for each parameter \cite{kingma2017adam, zeiler2012adadelta, duchi2011adaptive}.
This results in very fast convergence, however the solutions found by these methods tend to generalize worse than those trained with stochastic gradient descent \cite{wilson2017marginal}.
While these alternatives are effective in certain contexts, stochastic gradient descent is the most popular optimization algorithm used in deep learning for its reliability, simplicity, and generality.


\subsection{Architecture}

Recent advancements in neural network architectures have led to breakthrough results across a wide variety of machine learning tasks.
Many of these have been driven by the multilayer perceptron's notoriously poor performance with high dimensional image data.
These networks do not inherently account for spatial hierarchies as the input is flattened into a large one dimensional vector where each input neuron corresponds to a single feature.
As a consequence, these networks are incredibly prone to overfitting noisy pixel level details and even small images lead to an explosion in the number of trainable weights.

Convolutional layers were introduced to mitigate these problems by drawing inspiration from the biological network structures found in animal visual cortexes \cite{fukushima1980neocognitron}.
These layers scan the input through a convolution operation, using a set of filters to produce feature maps that capture spatial relationships between neighboring pixels.
Each filter uses the same shared window of weights for the entire input which drastically reduces the number of parameters required.
These layers enabled much deeper networks that are able to capture complex hierarchical features without overfitting.

\begin{figure}
    \centering
    \includegraphics[width=0.3\textwidth]{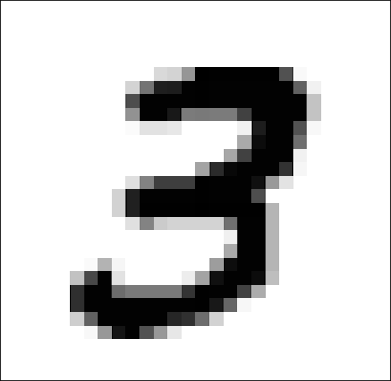}
    \hspace{0.2in}
    \includegraphics[width=0.3\textwidth]{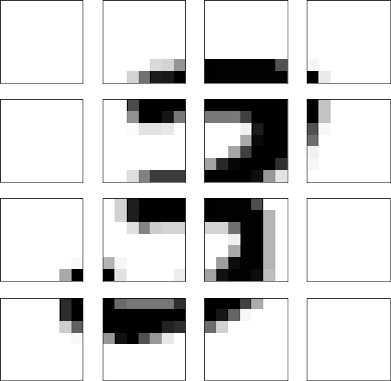}
    \hspace{0.2in}
    \includegraphics[width=0.3\textwidth]{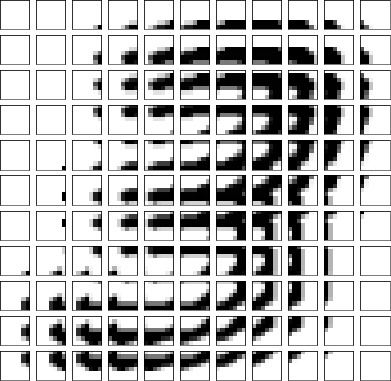}

    \caption{Visualizations of how convolutional layers operate on input images. Each convolutional filter in a layer scans over the input image according to a stride length. The middle image displays how a 7x7 filter with a stride of 7 applies to the input image. The right most image displays a 7x7 filter with overlap using a stride of 2.}
    \label{fig:my_label}
\end{figure}

Convolutional networks have continued to establish state of the art results on many benchmarks, spurred by the introduction of MNIST and the five layer LeNet in 1989 \cite{lecun1989backpropagation}.
Improvements have been progressive in the following years and limited mostly by the computational limitations of hardware at the time.
Notable architectures include the eight layer AlexNet, which kickstarted the deep learning revolution by winning the ImageNet competition in 2012, and the sixteen layer VGG model that won the ImageNet localization challenge in 2014 \cite{krizhevsky2012learning, simonyan2015deep}.
However, the use of deeper neural networks presented optimization challenges in the form of vanishing gradients \cite{pascanu2013difficulty}. 
This happens because backpropagation involves the repeated multiplication of partial derivatives through each layer from the output back to the input.
If these derivative are small, then the repeated multiplication can cause the gradient to diminish quickly such that weight updates in the early layers approach zero.
This is exacerbated by sigmoidal activation functions that saturate as their inputs become larger.

Residual Networks have since mitigated this problem by introducing skip connections between blocks of convolutional layers \cite{he2015deep}.
These connections bypass one or more layers by performing identity mapping where the input to a block is summed with the output of the layers being bypassed.
This reframes the underlying mapping $H(x)$ for a group of layers to a residual mapping $F(x) = H(x) - x$, creating shortcuts that allow gradients to propagate more effectively throughout the network.
Residual connections have enabled the effective training of substantially deeper networks while also exhibiting better convergence behavior during training, usually requiring fewer epochs to reach a stable and more accurate solution.

Vision Transformers have recently emerged as a powerful alternative to convolutional networks \cite{dosovitskiy2020image}.
Unlike convolutional layers that use small kernels to capture local features, self-attention mechanisms in Vision Transformers allow for global interactions where pixels or image patches are weighted in relation to every other pixel or image patch.
This approach offers the potential for better capturing spatial hierarchies, however it can be computationally expensive and require much more training data to avoid overfitting. 
Despite the promising results for Vision Transformers on massive datasets, deep residual convolutional networks are the de facto standard for computer vision tasks as they continue to achieve state-of-the-art results on many benchmarks.


\subsection{Subnetworks}

There has been a long history investigating the role of sparse subnetwork structures in deep learning research.
Drawing inspiration from the high degree of sparsity observed in biological neural networks, these structures offer significant potential for improving parametric efficiency, inference speed, memory requirements, and generalization.

A large body of research has emerged to explore how network pruning can find accurate subnetworks within deep neural networks by selectively removing unimportant weights \cite{blalock2020state}.
These methods offer significant potential for reducing computational complexity as a large number of parameters can be removed from a trained network while maintaining high performance after a small amount of continued training \cite{blalock2020state}.
Many techniques have been introduced to select unnecessary weights by using importance measures like weight magnitude, gradient saliency, connection sensitivity, and synaptic flow \cite{lecun1990optimal, lee2019snip, tanaka2020pruning}. 
While these methods can find extremely sparse subnetworks that achieve high accuracy, even random pruning has been shown to result in accurate subnetworks up to moderate levels of sparsity \cite{liu2022unreasonable, li2022revisiting}.


Sparse subnetwork structures are also implicitly leveraged in several important algorithmic advancements in deep learning.
The rectified linear unit (ReLU), defined as $f(x) = max(0, x)$, has gained widespread adoption as an activation function in deep neural networks. 
The activation patterns of ReLU neurons exhibit high levels of sparsity which are thought to improve the separability of high dimensional data \cite{glorot2011relu}.
Certain regularization methods leverage sparsity in parameter space in order to reduce overfitting and prevent feature reuse.
Dropout, DropConnect, and Stochastic Depth Networks are the most popular examples where random neurons, connections, or layers are masked on each forward pass through the network \cite{srivastava2014dropout, wan2013regularization, huang2016deep}.
Other effective regularization techniques include L1 and L2 weight decay penalties that aim to penalize large weights and push unimportant ones towards zero during training \cite{krogh1991simple}.
Weight decay can help to prevent overfitting, identify important features, and compress model sizes by learning sparse structures \cite{ma2019transformed}.


Other promising areas of neural network research aim to leverage the unique qualities of sparse networks to explore new models of neural computation.
Spiking neural networks use biologically inspired action potentials over time where neurons only activate if a threshold is crossed \cite{ghosh2009spiking, tavanaei2019deep}. 
These models hold potential for significantly faster inference and reduced energy consumption for resource constrained environments.
Echo State Networks and Liquid State Machines use recurrent reservoirs of sparsely connected neurons with fixed weights that have been proven to be particularly effective for dynamic time series modeling \cite{jaeger2007echo, maass2002real}.
Neuroevolutionary algorithms often evolve sparse network architecture along with the weights, excelling on tasks where gradient information is noisy or unavailable \cite{floreano2008neuroevolution, stanley2002neat, gaier2019wann}.
Super Networks are a theorized concept about how giant networks of the future will be too big to be trained by any one system, and instead portions of the network will need to be trained in a distributed manner by isolating sparse pathways \cite{fernando2017pathnet}.

\section{Ensemble Learning}

Many supervised learning algorithms can be described as search algorithms in a hypothesis space, where the goal is to train a set of model parameters to some candidate solution that best fits the target distribution \cite{blockeel2010hypothesis}. 
These hypothesis spaces are very large and complex in deep learning which makes finding the global optimum very difficult.
In practice, there are often many local optima that are good enough, however each will invariably have its own set of inherent biases \cite{choromanska2015loss}. 
Ensemble learning tries to remedy this problem by training several models that each converge to different local optima that are then combined together to form a better overall hypothesis.

This approach to improving generalization error is often explained in ensemble literature with an example of the bias-variance decomposition of the mean squared error (MSE) \cite{pretorius2016bias, ueda1996generalization, brown2005diversity}. Given a model prediction $\hat{y}$ and a true target value from an unknown test distribution $y$, the mean squared error is defined to be the expectation of the squared distance between the models predictions and the true target distribution. Bias is the difference between the expectation of the model and the true targets and variance is the squared difference between the models predictions and its mean.
\begin{gather}
bias = E[\hat{y}] - y \\
var = E[(\hat{y} - E[\hat{y}])^2] \\
MSE = E[(\hat{y} - y)^2] = (E[\hat{y}] - y)^2 + E[(\hat{y} - E[\hat{y}])^2] \\
MSE = bias^2  + var
\end{gather}

Given an ensemble of $M$ equally weighted estimators, the decomposition can be further extended to produce the bias-variance-covariance decomposition \cite{brown2005diversity}.
\begin{gather}
    \overline{bias} = \frac{1}{M} \sum_i(E[\hat{y}_i] - y) \\ 
    \overline{var} = \frac{1}{M} \sum_i E[(\hat{y}_i - E[\hat{y}_i])^2]\\
    \overline{covar} = \frac{1}{M(M-1)} \sum_i \sum_{j \neq i} E[(\hat{y}_i - E[\hat{y}_i])(\hat{y}_j - E[\hat{y}_j])] \\
    MSE = \overline{bias^2} + \frac{1}{M} \overline{var} + (1 - \frac{1}{M}) \overline{covar}
\end{gather}

Generalization error for single models relies upon the optimization of both bias and variance, where the tuning of a model towards high bias can cause it to miss important features and the tuning of a model toward high variance can cause it to be highly sensitive to noise. When the decomposition is extended to an ensemble, the generalization performance additionally depends on the covariance between models. Ideally, ensemble methods that promote diversity will be able to reduce covariance without increasing the bias or variance of individual models \cite{ueda1996generalization, brown2005diversity}.


Diversity is primarily encouraged through variation of the training data, network architecture, or training procedure for each ensemble member.
Several of the most popular traditional ensemble methods predate the deep learning revolution and instead use simple and fast classifiers (e.g. decision trees, support vector machines, logistic regression, etc.) to produce large numbers of models optimized on different variations of the training data \cite{schapire1990strength}.
Bootstrap Aggregation does this by training each model on a random subset of the training data with replacement \cite{breiman1996bagging}.
The Random Subspace Method instead trains each model on a random subset of features for each training sample \cite{ho1998random}.
Boosting strategically samples subsets of data to train models in sequence, where each subsequent model is trained to correct the mistakes of its predecessors \cite{drucker1995boosting, freund1996adaboost, chen2016xgboost, guolin2017lightgbm}.

Traditional ensemble methods are well founded mathematically by leveraging large numbers of diverse models to overcome weak base classifiers. 
However, this approach is ill-suited to the computationally intensive nature of deep neural networks, which tend to be data hungry and require large numbers of training samples.
These methods instead focus on training fewer models for longer periods, aiming to achieve high individual accuracy.
Data variation techniques that reduce the size of the training set for each member tend to result in lower individual accuracy and worse ensemble performance overall \cite{lakshminarayanan2017simple}.
Deep ensembles instead rely mostly on variations to network architecture (e.g. weight initializations, layer configurations, etc.) and training procedure (e.g. optimizer hyperparameters, loss functions, stochastic order of training samples, etc.) to induce diversity for each member \cite{lakshminarayanan2017simple, fort2020deep, hansen1990neural, liu1999ensemble}.




\subsection{Low-Cost Ensemble Methods}

Efficient ensemble algorithms have become increasingly important as they hold the potential to significantly reduce the computational cost associated with traditional ensemble learning, while retaining many generalization benefits that ensemble learning affords.
Several methods have been introduced that we broadly categorize as pseudo-ensembles, temporal ensembles, or evolutionary ensembles.
Pseudo-ensembles train several models under the guise of a single monolithic architecture.
Temporal ensembles encompass methods that save snapshots or checkpoints of a single model throughout training.
Evolutionary ensembles leverage biologically inspired concepts like mutation, recombination, and selection to spawn new child networks. 
While these methods vary in their approach to reducing cost, the primary idea behind each is to share either network structure, gradient information, or learned parameters among ensemble members. 
Table 2.2 provides a broad overview on the relative benefits and drawbacks of each type of method across a number of categories: training cost, inference cost, storage cost, member diversity, and ensemble size.

\begin{table}
\caption{A generalized overview of the benefits and drawbacks of various low-cost ensemble learning strategies. Specific methods within each category may differ significantly in regards to their strengths or weaknesses. All categories are relatively and subjectively graded according to the primary idea behind their cost reduction, where a dash represents baseline performance, a single checkmark corresponds to relatively good performance, and two checkmarks corresponds to best performance.}
\begin{tabularx}{\textwidth}{X c c c c c}
\toprule
Method & Training & Inference & Storage & Diversity & Ensemble Size \\
\midrule
Pseudo & \checkmark  & \checkmark \checkmark & \checkmark \checkmark & - & - \\
Temporal & \checkmark \checkmark & - & -  & \checkmark & \checkmark \\
Evolutionary & \checkmark & - & - & \checkmark \checkmark & \checkmark \checkmark \\
\bottomrule
\end{tabularx}
\end{table}

\subsubsection{Pseudo-Ensembles}


Pseudo-ensembles encompass a variety of methods that implicitly train ensembles as if they were a part of a single architecture \cite{bachman2014learning}.
These methods typically exploit variations of computational paths within the model to simulate the existence of multiple submodels.
Pseudo-Ensembles also tend to share parameters among members, distinguishing these methods from other ensembles where independent members are created and trained separately.




Dropout is the canonical example of a pseudo-ensemble where neurons are randomly masked during training on each forward pass \cite{srivastava2014dropout}. 
This technique effectively trains a unique subnetwork, or ensemble member, for each batch of training data.
Dropout is deactivated during inference so that the entire network is utilized, which results in a computation where each of the subnetworks are implicitly ensembled together.
Several variants of Dropout have since been introduced that mask different network structures.
DropConnect masks individual connections rather than neurons \cite{wan2013regularization}. 
Spatial Dropout masks feature maps in convolutional layers rather than neurons \cite{tompson2015efficient}.
Stochastic Depth Networks mask entire layers of deep residual networks \cite{huang2016stochastic}. 
Since these techniques mask large parts of the network during training, significantly more forward passes are needed to fully train a network.
Given enough training time, the resulting networks tend to converge to more robust optima than standard deep neural networks.

Some pseudo-ensemble methods explicitly encode relationships into the network architecture design.
TreeNets are built on this idea where the network architecture splits into several independent branches that each have their own output head \cite{lee2015m}. The early layers in the network serve as generalized feature extractors, while the independent branches learn specialized representations. During training, the shared layers accumulate the gradients from each independent branch and during testing, each path from input to one of the outputs can be treated as an independent ensemble member.
Multiple-Input Multiple-Output architectures extend the ideas introduced with TreeNets to also incorporate independent input branches. Each input head is fed different samples and each of the output heads is trained to predict the corresponding input \cite{havasi2021training}. During inference, the same input is fed to all of the branches and the resulting outputs are ensembled together.

BatchEnsembles utilize parameter sharing by decomposing weight matrices into an element wise product between a single shared set of base weights and member-specific rank-1 matrices \cite{wen2020batchensemble}. During training, the shared weights capture features that are common across all ensemble members while the member-specific weights can specialize. The rank-1 matrices for each ensemble member is represented with two single column vectors, which significantly reduces the storage cost compared to storing full rank weight matrices. Additionally, the element wise product is much cheaper computationally than matrix multiplication which results in little computational overhead compared to inference with a single network.

Knowledge Distillation was introduced to alleviate the inference and storage costs of ensembles and large networks. This approach is based on the idea that you can transfer the knowledge learned from a large network to a small network by training the small network using the predictions of a trained large network, instead of the raw training labels. This enables the small network to learn features that would be hard to disentangle if trained from scratch. This can be done in the context of ensemble learning by training a single network on the predictions made by a fully trained ensemble. The single network then approximates the behavior of the ensemble without needing to store and test all of the models separately \cite{hinton2015distilling}.

Stochastic Weight Averaging is a recent optimization technique where the weights of a model are averaged over several states taken from the final training epochs \cite{izmailov2019averaging}.
This approach effectively integrates several local optima into a composite model with implicit pseudo-ensemble like inference.
Model Soups operate on a similar principle and have recently shown state-of-the-art performance on large image classification benchmarks. Instead of taking the final epoch checkpoints of a single model, Model Soups average the weights of several independent models that are derived from the same pre-trained network but are each fine tuned with different hyperparameters \cite{wortsman2022model}.

Pseudo-ensembles are much more memory efficient than other low-cost ensemble methods as members are embedded within a single network structure. These techniques are generally very fast at performing inference, as forward passes do not need to be independently computed for each ensemble member. This makes these techniques highly valuable for environments where memory and inference cost is prioritized, like in embedded systems or edge computing. However, pseudo-ensembles do tend to need more training epochs to converge and the resulting models are less diverse than other ensemble methods, as large parts of the network structure are shared.

\subsubsection{Temporal Ensembles}

Temporal ensembles are built on the idea that ensemble members can be created by selecting and saving checkpoints over a single network's training trajectory \cite{swann1998fast}.
These approaches capitalize on the variability of model weights as they converge to several distinct optima throughout training.
Temporal methods primarily differ in how they determine which checkpoints to save and how to encourage the model to converge to diverse and accurate optima.

Fast Committee Learning pioneered this concept by selecting time slices at even intervals throughout training \cite{swann1998fast}.
Horizontal Voting Ensembles were later introduced that saved a contiguous number of states from every epoch taken late in the training phase \cite{xie2013horizontal}. 
This sacrifices diversity for member accuracy as contiguous checkpoints tend to be highly correlated, while checkpoints taken early in training will be much less accurate.

Snapshot Ensembles introduced a highly effective technique for manipulating the learning rate schedule by alternating between large and small values over predetermined cycles \cite{huang2017snapshot}. The large rates encourage the model to move further in parameter space before converging to local optima with small learning rates. The model is saved at the end of each cycle before repeating, striking a nice balance between member diversity and accuracy.
Fast Geometric Ensembles extend the algorithm introduced in Snapshot Ensembles by analyzing the loss surfaces found between the local optima saved at the end of each Snapshot checkpoint \cite{garipov2018loss}. The landscape analysis revealed that these local optima tend to be connected by high accuracy pathways. Fast Geometric Ensembles leverage these high accuracy pathways by saving model states along them. Interpolating the weights between these optima allows for much larger ensemble sizes as they are not constrained by the cycle length used in Snapshot Ensembles.

FreeTickets is a recent method that combines ideas from sparse neural network training and temporal ensemble learning \cite{liu2021freetickets}. This approach uses a dynamic sparse training algorithm that isolates a unique sparse subnetwork at each cycle which is then optimized with a cyclic learning rate schedule. At the end of every optimization cycle, the sparse model state is saved before updating connectivity for the next cycle. The unique topological structure of each member encourages greater diversity between members, however training sparse subnetworks from scratch tends to result in slower convergence.

Temporal Ensembles tend to be very computationally efficient for training as these algorithms save model states over a typical network training trajectory. The performance of these temporal ensembles often compete with traditional dense ensembles at a significant fraction of the compute cost. Inference and storage is more expensive than pseudo-ensembles since each ensemble member needs to be evaluated independently. Temporal ensembles also tend to have smaller ensemble sizes than evolutionary ensembles as member generation is dependent on training budget and the predetermined approach for selecting checkpoints. The members of certain temporal ensembles can struggle with diversity or accuracy as model checkpoints taken later in the training process tend to be highly correlated while earlier checkpoints tend to be less accurate.

\subsubsection{Evolutionary Ensembles}


Evolutionary methods utilize concepts inspired by biological evolution like natural selection, mutation, and recombination to optimize populations over time \cite{whitley1994genetic}.
These approaches have traditionally been concerned with single hypothesis testing, where optimization is geared towards selecting a single network with the best generalization performance from a population. 
However, several researchers have noted the natural connection between evolutionary populations and ensemble learning \cite{gagne2007evofree, liu2000evolutionary, yao2008evolving}. 

Evolution Strategies represent a popular subclass of evolutionary methods that describe a family of algorithms that generate child networks by selecting elite candidate models according to their fitness rankings and mutating them with noise perturbations \cite{beyer2002evolution}. These approaches are known to excel with wide scaling as candidate members can be cheaply generated and quickly evaluated in parallel \cite{salimans2017evolution}.

Several variations of evolution strategies have been introduced that alter the ways that child networks are created.
Natural Evolution Strategies generates new populations according to a weighted average over the previous population distribution \cite{wierstra2011natural}. This approach approximates a natural gradient as models are weighted according to their fitness such that mutations encourage models to move in the direction of previously successful models. 
Covariance Matrix Adaptation Evolution Strategies improves on evolution strategies by tracking dependencies between member parameters with a pairwise covariance matrix. The standard deviation can then control for the overall scale of the distribution, greatly improving exploration \cite{hansen2023cma, hansen2001completely}. However, despite the exploratory power of this approach, constructing the pairwise covariance matrix is computationally expensive and impractical for large networks.

Evolutionary algorithms are also commonly employed to evolve the topology of the network along with the weights. Neuroevolution of Augmenting Topologies is perhaps the most well known approach where networks are incrementally evolved from simple minimalist structures to more complex ones \cite{stanley2002neat}.
This approach also incorporates concepts like speciation to maintain a diverse population of network architectures with the goal of preventing premature convergence to suboptimal solutions.
Weight Agnostic Neural Networks employ a similar approach, however the weights between neurons remain fixed. \cite{gaier2019weight}. 
The focus then becomes entirely on evolving the architecture, where this approach interestingly found that network topology alone can encode solutions to complex reinforcement learning tasks.

Neuroevolutionary methods have primarily seen success with tasks that can be solved by relatively small networks, as these methods tend to be sample inefficient when scaling to larger models \cite{galvan2021neuroevolution}.
As a result, hybrid gradient/evolutionary methods have grown in popularity for deep neural networks as gradient optimization can be used to rapidly train the model while evolutionary processes can be used to improve exploration \cite{de2020fine}.
Evolutionary Ensembles with Negative Correlation Learning is a recent example of this hybrid methodology where gradient based optimization is applied to each member at each evolutionary generation. 
Multiple independent networks are trained simultaneously with the inclusion of a correlation penalty error term to encourage diversity. At each generation, the fittest individuals are selected as parents for the next generation, and child networks are spawned with normally distributed weight perturbations \cite{liu2000evolutionary}.
MotherNets is another method where a small base network is trained and used as a source to hatch larger child networks, where additional neurons and layers are added around the core network. The weights of the child networks are adjusted using function preserving network morphisms, in order to ensure that the added layers and neurons do not catastrophically affect the functionality of the core network. Diversity is encouraged through heterogenous network architectures where each child is optimized with different topological configurations \cite{wasay2020mothernets}.

These evolutionary approaches to optimization have several advantages over gradient based methods that are typically used in deep learning.
They are inherently scalable as population members can be evaluated independently and they excel at exploring landscapes where gradient information is noisy, flat, or unavailable \cite{salimans2017evolution}.
These low-cost evolutionary methods offer a unique advantage over other approaches in that it is generally cheap to generate new ensemble members. Once a parent network is trained, child networks tend to converge quickly while maintaining diversity in the parameter space. The decoupled nature of child generation and parent training enables more flexibility as large ensembles can be generated dynamically and pre-trained models could be used to save significant amounts of compute.

\subsubsection{Discussion}

Ensemble learning has long been known to be an effective and powerful way to improve performance and these low-cost methods have introduced ways to share information between members that can significantly reduce computational cost.
While there is no established taxonomy for low-cost methods in the literature, we note several commonalities where they can be organized under the categories of pseudo, temporal, and evolutionary ensembles.
With fixed training budgets and competing research objectives, these approaches often struggle to balance several tradeoffs relating to member accuracy, member diversity, training cost, inference cost, and storage cost.
These methods can suffer from restrictive algorithms that require special network structures or training procedures, which can make it difficult to utilize off the shelf models.
It can also be costly to generate and train new ensemble members which results in smaller ensemble sizes that can limit generalization potential.
These are increasingly important properties for ensemble methods as deep neural networks continue to grow in size.
To address these multifaceted challenges, we aim to introduce a framework that allows for flexibility in its implementation while also optimizing these tradeoffs to produce accurate, robust, and efficient ensembles.




\chapter{Theoretical Framework}
\label{chap:neural}
\section{Introduction}

We introduce Subnetwork Ensembles as an efficient low-cost approach to creating ensembles of deep neural networks.
The goal with this research is to develop a generalized framework for describing a process where a large parent network is trained to convergence and then used to generate child networks through a perturbative process applied to randomly sampled subnetworks in the parent model.
The spawned child networks can then be further optimized or tuned in order to recover the lost accuracy as a result of the perturbations.

The perturbative process, subnetwork sampling method, and tuning procedure used to create child networks can take many forms.
We explore several approaches in Chapters 4, 5, and 6 called Noisy Subnetwork Ensembles, Sparse Subnetwork Ensembles, and Stochastic Subnetwork Ensembles respectively.
Noisy Subnetwork Ensembles apply noise mutations to modify the weights of sampled subnetworks.
Sparse Subnetwork Ensembles prune the weights of sampled subnetworks to create sparse child networks.
Stochastic Subnetwork Ensembles uses probability distributions and Bernoulli realizations to create children with stochastic properties.
Figure 3.1 provides a visualization of child networks formed through these three methodologies.

\begin{figure}[t]
    \centering
    \includegraphics[width=\textwidth]{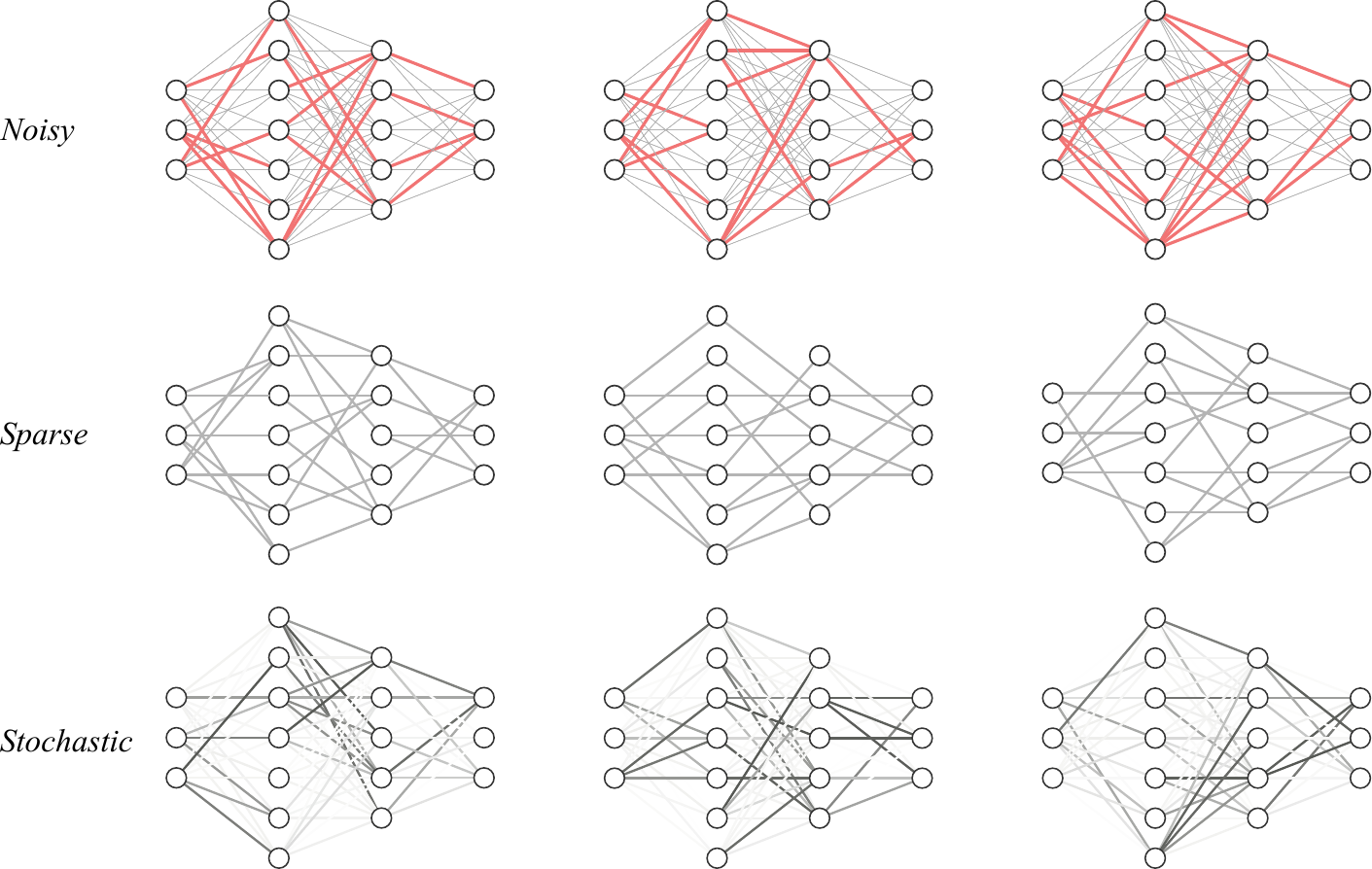}
    \caption{A visualization of child networks formed through three distinct methodologies that we investigate in Chapters 4, 5, and 6. Noisy Subnetwork Ensembles perturb subnetworks with noise. Sparse Subnetwork Ensembles prune subnetworks from the parent. Stochastic Subnetwork Ensembles use probability scores to determine active parameters on every pass through the network.}
    \label{fig:enter-label}
\end{figure}

Subnetwork Ensembles are motivated by several principles observed in deep learning optimization research.
Using a large parent network enables improved training dynamics and better convergence efficiency. Smaller networks trained from scratch often fail to achieve the same performance \cite{li2020train, tan2020efficientnet, hestness2017deep}.
Large parent networks are extremely overparameterized and there's a notion that trained models contain large amounts of redundancy \cite{blalock2020state, liu2022unreasonable}. Creating children from the subspaces of parent networks leverages unrealized capacity to improve parametric utilization.
Spawned child networks inherit optimized parameters that enable convergence to new local optima quickly. This property has long been observed in both pruning and transfer learning literature \cite{blalock2020state, zhuang2020comprehensive, yosinski2014transferable}.
Perturbing different subspaces of the parent network encourages diverse representations among child networks. This is especially true for sparse subnetworks where altered network topologies impact the underlying optimization landscape \cite{stier2019structural}.

This approach also offers several unique advantages over competing low-cost ensemble methods.
Since the parent and child network procedures are completely decoupled, any parent network architecture or optimization algorithm can be used.
Modern deep learning best practices (dropout, normalization layers, weight decay, adaptive optimizers, data augmentations, learning rate schedules, etc.) can be used to maximize performance of the parent \cite{kukacka2017regularization}.
This also opens the door to leveraging open source pre-trained models, which is an increasingly valuable area of research as models and datasets continue to grow in size.
It is also relatively fast and cheap to generate and train new child networks. 
This allows for the ability to dynamically grow ensemble sizes as needed, which can be very useful for some types of tasks where the problem space grows \cite{caruana1997multitask, wang2023comprehensive}.
This also allows for wide scaling as child networks can be generated and trained in parallel, avoiding the limitations of low-cost ensemble methods that are restricted to taking models along a sequential training trajectory \cite{huang2017snapshot, garipov2018loss, laine2017temporal}.
Sparse Subnetwork Ensembles that leverage pruning mechanisms result in much smaller ensembles that significantly reduces memory requirements.

\section{Formalization}

Consider a parent neural network $F$ parameterized with a set of weights $\theta$ optimized via some standard training procedure. 
We create an ensemble $E$ consisting of $k$ child networks, where each child network $f_i$ has weights $\hat{\theta}_i$ obtained through a perturbative process $\xi$ applied to a subset of weights from the parent network determined using a stochastic sampling process $\phi$.
\begin{align}
E &= \{f_1, f_2, ..., f_k\} \\
f_i(x; \hat{\theta}_i) &= F(x; \theta; \xi, \phi)
\end{align}

The generality of this approach offers a lot of freedom for deciding how Subnetwork Ensembles are formed. 
The perturbative and subnetwork sampling processes used to create child networks can take arbitrary forms which can be useful for tailoring this algorithm to a wide variety of tasks.
Any network architecture can be used with this approach, however it is important to note that the parent network does serve a foundational role in ensemble performance, such that the parent network should be both accurate and have sufficient capacity to enable the creation of high performing children.

\subsection{Subnetwork Sampling}

Subnetworks are represented with binary bit mask matrices with the same dimensions of the weight matrices for each layer in the parent network. Consider a weight matrix $W \in \mathbb{R}^{m \times n}$ representing the weights of a particular layer in a neural network. We generate a matrix $M \in \{0,1\}^{m \times n}$, using a stochastic process $\phi$, with the same dimensions as W. The elements of M are binary values, where $M_{ij} = 1$ if the corresponding weight $W_{ij}$ is retained, and $M_{ij} = 0$ if the weight is masked. The subnetwork weights $\hat{W}$ can then be computed by taking the element-wise (Hadamard) product of W and M.
\begin{align}
M &\sim \phi^{m \times n} \\
\hat{W} &= W \circ M
\end{align}



The topology of the subnetwork can be further described by the granularity in which parameters are masked. 
This granularity can be unstructured or structured, where the former refers to weight-level or connection-level pruning and the latter refers to neuron-level, channel-level, or layer-level pruning.
Unstructured pruning tends to result in networks with better predictive capacity while structured pruning results in better computational efficiency. 
Removing entire rows, columns, or blocks from a network is more easily leveraged by hardware optimizations \cite{blalock2020state, li2017fast}. 
However, emerging hardware technology offers a promising future for exploiting unstructured sparse computation \cite{nvidia2020a100, graphcore2022ipu, cerebras2022wafer, elsen2019fastsparse}.
\begin{equation}
(\hat{W}_{uns},\  \hat{W}_{str}) = \left( \begin{bmatrix}
w_{11} & w_{12} & 0 & w_{14} \\
0 & w_{22} & 0 & 0 \\
w_{31} & 0 & 0 & w_{34} \\
0 & 0 & w_{43} & w_{44}
\end{bmatrix} 
\begin{matrix} \\ \\ \\, \end{matrix}
\begin{bmatrix}
w_{11} & w_{12} & w_{13} & w_{14} \\
0 & 0 & 0 & 0 \\
w_{31} & w_{32} & w_{33} & w_{34} \\
0 & 0 & 0 & 0
\end{bmatrix}
\right)
\end{equation}

It's important to also consider the distribution of masked weights throughout non-homogenous networks. It's common for layer configurations in deep neural networks to vary significantly in size. Severe pruning of small layers may result in bottlenecks that restrict gradient flow. The distribution of masked weights can be controlled with global and local masking methods. Global methods are applied uniformly across the entire network while local methods are applied independently within each layer or sub-region. Global methods tend to result in higher compression rates as more parameters from the larger layers are removed whereas local methods offer more fine-grained control and reduced variance.

\subsection{Neural Partitioning}

The transformation of unique subnetworks among child networks is critical at encouraging diverse behavior.
However, the random nature of subnetwork sampling can be a large source of variance when constructing child networks.
It is prudent to reduce the amount of overlap between the subnetworks inherited by the parent in order produce less correlated models.
We introduce Neural Partitioning as an effective tool for achieving variance reduction by dividing neural networks into sets of opposed child networks.



These ideas are inspired by antirandom testing and antithetic sampling \cite{malaiya1995antirandom,shen2008antirandom, hammersley1956antithetic}. This is also known as mirrored sampling in evolutionary communities, where whenever a random vector $v = \{x_1, ..., x_n\}$ is generated, another vector is generated with the opposite signs $v' = \{-x_1, ..., -x_n\}$ \cite{brockhoff2010mirror}. Antithetic sampling takes advantage of the fact that the variance of dependent random variables depends on the covariance between them.
\begin{align}
    var(X) &= E[(X - E[X])^2] \\
    var(X + Y) &= E[((X+Y) - E[(X+Y)])^2] \\
    &= E[(X - E[X])^2] + E[(Y - E[Y])^2] + E[2(X - E[X])(Y - E[Y])] \\
    &= var(X) + var(Y) + 2 cov(X,Y)
\end{align}

If X and Y are independent, then the variance of the sum of them is equal to the sum of each. However, if we sample Y such that it's dependent on X, then the variance is reduced if the covariance between them is negative.

With Neural Partitioning, we can leverage this idea by generating groups of subnetworks such that the inherited parameters are antithetical and form a disjoint union over the parameters of the parent network. Consider a weight matrix $W \in \mathbb{R}^{m \times n}$ for a particular layer in a network. Consider a mask $M \in \{0, 1\}^{m \times n}$ that has been generated via some subnetwork sampling process. The corresponding subnetwork mask with no parameter overlap, also called the anti-mask $M'$, is one in which the polarity of all of the mask bits are flipped $M' = 1 - M$.
\begin{equation}
(M,\  M') = \left( \begin{bmatrix}
1 & 1 & 0 & 1 \\
0 & 1 & 0 & 0 \\
1 & 0 & 0 & 1 \\
0 & 0 & 1 & 1
\end{bmatrix} 
\begin{matrix} \\ \\ \\, \end{matrix}
\begin{bmatrix}
0 & 0 & 1 & 0 \\
1 & 0 & 1 & 1 \\
0 & 1 & 1 & 0 \\
1 & 1 & 0 & 0
\end{bmatrix}
\right)
\end{equation}


For large networks, this idea can be extended to partition the parameter space into $k$ sibling subnetworks with a sparsity level of $1 - \frac{1}{k}$. However, creating too many partitions can lead to increasingly sparse subnetworks that exhibit bottleneck effects.
This can reduce information flow throughout the network and adversely affect both model performance and training efficiency.
\begin{equation}
(M_1,\  M_2,\ M_3) = \left( 
\begin{bmatrix}
0 & 1 & 0 & 1 \\
0 & 1 & 0 & 0 \\
1 & 0 & 0 & 0 \\
0 & 0 & 1 & 1
\end{bmatrix} 
\begin{matrix} \\ \\ \\, \end{matrix}
\begin{bmatrix}
0 & 0 & 1 & 0 \\
1 & 0 & 0 & 1 \\
0 & 0 & 1 & 0 \\
1 & 0 & 0 & 0
\end{bmatrix}
\begin{matrix} \\ \\ \\, \end{matrix}
\begin{bmatrix}
1 & 0 & 0 & 0 \\
0 & 0 & 1 & 0 \\
0 & 1 & 0 & 1 \\
0 & 1 & 0 & 0
\end{bmatrix}
\right)
\end{equation}

\begin{equation}
(M_1,\  M_2,\ M_3) = \left( 
\begin{bmatrix}
0 & 1 & 0 & 1 \\
0 & 1 & 0 & 0 \\
1 & 0 & 0 & 0 \\
0 & 0 & 1 & 1
\end{bmatrix} 
\begin{matrix} \\ \\ \\, \end{matrix}
\begin{bmatrix}
0 & 0 & 1 & 0 \\
1 & 0 & 0 & 1 \\
0 & 0 & 1 & 0 \\
1 & 0 & 0 & 0
\end{bmatrix}
\begin{matrix} \\ \\ \\, \end{matrix}
\begin{bmatrix}
1 & 0 & 0 & 0 \\
0 & 0 & 1 & 0 \\
0 & 1 & 0 & 1 \\
0 & 1 & 0 & 0
\end{bmatrix}
\right)
\end{equation}

The goal of Neural Partitioning is to generate subnetwork masks that maximize the Cartesian distance between them. For an ensemble of $k$ subnetwork masks $E = \{M_1, M_2, ... M_k\}$, we wish to maximize the total pairwise distance between all ensemble subnetworks. Consider the masks for weight matrices flattened into vectors $A = \{a_1, ..., a_N\}$ and $B = \{b_1, ..., b_N\}$. The Cartesian distance ($CD$) and total Cartesian distance ($TD$) for an ensemble can be described with:
\begin{align}
CD(A,B) &= \sqrt{|a_1 - b_1| + ... + |a_N - b_N|} \\
TD &= \sum_{i=1}^{k} \sum_{j=1}^{k} CD(M_i, M_j)
\end{align}

\subsection{Perturbative Processes}

There are a wide variety of ways for defining and applying a perturbative process to a parent network in order to generate child networks.
The choice of perturbation mechanism can influence distinct aspects of ensemble performance.
We explore three methodologies in Chapters 4, 5, and 6 that have various levels of regularization effects and computational requirements.

In chapter 4, we explore how child networks can be created simply by perturbing subnetwork weights with random noise. Consider a noise matrix $N \in \mathbb{R}^{m \times n}$, with the same dimensions of the corresponding layer's weight matrix, generated with random values drawn from a Gaussian distribution $\mathcal{N}(\mu, \sigma^2)$. Assume that a bit mask $M \in \{0, 1\}^{m \times n}$ was generated with a sampling process $\phi$ to determine the subnetwork that the perturbations should apply to. The perturbed weights $\hat{W}$ are then computed by adding the masked noise values to the original weights $W$.
\begin{align}
M &\sim \phi^{m \times n} \\
N &\sim \mathcal{N}(\mu, \sigma^2)^{m \times n} \\
\hat{W} &= W + (N \circ M)
\end{align}

In chapter 5, we prune subnetwork weights to create sparse topological structures. The perturbation can be described as setting certain parameters indicated by the subnetwork mask to zero and freezing them so that no further weight updates affect them during optimization. This form of perturbation offers a stronger regularization effect than those created with noise perturbations. However, the child networks need to be optimized further to recover the accuracy lost via pruning.
\begin{equation}
\hat{W} = \begin{bmatrix}
w_{11} & w_{12} & 0 & w_{14} \\
0 & w_{22} & 0 & 0 \\
w_{31} & 0 & 0 & w_{34} \\
0 & 0 & w_{43} & w_{44}
\end{bmatrix} 
\end{equation}

In chapter 6, we explore how stochasticity can be injected into the perturbation process by leveraging probability matrices that determine how likely it is that a given parameter is retained. Consider a subnetwork mask M generated from a random sampling process $\phi$. The mask is then used as a foundation for a function $\psi(M)$ to construct a probability matrix $P \in [0, 1]^{m \times n}$ where each value corresponds to the probability that a parameter is retained or masked. The subnetwork mask is determined on every forward pass through a Bernoulli realization of the probability matrix. We further experiment with annealing the probability matrix over time such that the subnetwork mask evolves to become deterministic throughout training.
\begin{align}
M &\sim \phi^{m \times n} \\
P &\sim \psi(M)^{m \times n} \\
\hat{W} &= W \circ Bernoulli(P)
\end{align}



\subsection{Network Optimization}

Once we have a collection of child networks, they can be further optimized to recover lost accuracy as a result of the perturbative processes.
The fine tuning procedure generally consists of continued training of each child network independently.
Since the child networks inherit parameters from a fully trained parent, they tend to converge in relatively few epochs.
Moreover, unique topological transformations as a result of pruning or stochastic masking can promote convergence to diverse local optima.
Neuroevolutionary methods that leverage fitness evaluations and model selection strategies can also be used as a form of ensemble optimization. Ultimately, the goal is to construct an ensemble that minimizes the expected loss $\mathcal{L}(\hat{y}, y)$ of the averaged predictions $\bar{f}(x)$ for each $(x,y)$ pair over the dataset $d_{xy}$.
\begin{equation}
\text{minimize}  \underset{(x,y) \sim d_{xy}}{\mathbb{E}} \mathcal{L}(\bar{f}(x), y)
\end{equation}

\subsection{Ensemble Predictions}

There are several approaches for combining model predictions in the ensemble. The two most popular methods include majority vote and weighted model averaging, where voting methods operate on the predicted class labels and averaging methods take the weighted mean of the output distributions of each member \cite{fragoso2017bayesian, dzeroski2004combining, vanderlaan2007super}.  
\begin{align}
\bar{f}_{\text{vote}}(x) &= \text{mode} \{\text{argmax}(f_i(x)) : i \in 1, 2, ..., k\} \\
\bar{f}_{\text{avg}}(x) &= \text{argmax} \left( \frac{1}{k} \sum_{i=1}^{k} w_i f_i(x)\right)
\end{align}

While majority vote has a long history with traditional ensemble methods, averaged predictions have become the standardized approach for ensembles of deep neural networks. Throughout this research, we consider ensemble predictions given by the mean of equally weighted ensemble members. Perturbations can result in networks that produce outputs of very different magnitudes. In order to prevent the outsized influence by specific models, the raw output logits of each model are normalized by a softmax function before averaging. This is typically done in classification tasks to convert outputs into probability distributions.
\begin{equation}
softmax(x_i) = \frac{e^{x_i}}{\sum_j e^{x_j}}
\end{equation}

\subsection{Evaluation Metrics}

The bulk of our benchmark experiments in the following chapters focus on typical image classification tasks. In these contexts we report standardized metrics including the accuracy, negative log likelihood, and expected calibration error of ensemble predictions on holdout test sets \cite{kumar2020verified}. Accuracy reports the fraction of correct predictions (Eqn. 3.15). Negative Log Likelihood is a common loss function used for classification tasks in deep learning that measures the difference between predicted probability distributions and true output distributions (Eqn. 3.16). Expected Calibration Error is used for uncertainty estimation to measure how well a model's estimated probabilities match the true observed probabilities of its predictions (Eqn. 3.17). This involves splitting the data into several bins and taking the average difference between the accuracy and confidence of the predictions within each bin.
\begin{align}
Acc &= \frac{1}{N} \sum_{i=1}^N 1(argmax \ \bar{f}(x_i) \equiv argmax \ y_i) \\
NLL &= - \frac{1}{N} \sum_{i=1}^N \sum_{c=1}^C y_{i,c}\ log(\bar{f}(x_{i})_c) \\
ECE &= \sum_{i=1}^B \frac{|B_i|}{N} \left| (acc(B_i) - conf(B_i) \right|
\end{align}

\noindent where $\bar{f}(x)$ represents the prediction made by the ensemble, $y$ represents the correct output for input $x$, $N$ is the total number of test samples, $|B_i|$ is the number of predictions in the bin, $acc(B_i)$ is the accuracy of the predictions made in the bin and $conf(B_i)$ is the average confidence of model predictions in the bin.

\chapter{Noisy Subnetwork Ensembles}
\label{chap:noisy}
\section{Introduction}

\begin{figure}
    \centering
    \includegraphics[width=\textwidth]{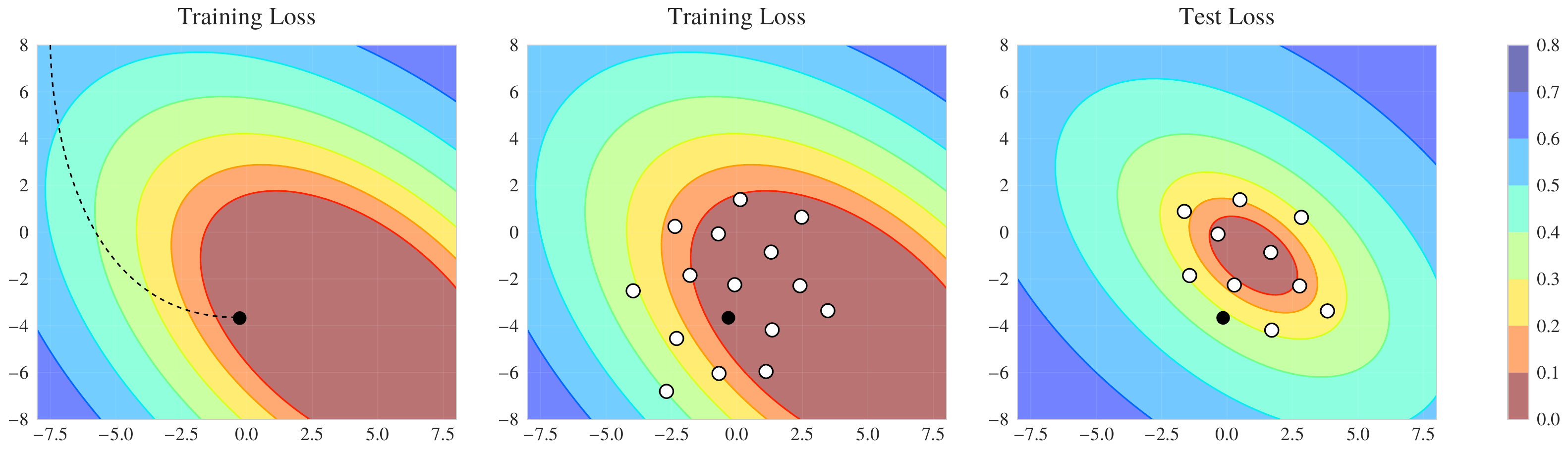}
    \caption{Neuroevolution excels at fine tuning fully-trained networks that get stuck in flat loss basins. The left graph displays a typical SGD training trajectory where models tend to converge to the edges of flat optima \cite{izmailov2019averaging}. The middle graph displays white dots as child networks that are generated from sparse mutations. The right graph shows the final ensemble consisting of top candidates selected from evaluation on a separate validation set. A key insight to the success of gradient-free methods for fine tuning is that the loss landscapes of test distributions rarely match the training distribution exactly. Ensembling over wide areas of good validation performance is key to improving generalization.}
    \label{fig:my_label}
\end{figure}


Neuroevolution is an area of machine learning research that aims to combine evolutionary algorithms and neural networks by using biologically inspired concepts like natural selection, mutation, and reproduction \cite{back1993overview,whitley2001overview, whitley1994genetic}. 
A core component of evolution strategies involves the generation of child networks by applying noise perturbations to one or more parent networks \cite{beyer2002evolution}.
However, as neural networks continue to grow in size, the effect of noise perturbations on network behavior becomes increasingly pronounced.

There is often a critical region of mutation where weak perturbations fail to provide any diversity while strong perturbations lead to complete performance collapse. 
The optimal mutation window can be vanishingly small with modern deep neural networks that contain millions or billions of parameters.
Due to the difficulty in mutating these large models, neuroevolutionary methods have primarily seen success with tasks that can be solved by relatively small networks \cite{galvan2021neuroevolution}.
One approach to alleviating this challenge with large models is to implement sparse mutations, where only a small subset of parameters are modified.
This can widen the critical mutation window which allows models to handle stronger perturbations before succumbing to performance collapse.
While sparse mutations have long been used in other areas of evolutionary and genetic programming, they have rarely been explored with evolution strategies using deep neural networks \cite{beyer2002evolution, hansen2023cma, khadka2018evolution}.


These observations of neuroevolutionary methods connect naturally to our work with Subnetwork Ensembles. 
In this chapter, we introduce a formulation that we call \textit{Noisy Subnetwork Ensembles} where child networks are generated through noise perturbations applied to random subnetworks in a large, pre-trained parent network.
Generating child networks in this manner can alleviate the problems with sample inefficiency that typically plague neuroevolutionary methods.
Since child networks retain their network structure and are only perturbed by noise, we eschew continued training of the children and instead implement a standard evolutionary model selection strategy, where we evaluate a large population of models and form the ensemble with those that perform best on a small validation set.

We conduct several ablation studies and visualizations designed to gain insight into the effect that mutation strength and sparsity have on network behavior.
We evaluate Noisy Subnetwork Ensembles on the benchmark reinforcement learning environment ProcGen as well as the large scale ImageNet classification task \cite{cobbe2020leveraging, krizhevsky2012learning}.
We observe reliable and consistent generalization improvements with over a dozen different deep neural network architectures on several difficult tasks.
This approach is flexible and strikes a nice balance with computational cost as these ensembles can improve performance over fully trained parent networks with no additional training required.

\section{Implementation}

Noisy Subnetwork Ensembles begin with the standard training of a large parent network $F$, parameterized with weights $\theta$.
We then spawn $k$ child networks by copying and perturbing a subset of the weights. 
The binary subnetwork mask $M \in \{0, 1\}$ is generated for each layer with the same dimensions as the layer weights $W$. The mask can be determined according to any random process $\phi$ for a given sparsity target $\rho$. In our implementation, we use a layerwise unstructured approach where we randomly assign the value for each parameter in the mask by sampling a Bernoulli distribution with probability $\rho$, where a successful trial denotes that the parameter will be retained.
We also generate a noise matrix $N$ with the same dimensions as the weight matrix according to some given random distribution.
We use a Gaussian distribution with tunable parameters for the mean $\mu$ and variance $\sigma^2$.
The sparse mutation for each layer is created by taking the Hadamard product $\circ$ of the subnetwork mask $M$ with the noise matrix $N$.
\begin{align}
M &\sim Bernoulli(\rho) \\
N &\sim \mathcal{N}(\mu, \sigma^2) \\  
\hat{W} &= W + (N \circ M)
\end{align}

\subsection{Trust-Region Search}

It is important to determine appropriate values for both the mean $\mu$ and variance $\sigma^2$ of the noise distribution as well as the amount of sparsity $\rho$ in the subnetwork mask.
This can be challenging to determine before hand due to differences between the size and complexity of datasets, model architectures, optimization hyperparameters, etc.

We implement a grid search over a small validation set, where we measure the average differences between the outputs of a network before and after mutations are applied. Ensuring that the outputs of the network diverge within some bound is referred to as trust region optimization in machine learning contexts \cite{conn2000trust}. This is a popular technique in reinforcement learning algorithms as they promote stable policy updates in environments where too large of a change may lead to performance collapse such that the model will struggle to recover \cite{schulman2017trust, schulman2017proximal}.

A standard approach for establishing the trust region would be to measure the squared difference between output predictions before and after mutation. Given a network $F$ that is parameterized by $\theta$ with $C$ outputs and a network that is perturbed with a noise mutation $\gamma$, the mean squared difference over a set of $N$ total input samples $X$ can be described as:
\begin{equation}
MSE = \frac{1}{N} \sum_{n=1}^N \sum_{c=1}^C (F(X_n; \theta)_c - F(X_n; \theta + \gamma)_c)^2
\end{equation}

However, since classification networks are trained with cross entropy loss, where outputs are treated as probability distributions, the mean squared error may not be the best fit for approximating the effect of that perturbation as it can be prone to increased variance. Instead, it is more common to describe the difference between output distributions with the Kullback-Leibler Divergence, also known as relative entropy. This allows for a stable measure of comparison that works regardless of network architecture or noise distribution.
\begin{equation}
KL = \frac{1}{N} \sum_{n=1}^N \sum_{c=1}^C F(X_n; \theta)_c \log \left( \frac{F(X_n; \theta)_c}{F(X_n; \theta+\gamma)_c} \right)
\end{equation}

A grid search algorithm can then be implemented to try out several values for the mutation strength and sparsity parameters in order to target a divergence score that represents the boundary of a trust region. 
One nice property revealed in our experiments is that there is a roughly linear relationship between KL divergence and accuracy degradation in trained models. 
This insight can be used to tweak mutation parameters according to problem complexity, ensemble size, or parameter sensitivity where a larger KL target can allow for stronger mutations and better exploration at the cost of potentially less accurate candidate models.

\begin{algorithm}
\SetAlgoLined
\KwIn{N, the number of candidate models}
\KwIn{K, the number of ensemble members}
\KwIn{($X_{train}, X_{val}, X_{test}$), \text{the training, validation, and test data}}
\KwIn{$\eta$, \text{the optimizer hyperparameters}}
\KwIn{$\Delta$, trust region boundary target}
\KwIn{$\phi(\rho)$, \text{the subnetwork sampling process with target sparsity $\rho$}}

\texttt{\\}
$O = sgd(\eta)$ \\
$F = initialize()$ \\
$F.w = train(F, O, X_{train})$ \\
\texttt{\\}
$\rho, \mu, \sigma^2 = trustRegionSearch(F,\ X_{val},\ \Delta)$ \\
\texttt{\\}
$candidates = []$ \\
\texttt{\\}
\For{i in 1 to N}{
    $f_i = initialize()$ \\
    $f_i.w = F.w$ \\
    \texttt{\\}
    \For{$j$ in $f_i.layers$}{
        $M \sim \phi(\rho)^{f_{i,j}.w}$ \\
        $N \sim \mathcal{N}(\mu, \sigma^2)^{f_{i,j}.w}$ \\
        $f_{i,j}.w = F_{i,j}.w + (N \circ M)$ \\
    }
    \texttt{\\}
    $fitness = evaluate(f_i, X_{val})$ \\
    $candidates \leftarrow (f_i, fitness)$ \\
}
\texttt{\\}
$ensemble = select(candidates, K)$ \\
\texttt{\\}
\For{$(x, y)$ in $X_{test}$}{
    $outputs = [F(x)$] \\
    \texttt{\\}
    \For{c in ensemble}{
        $outputs \leftarrow c(x)$ \\
    }
    \texttt{\\}
    $predictions = softmax(mean(outputs))$ \\
    $loss = nll(predictions, y)$ \\
    $accuracy = acc(predictions, y)$ \\
}
\caption{Noisy Subnetwork Ensembles}
\end{algorithm}

\subsection{Mirrored and Partitioned Noise}

Using the above techniques, we can then generate a collection of child networks with subnetworks perturbed according to parameters found through this trust region grid search.
Neural Partitioning and Antithetical Sampling can also be leveraged when generating child networks \cite{brockhoff2010mirror}. A single generated subnetwork mask $M$ and noise matrix $N$ can be used to generate four child networks $\hat{W}_i$ perturbed in opposite directions and subspaces, resulting in reduced variance and a balanced exploration of the parameter space.
\begin{align}
M &\sim \text{Bernoulli}(\rho) \\
N &\sim \mathcal{N}(\mu, \sigma^2) \\
\hat{W}_1 &= W + (N \circ M) \\
\hat{W}_2 &= W + (N \circ (1 - M)) \\
\hat{W}_3 &= W - (N \circ M) \\
\hat{W}_4 &= W - (N \circ (1 - M))
\end{align}

We can gain further improvements (without conducting any additional optimization) by implementing a model selection process.
For each candidate child network in the population, we evaluate its performance on a holdout validation set where the accuracy is recorded as a measure of its fitness. 
The top $k$ models with the best performance will then be selected to become members of the final ensemble. 




\section{Experiments}


\subsection{Decision Boundaries}

\begin{figure}[t]
    \includegraphics[width=\textwidth]{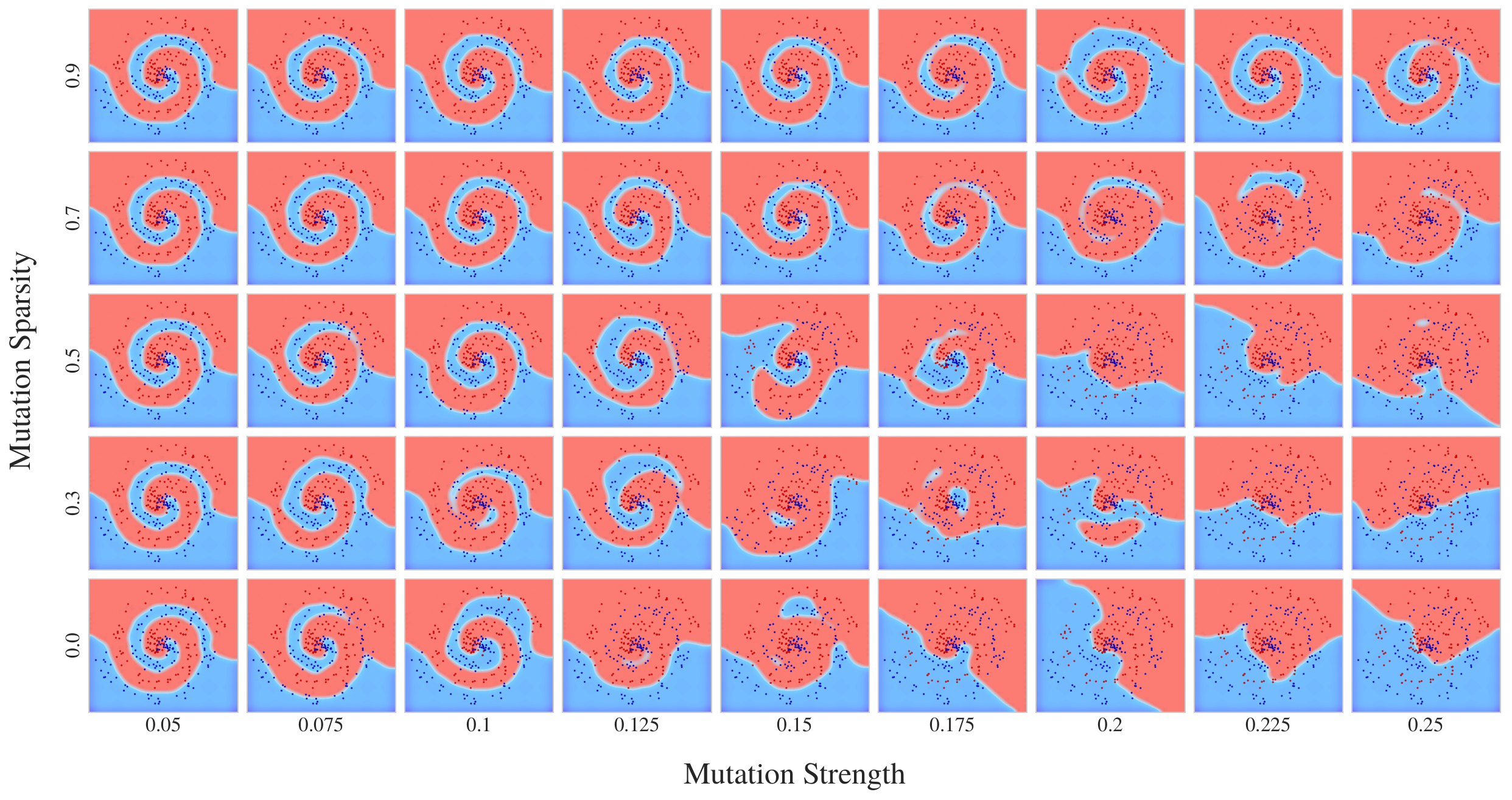}
    \caption{The images above detail decision boundaries for a trained three layer multilayer perceptron after being perturbed with various mutations. With dense perturbations, there is a comparatively small window between a functional decision boundary and complete performance collapse. As mutations become more sparse, the model is better able to retain its behavior while the strength of parameter mutation increases.}
\end{figure}

We begin by exploring the interplay between mutation sparsity and mutation strength by visualizing how changing these values affects network predictions.
We use a simple three layer fully connected multilayer perceptron that contains 64 neurons in each layer.
We train this model on a binary interleaved spiral dataset that contains 2500 sample (x, y) points. 
We train for 10 epochs using the Adam optimizer \cite{kingma2017adam} with a learning rate $0.001$ and use this trained model as the starting point for all mutations.

We then perturb the model with mutations sampled from a Gaussian Distribution $N \sim \mathcal{N}(0, \sigma^2)$ applied to random subnetworks with varying levels of sparsity. The subnetwork masks are generated in a layerwise unstructured fashion by sampling $M \sim Bernoulli(\rho)$ where $\rho$ corresponds to the sparse probability. We evaluate several values for the variance of the noise perturbation $\sigma \in [0.05, 0.25]$ and for subnetwork sparsities $\rho \in [0.0, 0.9]$.

Using the perturbed models, we make predictions on a holdout test set containing 250 samples. In figure 4.2, we display a grid of the decision boundaries of these perturbed models on the test set as we ablate between mutation strength and sparsity.
When mutations are dense, we see a very quick collapse of performance for small perturbations. The window for optimal dense mutation is quite small even for this toy network and simple dataset. When mutations become more sparse, the model is able to maintain good classification boundaries while displaying small variations of prediction diversity. This kind of behavior is desirable for neuroevolutionary populations as ensemble learning research has shown that ensembles perform better with large numbers of both accurate and diverse members \cite{bonab2016theoretical}.

\subsection{Ablations}

Next, we aim to explore whether the intuitions about how mutations affect predictions translate to a much larger convolutional network on benchmark computer vision datasets.
For this experiment, we conduct ablations on the CIFAR-10 and CIFAR-100 datasets \cite{krizhevsky2012learning}. These are popular benchmark datasets and their use is widespread in computer vision research. They each contain 50,000 training and 10,000 test samples of colored 32x32 pixel images. CIFAR-10 contains samples belonging to one of 10 classes while CIFAR-100 contains samples belonging to one of 100 classes.
We use a WideResNet-28x10 model for our parent network, which is a highly accurate network architecture that contains $\sim 36M$ parameters. This network is a variant on the popular ResNet that decreases the depth and increases the width of each convolutional layer \cite{zagoruyko2017wide}.

We implement a standard training algorithm for this type of model where we train for 100 epochs using Stochastic Gradient Descent with momentum \cite{sutskever13nesterov}.
A stepwise learning rate decay is used where an inital value of 0.1 decays to 0.01 after 50\% of training and decays again to 0.001 for the final 10\% of training.
We use standard data augmentation schemes for CIFAR that includes a random crop and random horizontal flip along with mean standard normalization.
We split the test set in half in order to construct a validation set used for model selection.

\begin{figure}[!ht]
    \centering
    \includegraphics[width=\textwidth]{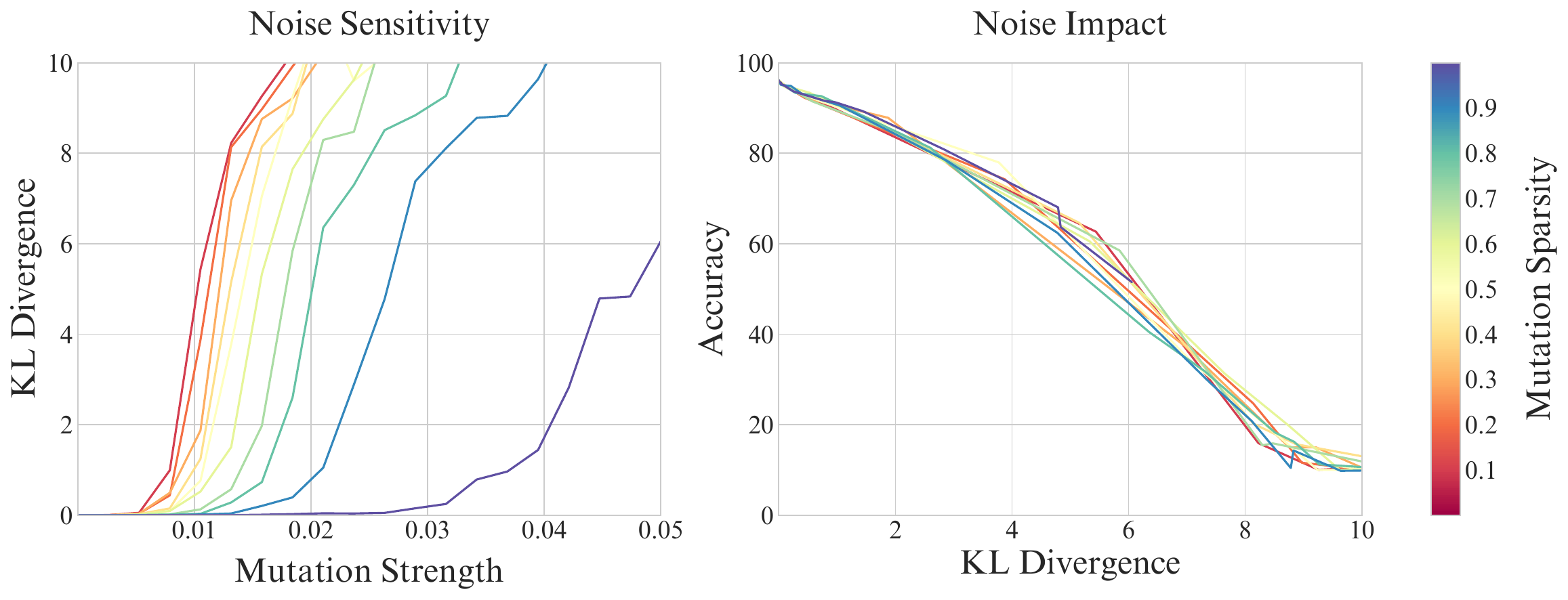}

    \vspace{0.1in}
    
    \includegraphics[width=\textwidth]{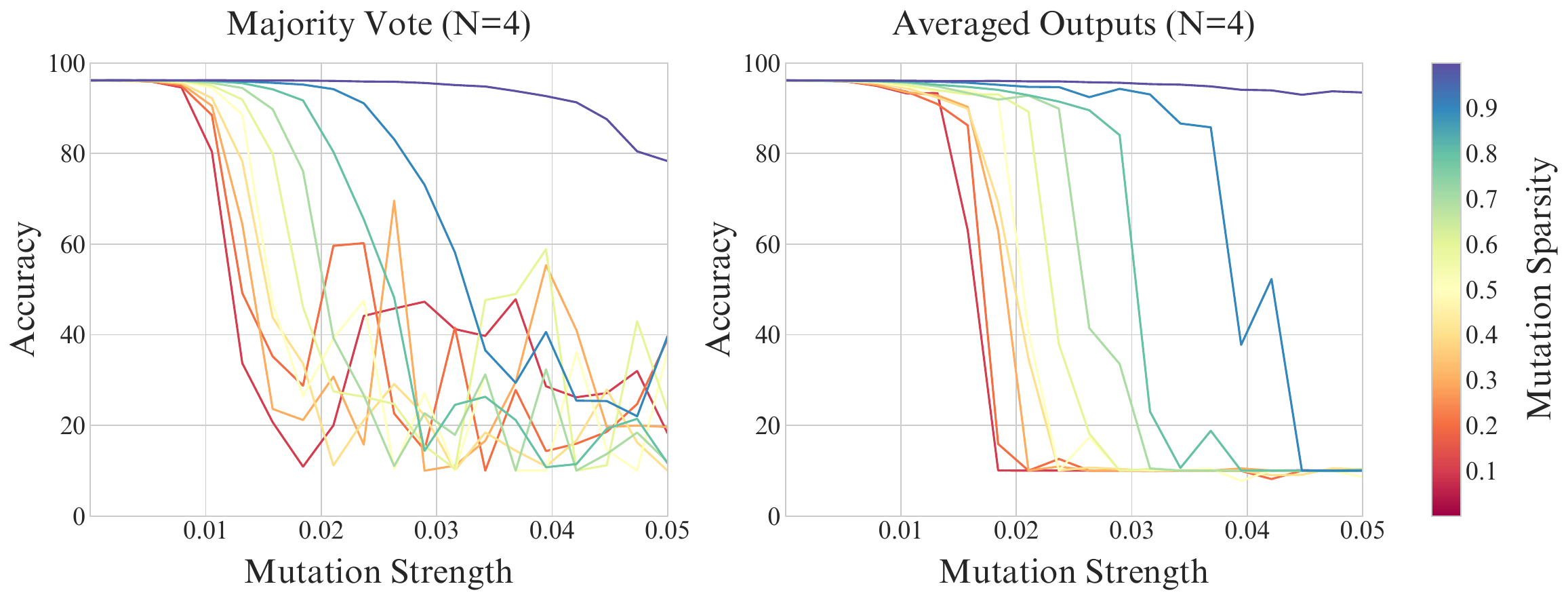}
    \caption{Results of mutation ablations for a trained WideResNet-28x10 model on CIFAR-10. Lines of different colors correspond to the percentage of sparsity for the mutations. The top left graph reports the average KL divergence as a result of mutation strength. The top right graph explores the relationship between KL divergence and accuracy. The bottom left reports the accuracy of an an ensemble evaluated with majority vote predictions. The bottom right  reports accuracy of an ensemble evaluated with averaged predictions.}
    \vspace{0.2in}
\end{figure}

The parent networks achieve an accuracy of approximately 96\% on CIFAR-10 and approximately 80\% on CIFAR-100.
Using these saved models as parent networks, we then perturb the models with varying amounts of mutation strengths, $\sigma \in (0.0, 0.05]$, and sparsities, $\rho \in [0.01, 0.99]$, and we evaluate their performance on the test sets.
Figure 4.3 displays the results of these experiments where we start by measuring the effect that mutations have on KL Divergence.
Predictably, we see that KL divergence quickly increases as the density and strength of mutations increase.
The rapid increase in KL divergence for dense perturbations illustrates how quickly performance collapses between very small changes in mutation strength.
We also observe an interesting linear relationship between KL divergence and model accuracy.

We then implement Noisy Subnetwork Ensembles where a population of 16 models are created by perturbing the parent model according to a given mutation strength and sparsity.
The top 4 models with the best accuracy on the validation set are then selected.
We report the accuracy on the test set with this ensemble when predictions are decided via majority vote and averaged output distributions.
The averaged predictions approach is much more consistent than majority vote, which display large amounts of variance as mutations become stronger.
The accuracy of both combination methods are comparable with no noticeable difference at low levels of mutation.
In settings with strong mutations, the averaged predictions model exhibits more robust behavior and better accuracy than majority vote.
The point where the mutation started to negatively impact member accuracy corresponds to an observed KL divergence approximately between 0.02 to 0.05.

\subsection{Benchmarks}

\subsubsection{ImageNet}

We then explore our approach on the difficult benchmark computer vision dataset, ImageNet \cite{krizhevsky2012imagenet, deng2009imagenet}. ImageNet is a large scale collection of images that have been hand labelled for use in machine learning tasks and is organized according to the wordnet dataset hierarchy. Over 14 million images and 20,000 labels have been collected in total. We use the 2012 ImageNet collection which consists of a training set of 1.2 million images and a validation set of 50,000 images, each belonging to one of 1000 categories. Images have varying sizes and aspect ratios and consist of both colored and grayscale photos. Images are normalized with mean standard normalization and we implement standard data augmentations which consists of resizing to 256x256 pixels and center cropping to 224x224 pixels.

We evaluate our approach with ten popular deep neural network architectures of varying sizes and generalization capacities in order to demonstrate the generality of this approach across different contexts \cite{krizhevsky2012learning, huang2018densely, szegedy2014going, howard2017mobilenets, he2015deep, xie2017aggregated, zhang2017shufflenet, iandola2016squeezenet, simonyan2015deep, zagoruyko2017wide}. All networks are pretrained and available from the torchvision repository \cite{pytorch2022torchvision}. 

We break the ImageNet competition test collection (ILSVRC 2012) of 50,000 images into a 80/20 split between test and and validation sets that now each contain 40,000 and 10,000 samples respectively.
All fitness evaluations and model selections use the validation set and all reported results are evaluated on the holdout test set.
We report accuracy, negative log likelihood, and expected calibration error for the parent network and the ensemble \cite{kumar2020verified}. We ran each model twice and report the best results.

\begin{table}[t]
    \caption{Results for sparse mutation ensembles on ImageNet with a wide variety of models. Mutation sparsity and strength are determined from a small hyperparameter grid search. Accuracy (Acc), Negative Log Likelihood (NLL), and expected calibration error (ECE) are reported for the parent and the ensemble. Metrics prepended with $e$ refer to ensemble results. $\Delta$Acc is the change in accuracy between the parent and the ensemble, $\sigma$ is the mutation strength, and $\overline{KL}$ is the average output divergence between the mutated models and the parent models. 16 models are generated and the 8 most accurate candidates on a validation set are used together as an ensemble.}
    \scriptsize
    \begin{tabularx}{\textwidth}{X l l l l l l c c c c}
    \toprule
    Model & Acc $\uparrow$ & NLL $\downarrow$ & ECE $\downarrow$ & eAcc $\uparrow$ & eNLL $\downarrow$ & eECE $\downarrow$ & $\Delta$Acc $\uparrow$ & Parameters & \multicolumn{1}{c}{$\sigma$} & $\overline{KL}$  \\
    \midrule
    AlexNet & 56.46 & 1.904 & 0.021 & 56.52 & 1.903 & 0.019 & 0.06 & 61.1M & 0.0025 & 0.046 \\
    \text{DenseNet-121} & 74.43 & 1.014 & 0.024 & 74.68 & 1.004 & 0.021 & 0.25 & 8.0M & 0.0012 & 0.040 \\
    \text{Inception-V3} & 69.57 & 1.819 & 0.184 & 69.98 & 1.681 & 0.169 & 0.41 & 27.2M & 0.0007 & 0.049 \\
    \text{MobileNet-V2} & 72.12 & 1.136 & 0.072 & 72.14 & 1.132 & 0.024 & 0.02 & 3.5M & 0.0010 & 0.045 \\
    \text{ResNet-18}  & 69.76 & 1.247 & 0.026 & 69.93 & 1.238 & 0.022 & 0.17 & 11.7M & 0.0020 & 0.046 \\
    \text{ResNext-50} & 77.64 & 0.945 & 0.065 & 77.72 & 0.929 & 0.059 & 0.08 & 25.0M & 0.0012 & 0.051 \\
    \text{ShuffleNet-V2} & 69.51 & 1.354 & 0.072 & 69.57 & 1.351 & 0.071 & 0.06 & 2.3M & 0.0020 & 0.052 \\
    SqueezeNet & 58.10 & 1.852 & 0.017 & 58.14 & 1.852 & 0.017 & 0.04 & 1.2M & 0.0020 & 0.026 \\
    \text{VGG-16} & 71.62 & 1.140 & 0.027 & 71.64 & 1.138 & 0.028 & 0.02 & 138.4M & 0.0025 & 0.048 \\
    \text{WideResNet-50} & 78.47 & 0.879 & 0.054 & 78.60 & 0.852 & 0.038 & 0.13 & 68.9M & 0.0015 & 0.049 \\
    \bottomrule
    \end{tabularx}
\end{table}

We conduct a hyperparameter grid search in order to find appropriate mutation strengths for each model.
Using a separate holdout dataset of 1000 samples we measure the average KL Divergence and accuracy while we ablate mutation strength $\sigma \in [0.001, 0.015]$ with a mutation sparsity of 0.5. We then choose both a mutation strength that maximizes accuracy while targeting a KL Divergence of approximately 0.02 which was found empirically to be a safe range for mutation where the the accuracy of candidate models is balanced with exploration.
We then construct a population of 16 models with neural partitioning and mirrored sampling.
We first generate a random noise mutation sampled from a Gaussian distribution modified with mutation strength. We construct a mirrored version of this random vector by flipping the signs of each value. We then use neural partitioning to generate two opposed unstructured subnetwork masks each with 50\% sparsity
Each child network is then evaluated on a validation set and the top 8 candidates with the best accuracy are selected.
We evaluate the top 8 candidates and combine their output predictions together to report the ensemble accuracy, negative log likelihood and expected calibration error.

Table 4.1 contains both the parent results and the ensemble results. We see consistent improvement in all metrics for each model with Noisy Subnetwork Ensembles. While the difference between generalization performance is very small in many cases, it is notable that convergence is stable and appears to be monotonic with optimal mutation hyperparameters. 
However, there's a question as to how much potential is able to be squeezed out of these fully trained networks without any further optimization to the child networks. 
There is no accepted method for determining the theoretical ceiling of generalization capacity for complex network architectures.
These small improvements may be considered significant in some contexts where an improvement of 0.5\% accuracy on a dataset of 50,000 images corresponds to 250 more correct predictions.

\subsubsection{Procgen}

\begin{figure}
    \includegraphics[width=\textwidth]{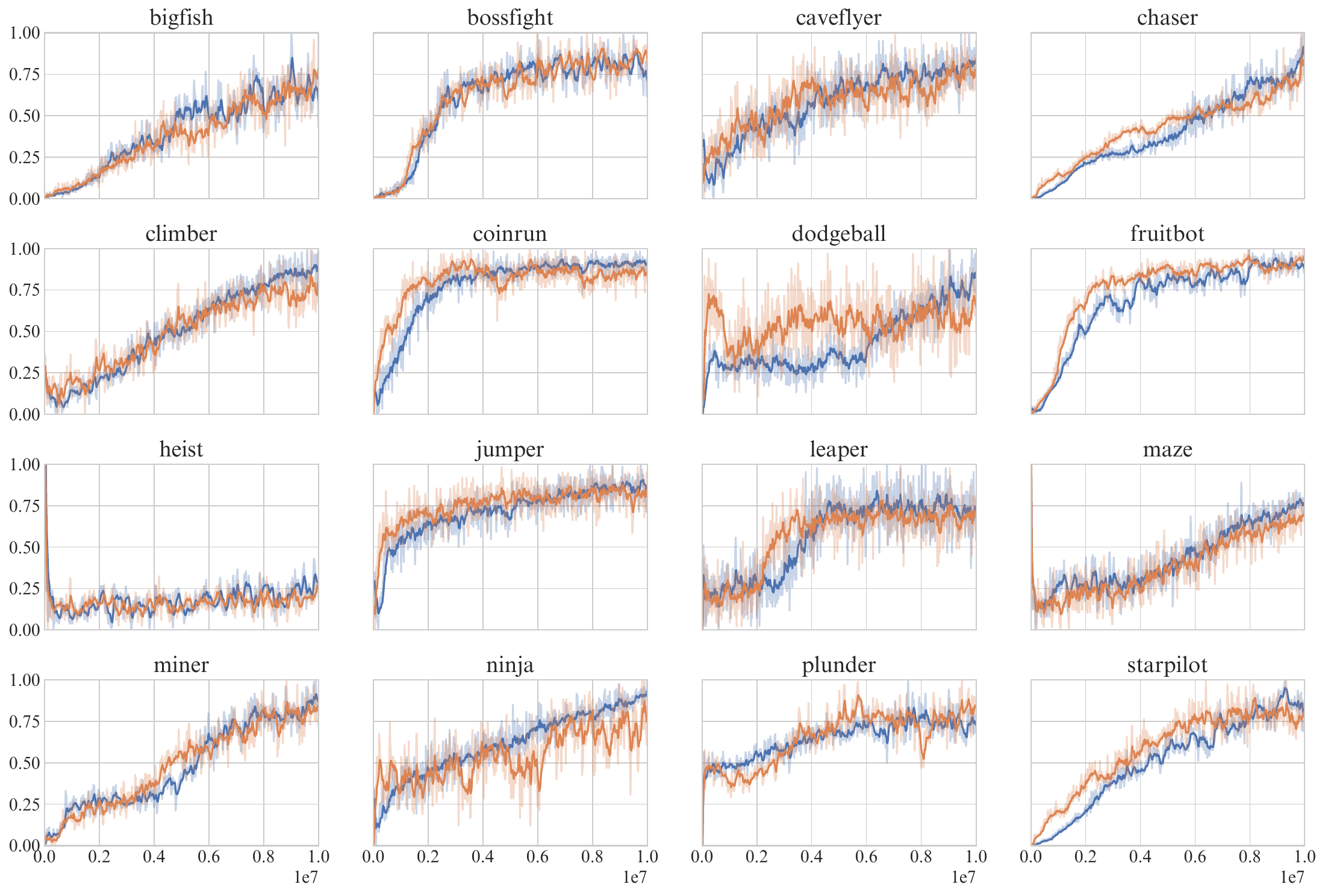}
    \caption{Training graphs on all 16 environments in the ProcGen benchmark. Orange lines correspond to single model evaluations and blue lines correspond to noisy subnetwork ensembles.}

    \vspace{0.1in}
    \captionof{table}{Proximal Policy Optimization hyperparameters and evaluation results over 10 episodes.}
    \vspace{0.1in}
    \footnotesize
    \begin{tabularx}{0.35\columnwidth}{X c}
    \toprule
    Parameter & Value \\
    \midrule
    $\gamma$ & 0.999 \\
    \addlinespace[0.01in]
    $\lambda$ & 0.95 \\
    Epochs/rollout & 3 \\
    \addlinespace[0.01in]
    Timesteps/rollout & 256 \\
    Minibatches & 8 \\
    \addlinespace[0.01in]
    Batch size & 2048 \\
    Entropy Bonus & 0.01 \\
    \addlinespace[0.01in]
    Target KL & 0.01 \\
    Clip Range & 0.2 \\
    \addlinespace[0.01in]
    Framestack & 2 \\
    Normalization & Reward \\
    \addlinespace[0.01in]
    Learning Rate & 5e-4 \\
    Workers & 1 \\
    \addlinespace[0.01in]
    Envs/worker & 64 \\
    Total Timesteps & 10m \\
    \addlinespace[0.01in]
    Network & Impala \\
    Block Config & $[32,64,128]$ \\
    \bottomrule
    \end{tabularx}
    \hspace{0.2in}
    \begin{tabularx}{0.6\columnwidth}{ l c c c } 
    \toprule
    Environment & Baseline & Noisy Subnetworks & $\sigma$ \\
    \midrule
    Bigfish & $13.7 \pm 5.47$ & $25.9 \pm 5.59$ & 0.06 \\ 
    Bossfight &  $9.1 \pm 1.93$ & $12.4 \pm 0.32$ & 0.05 \\
    Caveflyer & $3.9 \pm 1.53$ & $7.3 \pm 1.32$ & 0.07 \\
    Chaser & $2.3 \pm 0.22$ & $3.7 \pm 1.0$ & 0.01 \\
    Climber & $5.1 \pm 1.61$ & $7.0 \pm 1.75$ & 0.01 \\ 
    Coinrun & $8.0 \pm 1.26$ & $7.0 \pm 1.44$ & 0.06 \\
    Dodgeball & $0.8 \pm 0.51$ & $2.0 \pm 0.56$ & 0.05 \\
    Fruitbot & $30.2 \pm 3.16$ & $32.4 \pm 0.69$ & 0.04 \\
    Heist & $3.0 \pm 1.54$ & $4.0 \pm 1.55$ & 0.10 \\ 
    Jumper & $6.0 \pm 1.55$ & $7.0 \pm 1.45$ & 0.02 \\
    Leaper & $6.0 \pm 1.55$ & $9.0 \pm 0.94$ & 0.01 \\
    Maze & $6.0 \pm 1.54$ & $5.0 \pm 1.58$ & 0.04 \\
    Miner & $7.4 \pm 1.78$ & $4.7 \pm 1.74$ & 0.09 \\ 
    Ninja & $5.0 \pm 1.58$ & $7.0 \pm 1.45$ & 0.01 \\
    Plunder & $10.6 \pm 3.02$ & $13.6 \pm 2.97$ & 0.06 \\
    Starpilot & $20.1 \pm 5.42$ & $39.1 \pm 6.85$ & 0.01 \\
    \midrule
    Mean & $8.58 \pm 1.86$ & $11.69 \pm 2.74$ & $0.04 \pm 0.01$ \\
    \bottomrule
    \end{tabularx}
\end{figure}

We additionally explore the efficacy of Noisy Subnetwork Ensembles on the difficult Procgen reinforcement learning benchmark \cite{cobbe2020leveraging}.
Procgen consists of 16 unique game environments that use procedural generation to produce distinct and diverse levels for every run.
The observation space for all environments are 64x64x3 RGB images and the action space consists of 15 discrete options that correspond to behavior within the game (e.g. left, right, jump, shoot).
Procgen can generate endless numbers of unique levels which allows for a more robust evaluation of generalization, which has previously been challenging to evaluate in other reinforcement learning benchmarks \cite{zhang2018study}.

We use the Proximal Policy Optimization algorithm to train our models \cite{schulman2017proximal}. 
This is one of the most popular and reliable baseline algorithms in deep reinforcement learning.
Proximal Policy Optimization updates policies by taking the largest step possible to improve performance while satisfying a constraint on how close new and old policies can be. 
The motivation is to avoid a common failure point in reinforcement learning where too significant of a change in policy can result in catastrophic performance collapse.




We train an Impala Convolutional Neural Network for 10 million time steps on each environment \cite{espeholt2018impala}.
The Impala architecture is a benchmark residual convolutional network for reinforcement learning tasks that contains 15 layers and approximately 1.6 million parameters.
On evaluation rollouts, we create a Noisy Subnetwork Ensemble consisting of 8 models where a random selection of 20\% of the weights are perturbed by sampling from a Gaussian distribution with mutation strength scaled by $\sigma \in [0.01, 0.10]$. 
Perturbations are applied in a layerwise unstructured fashion.
Actions are chosen by averaging the softmax outputs of the ensemble members.



Figure 4.4 and Table 4.2 includes evaluation graphs and final results on each environment for both Noisy Subnetwork Ensembles as well as the single parent model. All results are reported as the mean and standard deviation over 10 runs.
Noisy Subnetwork Ensembles improved results in most environments with a few exceptions, potentially due to the relation between ensembles and probabilistic actions.
Proximal Policy Optimization is a stochastic algorithm that chooses actions according to a probability distribution. 
When combining the predictions of child networks, the output distribution tends to magnify the most likely action. 
This can be detrimental in environments where stochasticity is helpful for encouraging exploration as the deterministic choice may lead to situations where an agent gets stuck in a state where it is unable to recover.

\section{Discussion}

We introduce Noisy Subnetwork Ensembles as a formulation of Subnetwork Ensembles where child networks are generated by perturbing sampled subspaces of a parent network with random noise mutations.
This proves to be an effective approach to alleviating the challenges of mutating deep neural networks in evolution strategies.
Subnetwork perturbations widen the critical mutation window which allows for stronger mutations and more diversity before succumbing to performance collapse.
We explore how these subnetwork mutations reduce variance and we show how a model selection strategy can further improve performance in these ensembles.

We conduct several ablation studies in order to explore the interplay between sparsity and mutation strength on network behavior. We conduct a decision boundary experiment where visualizations are created that show the predictions of a model on a binary classification dataset. 
After being perturbed with dense mutations, the model sees a rapid decline in performance while sparsity significantly helps in maintaining accurate predictions. 
We then explore how these insights translate to a wide residual network on CIFAR where we explore parameter sensitivity and differences between averaged weights and averaged predictions for various levels of mutation strength and sparsity.
Our findings reaffirm the idea that subnetwork mutations produce accurate models more reliably than dense mutations.

We then introduce two large scale benchmark evaluations of Noisy Subnetwork Ensembles on ImageNet and Procgen, with a wide variety of deep neural network architectures.
We use a trained parent network to generate child networks with noise perturbation hyperparameters found through a short grid search algorithm.
We observe a small yet reliable improvement in generalization for every model on ImageNet and for most environments on Procgen.
This formulation of Noisy Subnetwork Ensembles offers an intriguing computational tradeoff  as child networks can be quickly generated to obtain small generalization improvements, without requiring any additional training.


\chapter{Sparse Subnetwork Ensembles}
\label{chap:sparse}
\section{Introduction}

\begin{figure}
    \centering
    \includegraphics[width=\textwidth]{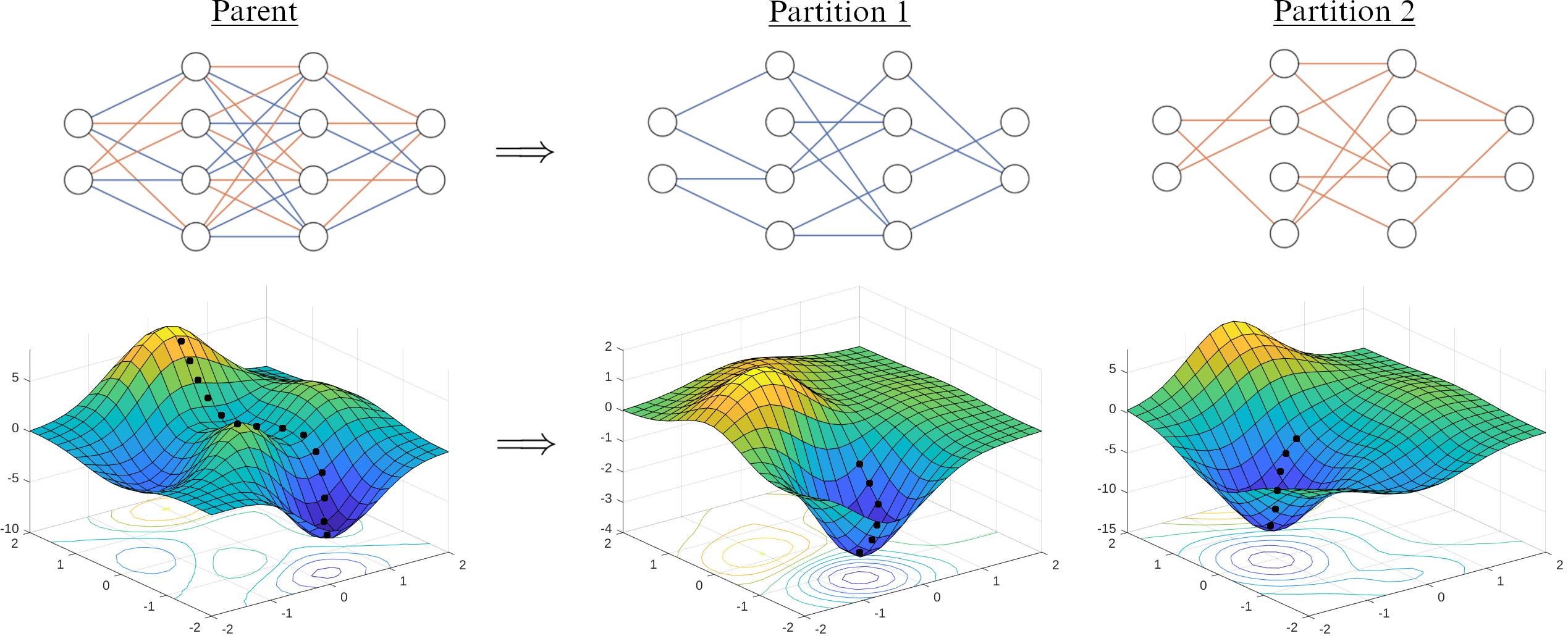}
    \caption{A diagram explaining network partitions in Sparse Subnetwork Ensembles. The parent network is optimized and settles into some optima. Child networks are created via neural partitioning. That is, two sparse children each inherit a different set of parameters from the parent. The optimization landscapes for each child are diverse, yet they converge quickly due to inheritance from a fully trained parent.}
    \label{fig:antirandom-split}
\end{figure}

Noisy Subnetwork Ensembles were introduced as a no-cost formulation of Subnetwork Ensembles where child networks are generated through subnetwork perturbations and ensembled together without applying any additional training. 
The combination of these perturbed models was shown to reliably improve generalization performance on holdout data, however the differences were relatively small. 
Without further optimization of the child networks, there is a limited potential for improving the predictive capacity of each child network. 
Additionally, continued training of the child networks in Noisy Subnetwork Ensembles would naturally lead to convergence towards highly correlated areas of the optimization landscape as they share the vast majority of their parameters and network structure.
This lack of diversity can limit performance and lead to less robust ensembles.

We extend this work by introducing \textit{Sparse Subnetwork Ensembles}, a formulation of Subnetwork Ensembles where child networks are created through the pruning of subnetworks in a trained parent network. 
The sparse child networks are then subsequently trained for a small number of epochs in order to recover the lost accuracy from pruning.
The unique network topology of each child encourages convergence to unique and diverse minima, resulting in more robust ensembles.
We also explore several methods for improving diversity among child networks, including Neural Partitioning for generating children with mirrored network structures, and a cyclic learning rate schedule that alternates between large and small values to promote more exploration in weight space before converging \cite{smith2018superconvergence}.

This approach is highly flexible and offers several unique benefits over other low-cost ensemble methods. 
Like Noisy Subnetwork Ensembles, additional child networks can be dynamically generated with little additional computation. 
Generating ensemble members through pruning results in much smaller networks that significantly reduce memory usage and computational cost during forward passes.
This approach attempts to strike the optimal balance between leveraging the improved training dynamics associated with using a large parent network, while also creating accurate and diverse child networks that converge quickly to good optima.

We compare Sparse Subnetwork Ensembles to several popular state-of-the-art ensemble learning algorithms with benchmark deep neural network architectures and datasets. We conduct hyperparameter ablation studies with child sparsity, pruning structures, tuning schedules, and ensemble sizes of up to 128 members. We demonstrate that Sparse Subnetwork Ensembles are both highly efficient, accurate, and robust with both small and large training budgets.

\section{Implementation}



Sparse Subnetwork Ensembles begin with the standard training of a large parent network $F$, parameterized with weights $\theta$.
Given a fixed training budget, it can be difficult to determine the appropriate amount of time to allot for training the parent as model architecture, dataset complexity, total training budget, ensemble size, and inference compute limitations can all have a strong impact on performance for different ensemble configurations.
As a general guideline, our large scale image classification experiments with Wide Resnets suggest a 70/30 split between parent/children training time for a budget of 200 epochs.

We then spawn $k$ child networks by copying the parent and pruning a subset of the weights. 
The binary subnetwork mask $M \in \{0, 1\}$ is generated for each layer with the same dimensions as the layer weights $W$. 
The values for each parameter in the mask is determined by sampling a Bernoulli distribution with probability $\rho$, where a successful trial denotes that the parameter will be retained.
We also explore an implementation where the mask is determined by sampling such that neurons are masked rather than individual connections.
The sparse subnetwork is created by taking the Hadamard product $\circ$ of the mask $M$ with the weights $W$ at each layer.

We additionally implement Neural Partitioning when constructing child networks of 50\% sparsity. In this case, whenever we generate a child network, we construct the anti-random subnetwork by flipping the polarity of the bits in the subnetwork mask of the original child. This can be repeated $n$ times to create an ensemble of size $2n$. The weights for each child network $\hat{W}$ and it's anti-random complement $\hat{W'}$ can be described with.
\begin{align}
M &\sim Bernoulli(\rho) \\
\hat{W} &= W \circ M \\
\hat{W'} &= W \circ (1 - M)
\end{align}

This works best with a child sparsity target of 50\% such that the size of all the child networks in the ensemble are roughly equivalent. Given a large enough parent network, the children can be partitioned to greater degrees (e.g. three child networks that inherit 33\% of the parent's parameters each). However, children with higher sparsity may require more training time to achieve similar performance.

Given the high dimensionality of deep neural networks, the probability that any two child networks will end up with similar topologies is small since the number of unique graph structures in random and complex graphs grow exponentially with the number of nodes \cite{harary2014graphical}.
The unique network topology of each child encourages optimization of diverse feature representations which helps to produce more robust ensembles.
Figure 5.2 illustrates how pruning convolutional filters applies unique geometric constraints which inherently forces the production of diverse feature maps.

\begin{figure}
    \centering
    \includegraphics[width=\textwidth]{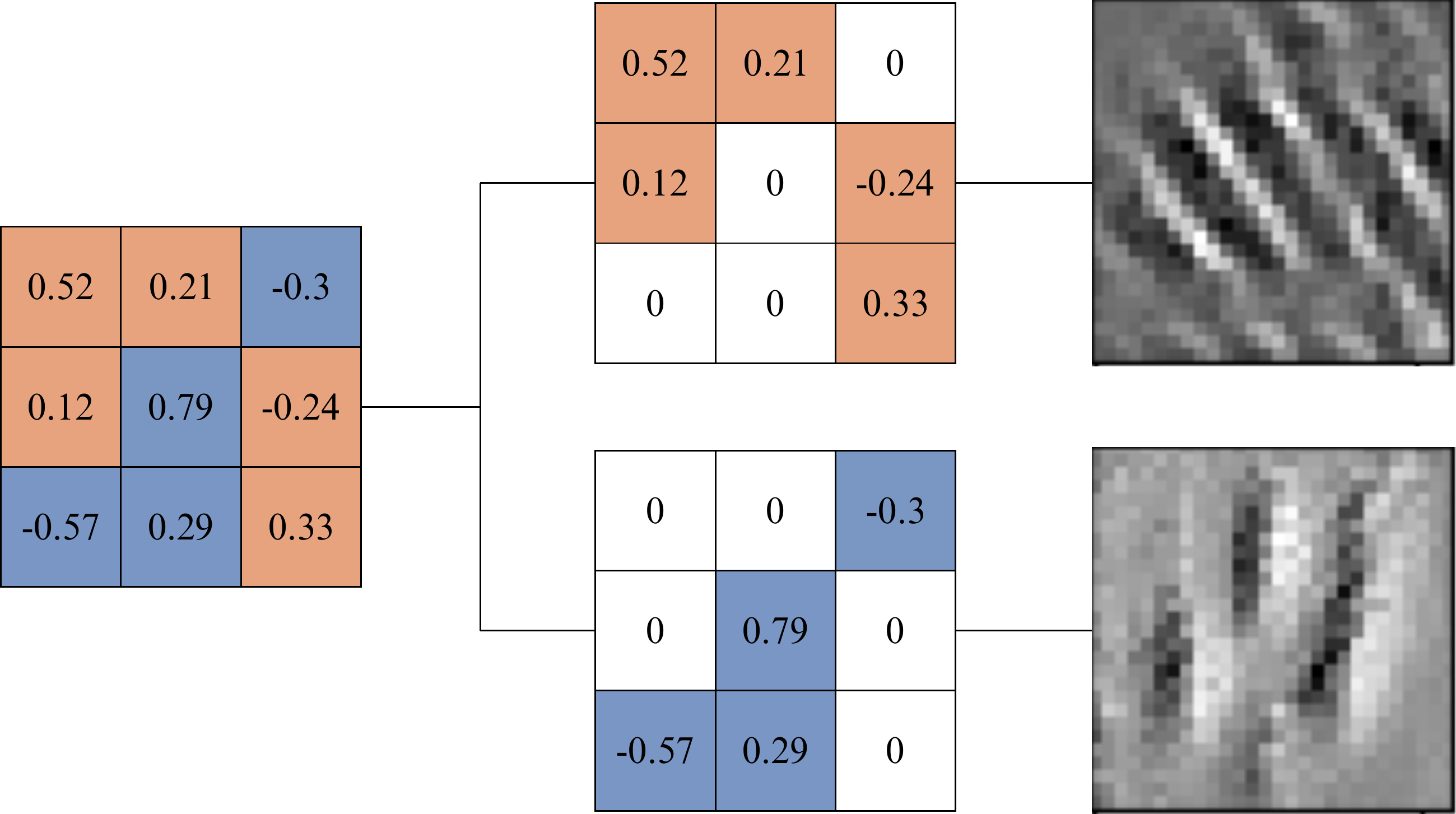}
    \caption{A convolutional parent network filter split into two partitioned child filters. The child networks learn diverse feature representations due to the geometric constraints of their unique topologies.}
    \label{fig:my_label}
\end{figure}

\begin{algorithm}
\SetAlgoLined
\KwIn{N, the number of ensemble members}
\KwIn{($X_{train}, X_{test}$), \text{the training and test data}}
\KwIn{($t_{parent}, t_{child}$), \text{the number of parent and child epochs to train for}}
\KwIn{($\eta_{parent}, \eta_{child}$), \text{the parent and child optimizer hyperparameters}}
\KwIn{$\phi(\rho)$, \text{the subnetwork sampling process with target sparsity $\rho$}}

\texttt{\\}
$O = sgd(\eta_{parent})$ \\
$F = initialize()$ \\
$F.w = train(F, O, X_{train}, t_{parent})$ \\
\texttt{\\}
$ensemble = []$ \\
\texttt{\\}
\For{i in 1 to $N$}{
    $f_i = initialize()$ \\
    $f_i.w = F.w$ \\
    \texttt{\\}
    \For{$j$ in $f_i.layers$}{
        $M \sim \phi(\rho)^{f_{i,j}.w}$ \\
        $f_{i,j}.w \leftarrow f_{i,j}.w \circ M$ \\
    }
    \texttt{\\}
    $O = sgd(\eta_{child})$ \\
    $f_i.w = train(f_i, O, X_{train}, t_{child})$ \\
    \texttt{\\}
    $ensemble \leftarrow f_i$ \\
}
\texttt{\\}
\For{$(x, y)$ in $X_{test}$}{
    $outputs = [F(x)$] \\
    \texttt{\\}
    \For{c in ensemble}{
        $outputs \leftarrow c(x)$ \\
    }
    \texttt{\\}
    $predictions = softmax(mean(outputs))$ \\
    $loss = nll(predictions, y)$ \\
    $accuracy = acc(predictions, y)$ \\
}
\caption{Sparse Subnetwork Ensembles}
\end{algorithm}


Each child network is then tuned for a small number of epochs. 
Since each child inherits its parameters from an accurate parent, the convergence to good solutions often occurs after a relatively small number of epochs \cite{blalock2020state}.
There are several tunable hyperparameters related to optimization that influence the performance of the fine tuning phase.
These include the number of epochs, learning rate schedule, batch size, weight decay, and data augmentation strategies.
These optimization hyperparameters are generally kept consistent with those used for training the parent network \cite{renda2020comparing}.
Parent networks are typically trained with a learning rate schedule that starts with a large value and slowly decays to a small value over time.
The most common method for tuning subnetworks in pruning literature is to use a small constant learning rate that is consistent with the final phase of training the parent.
However, recent work has found that mimicking the decaying schedules used for the parent network can be highly effective for tuning subnetworks \cite{renda2020comparing, le2021network}.

We experiment with a learning rate schedule called the one-cycle policy, which has been shown to be highly effective for rapidly training deep neural networks from scratch \cite{smith2018superconvergence}.
The one-cycle policy consists of a warm-up, cool-down, and annihilation phase, which alters the learning rate during training for each batch of data. 
The warm up phase begins with a small learning rate that ramps up to a large maximum rate. 
The cool-down phase gradually decays the maximum rate to a minimum rate that is lower than the initial learning rate. 
The annihilation phase takes place in the very final stages of training where the learning rate is decayed to 0.

The one-cycle policy is highly effective at encouraging diversity among child networks as the large learning rates enable each child to move greater distances away from the initial weights before converging to optima. This idea has been shown to be effective in other successful low-cost ensemble methods and is the primary insight underlying several popular temporal ensemble methods such as Snapshot Ensembles and Fast Geometric Ensembles \cite{huang2017snapshot, garipov2018loss}.

The original introduction of the one-cycle policy utilized linear annealing to modify the learning rates. However, cosine annealing tends to be more popular in subsequent implementations. 
In our experiments we use a revised one-cycle schedule that disregards the annihilation phase and uses only the warm-up and cool-down. 
Additionally, we use the inverse of the one-cycle policy for momentum when tuning with stochastic gradient descent \cite{sutskever13nesterov}. 
Both of these modifications have been shown empirically in unpublished work to perform better than the original implementations \cite{fastai2017onecycle}.

The value for the learning rate $\eta$ at iteration $t$, where $\eta_{init}$ is the initial learning rate value, $\eta_{max}$ is the maximium value, $\eta_{min}$ is the minimum value and $T$ is the total number of iterations can be described as:
\begin{align}
\eta_{warm}(t) &= \eta_{init} + \frac{1}{2}(\eta_{max} - \eta_{init}) \left( 1 - cos \left( \frac{\pi t}{T} \right) \right) \\
\eta_{cool}(t) &= \eta_{min} + \frac{1}{2}(\eta_{max} - \eta_{min}) \left( 1 + cos \left( \frac{\pi t}{T} \right) \right)
\end{align}

\begin{figure}
    \centering
    \includegraphics[width=0.45\textwidth]{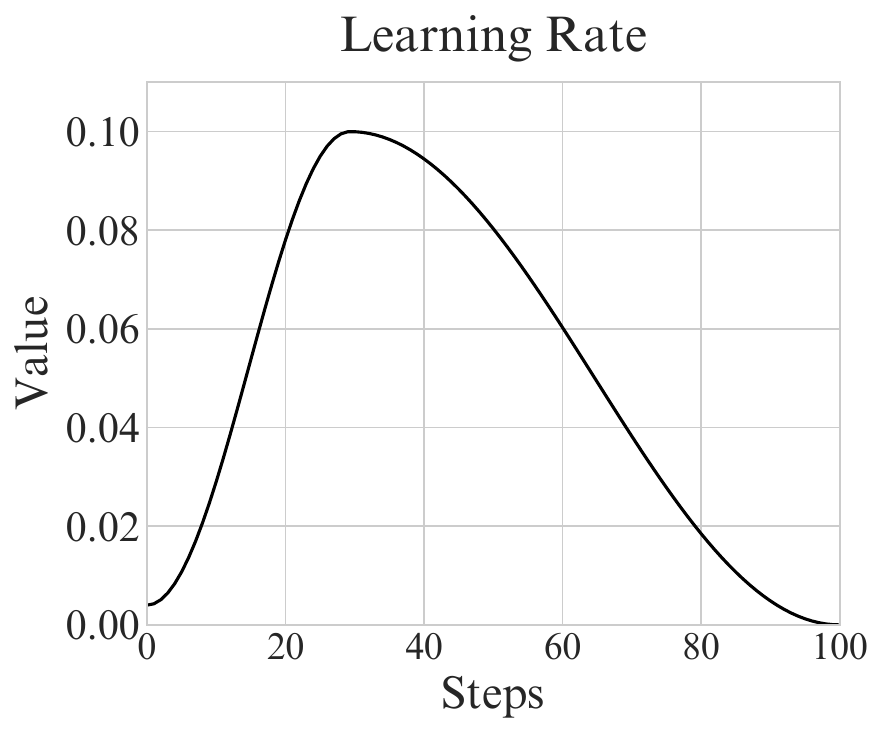}
    \hspace{0.2in}
    \includegraphics[width=0.45\textwidth]{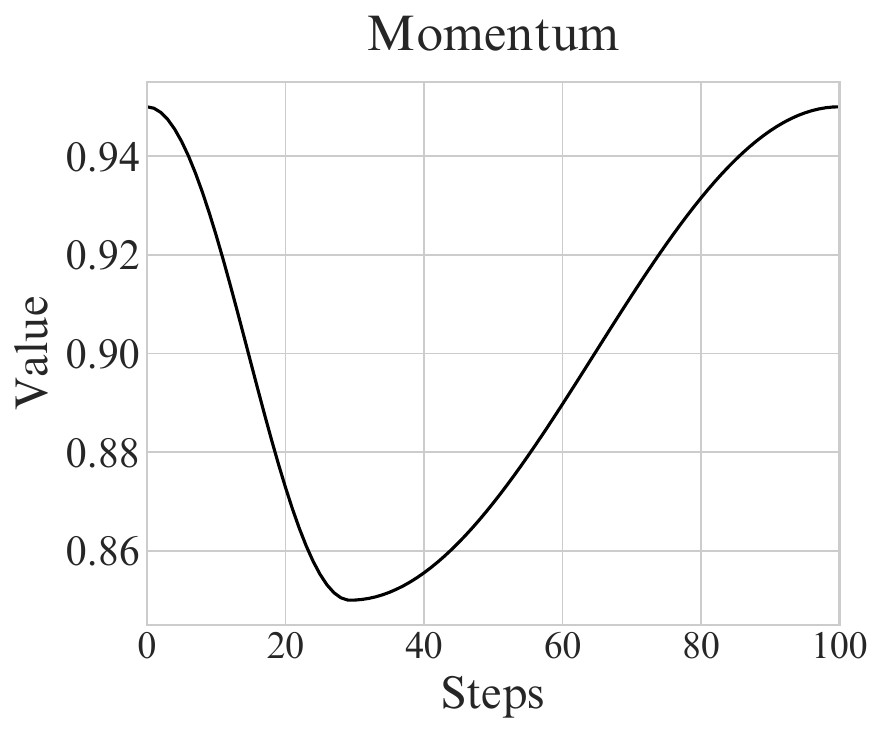}
    \caption{An example of the one-cycle learning rate and momentum schedule we use to fine-tune the child networks in Sparse Subnetwork Ensembles.}
    \label{fig:my_label}
\end{figure}

\section{Experiments}




\subsection{Ablations}

We explore the impact of several hyperparameters that govern how sparse Subnetwork Ensembles are created and optimized.
We use three different variations of the LeNet-5 model architecture where we vary the number of neurons in each hidden layer to produce a small, medium and large model that we call \textit{LeNet-S} (253,290 parameters), \textit{LeNet-M} (1,007,306 parameters), and \textit{LeNet-L} (4,017,546 parameters).
All ablation experiments use the Adam optimizer with a learning rate of $0.001$ and the benchmark CIFAR-10 and CIFAR-100 datasets, which consist of 60,000 small natural colored images of 32x32 pixels in size. Those 60,000 images are split up into 50,000 training images and 10,000 testing images. CIFAR-10 samples from 10 classes of images, while CIFAR-100 samples from 100 classes of images. 

Our first ablation aims to evaluate the effect that the amount of sparsity in the child networks has on both the size of the parent model and the resulting ensemble performance.
We train each parent model (\textit{LeNet-S}, \textit{LeNet-M}, and \textit{LeNet-L}) for 10 epochs.
An ensemble of 8 sparse subnetwork children are created for each parent by pruning to ablated target sparsity values $s \in [0.1, 0.9]$, and subsequently tuned for an additional 3 epochs.
Larger models are shown to have more capacity for pruning and learning more efficiently. Ensemble accuracy improves up to a critical region as child networks become more sparse without needing any additional epochs of training. More sparsity in the child networks can yield greater ensemble performance which may suggest better diversity.

Next we explore how structured (neuron-level) and unstructured (connection-level) pruning methods affect ensemble performance at different levels of child network sparsity.
We first train the \textit{LeNet-L} parent network for 10 epochs.
An ensemble of 8 children is created where each is pruned using either structured or unstructured pruning to ablated target sparsity levels $s \in [0.1, 0.9]$, and tuned for an additional 3 epochs.
Connection pruning performs much better at higher levels of sparsity compared to neuron pruning. 
However, we see little difference in performance for child network sparsities up to 50\%.

\begin{figure}
    \includegraphics[width=.46\textwidth]{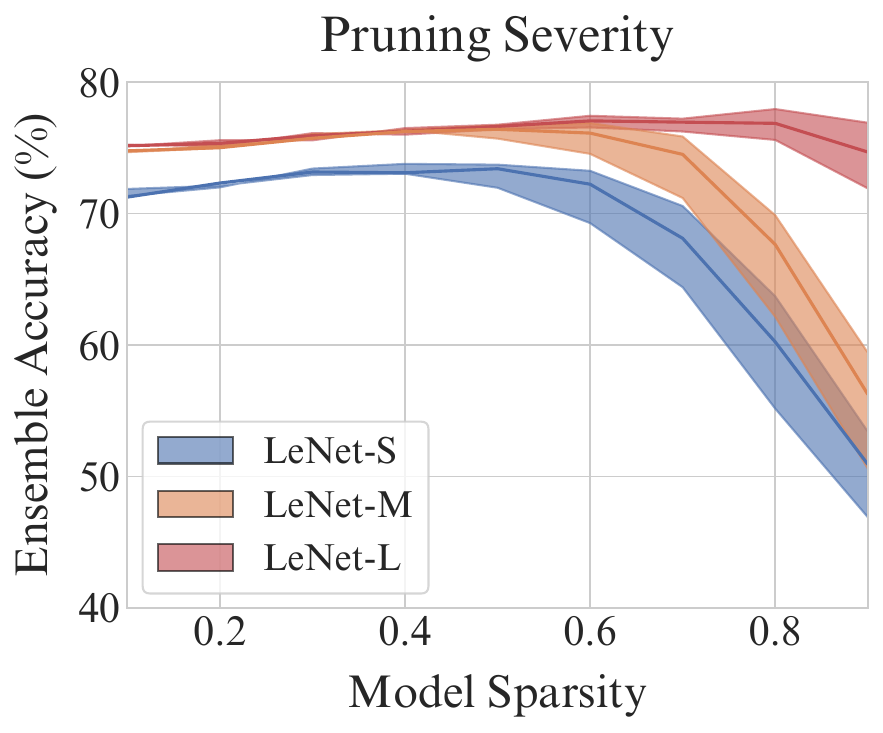}
    \hspace{0.4in}
    \includegraphics[width=.46\textwidth]{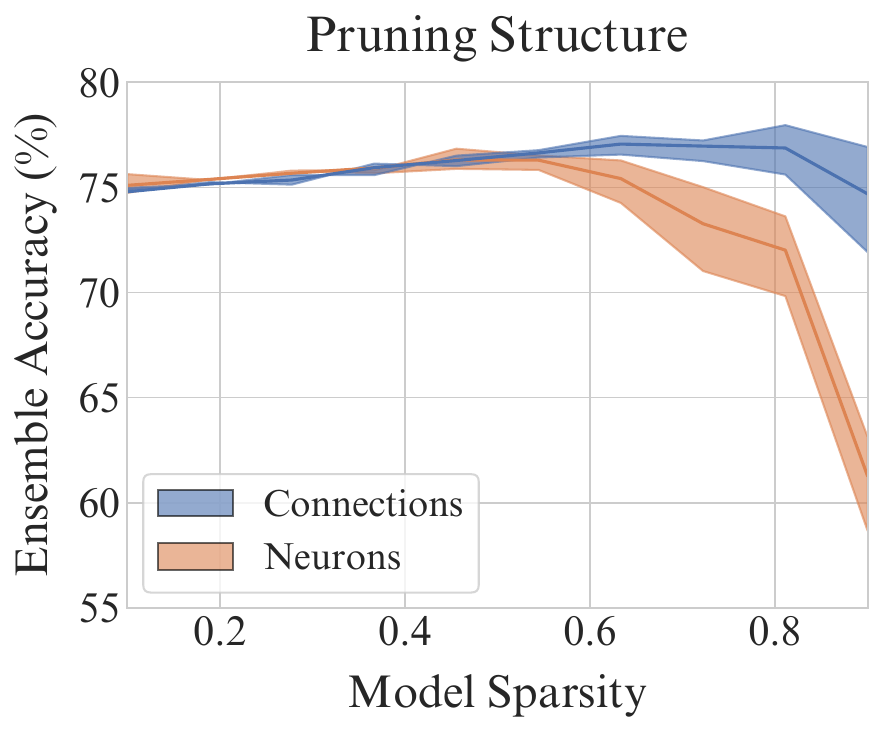}
    \caption{Ablations over pruning sparsity and structure for various model sizes. Each graph reports performance for an ensemble containing 8 models. The left figure reports the effect that pruning sparsity has on models of different sizes. The right figure displays the efficacy of neuron and connection pruning for LeNet-L.}
\end{figure}

One of the unique advantages of Subnetwork Ensembles is that it is relatively cheap to dynamically generate additional ensemble members.
We explore this by taking the \textit{LeNet-M} network and producing up to 128 candidate member networks.
We start by training the parent network for 10 epochs.
Child networks are created through unstructured pruning with a sparsity target of $0.5$. 
Each child is tuned for one additional epoch. 
We evaluate ensemble performance over 10 runs as we increase the size of the ensemble from 4 to 128.
Larger ensembles offer great potential for improved generalization as we see monotonically improved accuracy with reduced variance as the number of ensemble members increases.

We then aim to evaluate the efficacy of neural partitioning and the one-cycle learning rate policy.
We use a more modern training procedure for this experiment where the parent network (\textit{LeNet-L}) is optimized by stochastic gradient descent with Nesterov momentum for 20 epochs. 
We use an initial learning rate of $\eta_1 = 0.1$ for 50\% of the training budget which decays linearly to $\eta_2 = 0.001$ at 90\% of the training budget. The learning rate is kept constant at $\eta_2 = 0.001$ for the final 10\% of training. 
Child networks are created with a target sparsity of $0.5$ using either random pruning (R) or Neural Partitioning (NP).
Child networks are tuned with either a constant fine tuning (FT) learning rate of $\eta=0.01$ or with a One-Cycle (1C) schedule that ramps up from $\eta_1=0.001$ to $\eta_2=0.1$ at 10\% of tuning, then decaying to $\eta_3=1e-7$ at the end of 5 epochs.
We record results for 10 sample runs of each configuration (R+FT, NP+FT, R+1C, NP+1C), where we observe the best results with the combination of neural partitioning and one-cycle tuning on both CIFAR-10 and CIFAR-100.

\begin{figure}
    \includegraphics[width=.46\textwidth]{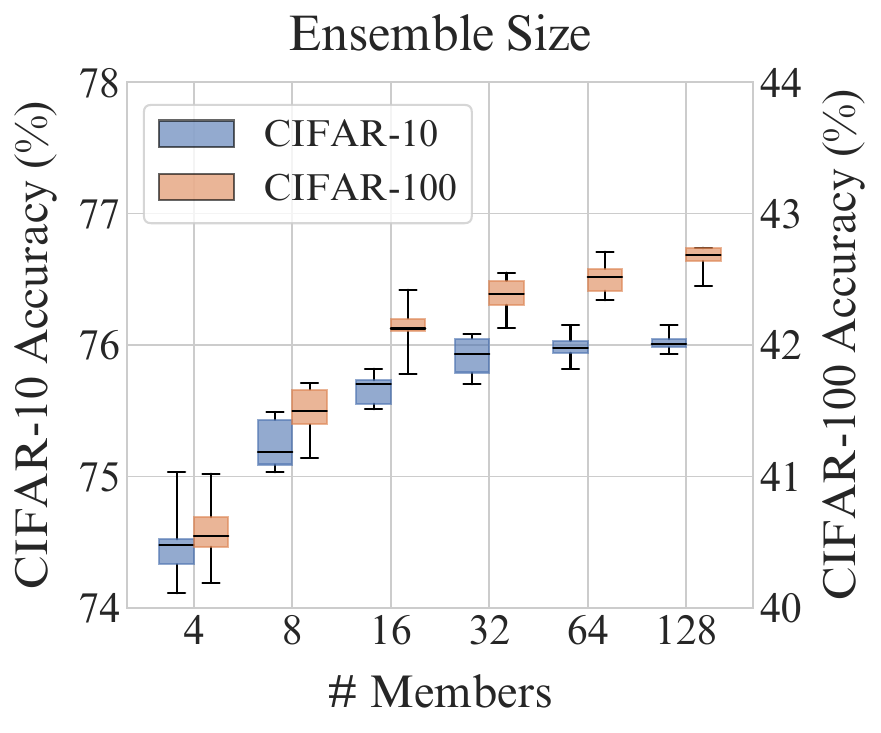}
    \hspace{0.4in}
    \includegraphics[width=.46\textwidth]{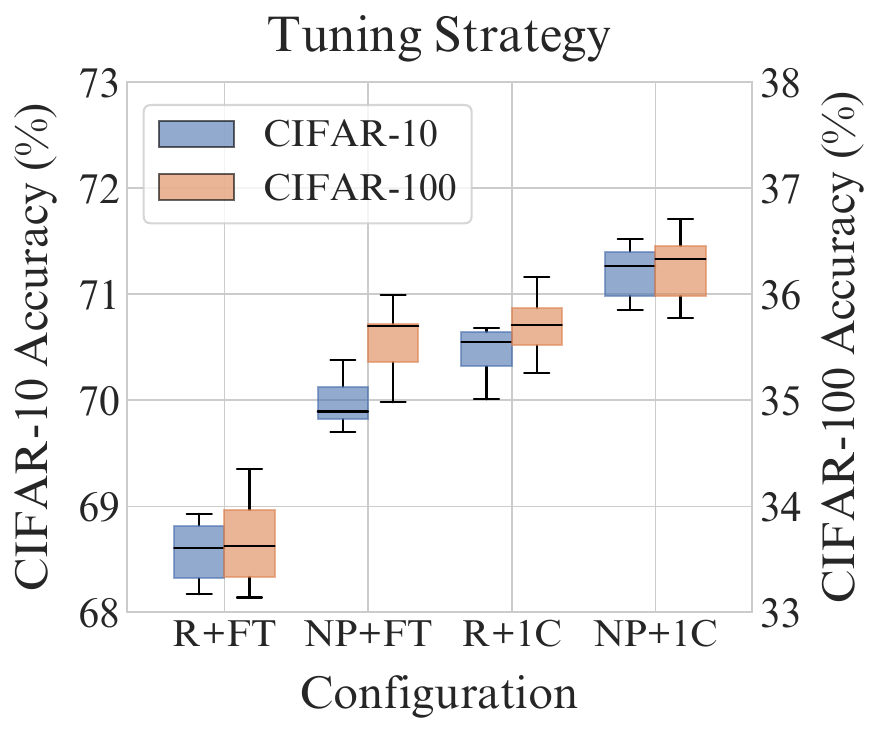}
    \caption{Ablations exploring the effect of ensemble size and various ensemble pruning/tuning strategies. The left graph displays the results for ever increasing ensemble sizes spawned from a single parent network. The right figure contains results for various configurations of random pruning and neural partitioning with constant rate fine tuning and one-cycle fine tuning.}
\end{figure}

We next explore the early training efficiency of Sparse Subnetwork Ensembles compared to other low-cost methods, MotherNets and Dropout, which are two conceptually competitive approaches \cite{wasay2020mothernets, srivastava2014dropout}.
Like Subnetwork Ensembles, MotherNets creates ensembles of child networks from a trained parent network.
However, this approach instead uses a small parent network to create larger child networks by adding new neurons and layers around the small shared structure.
Meanwhile, Dropout is a popular regularization technique that masks random neurons in the network during training such that a unique subnetwork is active for each forward pass.
The purpose of these comparisons is to explore the idea that we get better training dynamics by first training a large parent network and then decomposing it into small networks, rather than using smaller networks throughout training.

\begin{figure}
    \includegraphics[width=.46\columnwidth]{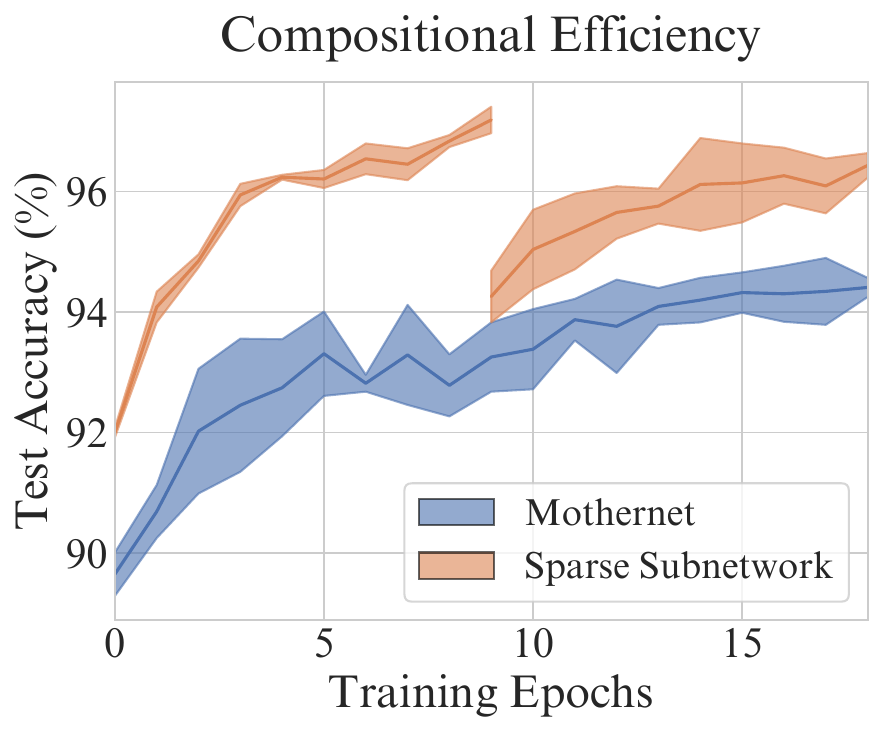}
    \hspace{0.4in}
    \includegraphics[width=.46\columnwidth]{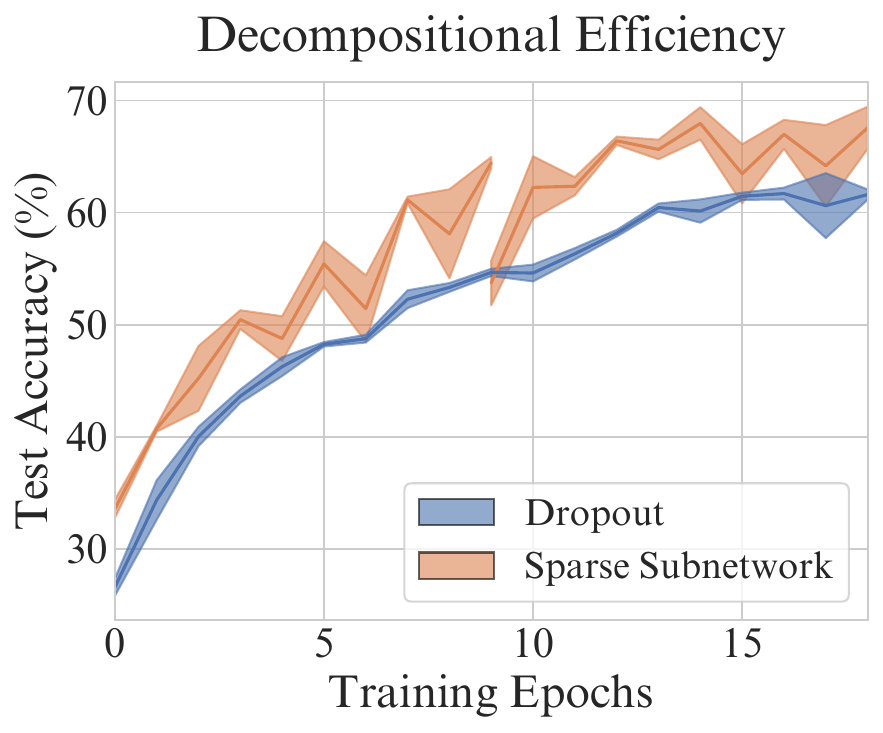}
    \caption{Exploring the early training efficiency of Sparse Subnetwork Ensembles with Mothernets and Dropout. The left figure compares Mothernets on the MNIST dataset with a multilayer perceptron. The right figure compares Dropout on CIFAR-10 with LeNet-L.}
\end{figure}

The comparison to MotherNets uses the small benchmark dataset MNIST, which is a collection of small handwritten black and white digits \cite{lecun1998mnist}. We start by training a MotherNet that contains one hidden layer of 16 neurons. A child network is then created where an additional 16 neurons are added to the original 16 neuron layer of the trained MotherNet. We compare this with a Sparse Subnetwork Ensemble parent that contains a single hidden layer of 64 neurons. A child network is then created by pruning the parent's hidden layer to 32 neurons. We train both parents and the resulting children for 10 epochs each using stochastic gradient descent with momentum. The resulting children for both methods are the same size and both approaches train for 20 total epochs.
The training graphs of these methods shows that Sparse Subnetwork Ensembles reach higher accuracies than MotherNets in fewer epochs.
There is a smooth and consistent training curve with the hatched children of MotherNets, while Subnetwork children undergo a large drop in accuracy before quickly recovering and exceeding the performance of the MotherNet children.

The comparison to Dropout is done on CIFAR-10 with the \textit{LeNet-L} architecture.
The Dropout model is identical to the parent network used for Sparse Subnetwork Ensembles, except that it contains dropout layers after every hidden layer with an activation probability of $0.5$ during training.
We train the parent network for 10 epochs, after which a sparse child network is created by pruning 50\% of the neurons and training for an additional 10 epochs.
Dropout does offer a smoother training trajectory with reduced variance, however Sparse Subnetwork Ensembles appear to train more efficiently with better accuracy after 20 epochs, despite the final child network containing half the number of neurons of the Dropout model.

\subsection{Landscapes}

Deep neural networks are known to be vulnerable to adversarial attacks where small manipulations, imperceptible to humans, are applied to input images in order to mislead the network \cite{szegedy2014intriguing}.
Previous works have theorized that this is potentially due to excessive overparameterization which causes spurious correlations to develop that are exploitable.
Recent research has demonstrated that pruning can be an effective tool for improving robustness against adversarial attacks \cite{li2022pruning, jordao2021effect, ye2021adversarial}.
It has also been observed that there is a correlation between robust solutions and flatness of the local optimization landscape \cite{stutz2021relating}.
We explore how the loss landscape of a sparse subnetwork evolves as it undergoes both pruning and tuning.

We first train a parent ResNet-56 model for 300 epochs using Stochastic Gradient Descent with momentum on CIFAR-10 with a batch size of 128. The learning rate starts at $0.1$ and decays to $0.01$ at 150 epochs, $0.001$ at 225 epochs and $0.0001$ at 275 epochs. We then prune the network by eliminating a random 50\% of the weights and continue training the sparse child network for three additional epochs with a constant learning rate $0.001$.

The visualizations are created by choosing a center point for the graph with the model weights $\theta$. Two orthogonal directional vectors $\delta$ and $\eta$ are chosen and the loss landscape is plotted by evaluating the network with weights that are determined by linearly interpolating over the two directional vectors \cite{li2018visualizing}.
\begin{equation}
f(\alpha, \beta) = \mathcal{L}(\theta + \alpha \delta + \beta \eta)
\end{equation}

\begin{figure}[t]
    \includegraphics[width=\textwidth]{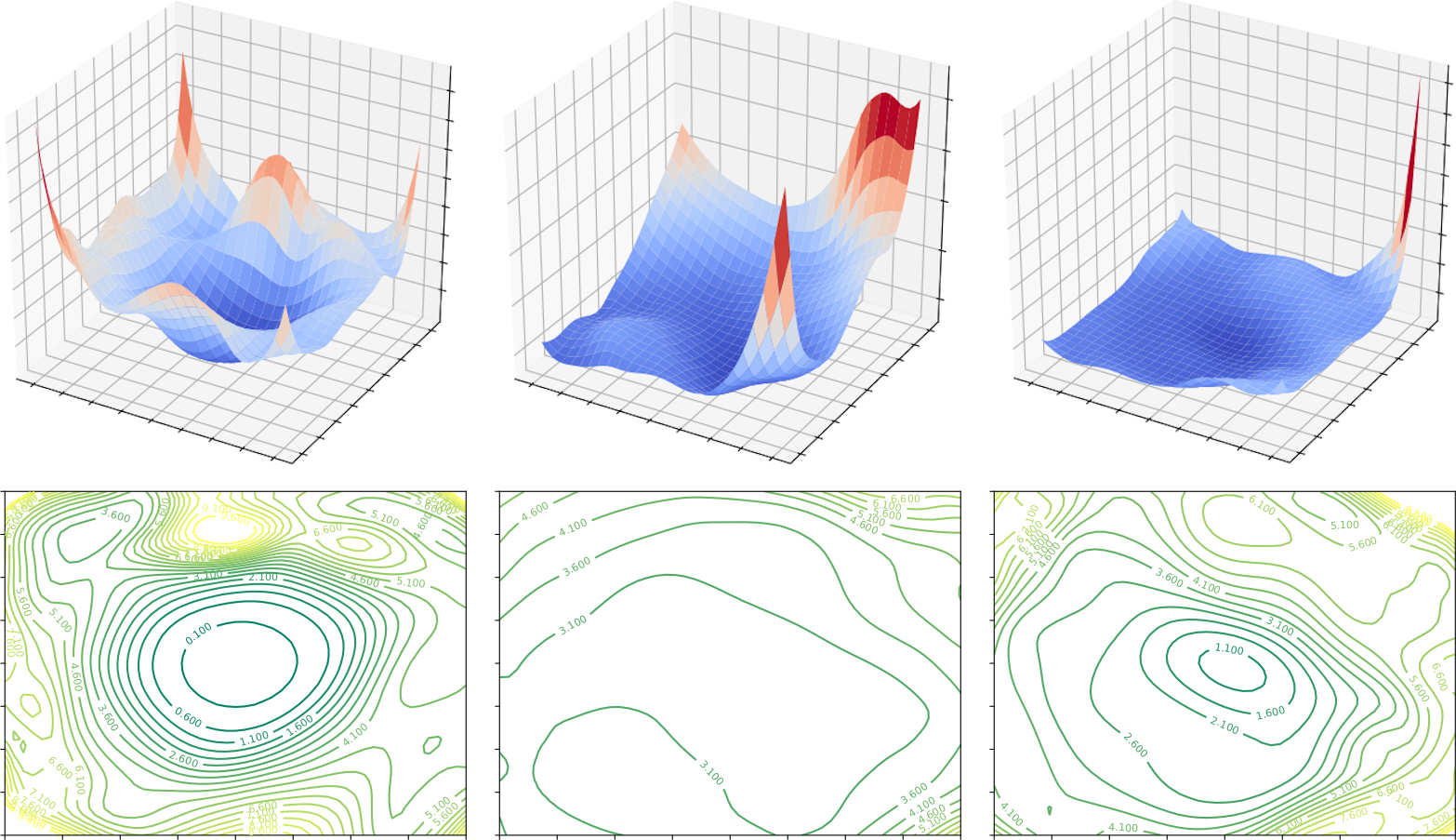}
    \caption{Visualizations of the loss landscapes of a ResNet-56 model as it undergoes pruning and tuning. The first column is the landscape of the fully trained parent model. The middle figure is the landscape of the model after removing 50\% of the parameters. The final figure is the landscape of the pruned model after three additional epochs of fine tuning.}
\end{figure}

Figure 5.7 displays the results of this visualization. We note the apparent simplification of the loss landscape of the sparse child network as it is tuned. The model appears to settle into an area of the landscape that is shallow and wide, suggesting a good location for robust generalization.

\subsection{Benchmarks}

\subsubsection{Small Budget}

\begin{table}
\captionof{table}{Accuracy of various methods for the small training budget benchmark experiment. All ensembles contain 8 models and are trained for a total of 16 epochs. Dropout rate is 50\%, Bootstrap Aggregation is trained on 90\% subsets.}
\small
\begin{tabularx}{\columnwidth}{l X c c}
\toprule
Model & Method & CIFAR-10 & CIFAR-100 \\
\midrule
LeNet-L  & Independent & 70.29 $\pm$ 0.21 & 34.39 $\pm$ 0.18 \\
        & Dropout & 70.06 $\pm$ 0.32  & 28.98 $\pm$ 0.33 \\
        & Bootstrap Aggregation & 70.97 $\pm$ 0.13  & 32.40 $\pm$ 0.11 \\
        & Sparse Subnetwork Ensemble (R) & 75.51 $\pm$ 0.06  & 41.68 $\pm$ 0.14 \\
        & Sparse Subnetwork Ensemble (NP) & \textbf{75.75} $\pm$ \textbf{0.06} & \textbf{41.85} $\pm$ \textbf{0.13} \\
\midrule
ResNet-18 & Independent & 80.99 $\pm$ 0.11 & 49.26 $\pm$ 0.09 \\
        & Dropout & 72.97 $\pm$ 0.41 & 37.69 $\pm$ 0.28 \\
        & Bootstrap Aggregation & 79.88 $\pm$ 0.05 & 46.18 $\pm$ 0.06 \\
        & Sparse Subnetwork Ensemble (R) & 84.61 $\pm$ 0.07  & 57.08 $\pm$ 0.07 \\
        & Sparse Subnetwork Ensemble (NP) & \textbf{84.68} $\pm$ \textbf{0.04} & \textbf{57.26} $\pm$ \textbf{0.06} \\
\midrule
DenseNet-121 & Independent & 84.14 $\pm$ 0.09 & 56.62 $\pm$ 0.09 \\
          & Dropout & 79.59 $\pm$ 0.36 & 44.02 $\pm$ 0.58 \\
          & Bootstrap Aggregation & 81.16 $\pm$ 0.04 & 48.48 $\pm$ 0.06 \\
          & Sparse Subnetwork Ensemble (R) & 86.26 $\pm$ 0.07  & 62.02 $\pm$ 0.07 \\
            & Sparse Subnetwork Ensemble (NP) & \textbf{86.35} $\pm$ \textbf{0.04} & \textbf{62.09} $\pm$ \textbf{0.06} \\
\bottomrule
\end{tabularx}
\end{table}

Little attention has been given to deep ensemble methods that work well with small training budgets and/or resource constrained environments.
While it may seem counter-intuitive to construct ensembles with so few epochs, it is a valuable area of research as insights from these kinds of benchmarks hold the potential to significantly improve training efficiency and reduce computational cost \cite{coleman2017DAWNBench, mattson2020mlperf}.

The experiment presented here limits the total training budget to 16 epochs.
We compare the efficiency of Sparse Subnetwork Ensembles to a single independent model, a dropout model, and a traditional bootstrap aggregation (bagged) ensemble.
Each method is evaluated with three different model architectures (LeNet-L, ResNet-18, and DenseNet-121) and all models are trained with Adam using a fixed learning rate of $0.001$.

The independent model and the dropout model are identical to each other, except that the dropout model has dropout layers inserted after every hidden layer with a 50\% activation probability during training.
This parameter is chosen to keep the number of parameters consistent with the child networks generated by Sparse Subnetwork Ensembles.
The bagged ensemble contains 8 full size models that are each trained for two epochs on a randomly sampled 90\% subset of the training data.
Sparse Subnetwork Ensembles trains a single parent network for 8 epochs, then creates 8 children using unstructured pruning with a 50\% sparsity target and tunes each of those for an additional epoch. 
We include results for child networks created with random pruning (R) and with Neural Partitioning (NP), where 4 sets of partitioned pairs of child networks are generated.

Table 5.1 displays the mean accuracy and standard error achieved for each method over 10 runs, except for Sparse Subnetwork Ensembles which have 30 runs.
In this small fixed training budget experiment, comparisons show that the Sparse Subnetwork Ensembles are significantly better than all other methods for early training efficiency.
While the extent of hyperparameter tuning is limited for all methods, this experiment is designed to create an initial baseline for small training budgets that can be further explored in future work.

\begin{table}[t]
\caption{Results for ensembles of WideResNet-28-10 models on both CIFAR-10 and CIFAR-100. Methods with $^*$ denote results obtained from \cite{havasi2021training, liu2021freetickets}. Best low-cost ensemble results are {\bf bold}. cAcc, cNLL, and cECE correspond to corrupted test sets.}
\scriptsize
\begin{tabularx}{\textwidth}{X l l l l l l c c}
\toprule
Methods (CIFAR-10/WRN-28-10) & Acc $\uparrow$ & NLL $\downarrow$ & ECE $\downarrow$ & cAcc $\uparrow$ & cNLL $\downarrow$ & cECE $\downarrow$ & FLOPs $\downarrow$ & Epochs $\downarrow$ \\
\midrule
Independent Model$^*$ & 96.0 & 0.159 & 0.023 & 76.1 & 1.050 & 0.153 & 3.6e17 & 200 \\
Monte Carlo Dropout$^*$ & 95.9 & 0.160 & 0.024 & 68.8 & 1.270 & 0.166 & 1.00x & 200 \\
Snapshot (M=5) & 96.3 & 0.131 & 0.015 & 76.0 & 1.060 & 0.121 & 1.00x & 200 \\
Fast Geometric (M=12) & 96.3 & 0.126 & 0.015 & 75.4 & 1.157 & 0.122 & 1.00x & 200 \\
Sparse Subnetworks (M=6) & {\bf 96.5} & {\bf 0.113} & {\bf 0.005} & {\bf 76.2} & {\bf 0.972} & {\bf 0.081} & {\bf 0.85x} & 200 \\
\midrule
TreeNet (M=3)$^*$ & 95.9 & 0.258 & 0.018 & 75.5 & 0.969 & 0.137 & 1.52x & 250 \\
BatchEnsemble (M=4)$^*$ & 96.2 & 0.143 & 0.021 & 77.5 & 1.020 & 0.129 & 4.40x & 250 \\
MIMO (M=3)$^*$ & 96.4 & 0.123 & 0.010 & 76.6 & 0.927 & 0.112 & 4.00x & 250 \\
EDST (M=7)$^*$ & 96.4 & 0.127 & 0.012 & 76.7 & 0.880 & 0.100 & 0.57x & 850 \\
DST (M=3)$^*$ & 96.4 & 0.124 & 0.011 & 77.6 & 0.840 & 0.090 & 1.01x & 750 \\
Dense Ensemble (M=4)$^*$ & 96.6 & 0.114 & 0.010 & 77.9 & 0.810 & 0.087 & 1.00x & 800 \\
\bottomrule
\end{tabularx}

\vspace{0.1in}

\begin{tabularx}{\textwidth}{X l l l l l l c c }
\toprule
Methods (CIFAR-100/WRN-28-10) & Acc $\uparrow$ & NLL $\downarrow$ & ECE $\downarrow$ & cAcc $\uparrow$ & cNLL $\downarrow$ & cECE $\downarrow$ & FLOPs $\downarrow$ & Epochs $\downarrow$ \\
\midrule
Independent Model$^*$ & 79.8 & 0.875 & 0.086 & 51.4 & 2.700 & 0.239 & 3.6e17 & 200 \\
Monte Carlo Dropout$^*$ & 79.6 & 0.830 & 0.050 & 42.6 & 2.900 & 0.202 & 1.00x & 200 \\
Snapshot (M=5) & 82.1 & 0.661 & 0.040 & 52.2 & 2.595 & 0.145 & 1.00x & 200 \\
Fast Geometric (M=12) & 82.3 & 0.653 & 0.038 & 51.7 & 2.638 & 0.137 & 1.00x & 200 \\
Sparse Subnetworks (M=6) & {\bf 82.7} & {\bf 0.634} & {\bf 0.013} & {\bf 52.7} & {\bf 2.487} & {\bf 0.131} & {\bf 0.85x} & 200 \\
\midrule
TreeNet (M=3)$^*$ & 80.8 & 0.777 & 0.047 & 53.5 & 2.295 & 0.176 & 1.52x & 250 \\
BatchEnsemble (M=4)$^*$ & 81.5 & 0.740 & 0.056 & 54.1 & 2.490 & 0.191 & 4.40x & 250 \\
MIMO (M=3)$^*$ & 82.0 & 0.690 & 0.022 & 53.7 & 2.284 & 0.129 & 4.00x & 250 \\
EDST (M=7)$^*$ & 82.6 & 0.653 & 0.036 & 52.7 & 2.410 & 0.170 & 0.57x & 850 \\
DST (M=3)$^*$ & 82.8 & 0.633 & 0.026 & 54.3 & 2.280 & 0.140 & 1.01x & 750 \\
Dense Ensemble (M=4)$^*$ & 82.7 & 0.666 & 0.021 & 54.1 & 2.270 & 0.138 & 1.00x & 800 \\
\bottomrule
\end{tabularx}
\end{table}

\subsubsection{Large Budget}

We next evaluate Sparse Subnetwork Ensembles in the context of a standardized benchmark in low-cost ensemble learning literature.
This experiment uses a larger training budget of 200 epochs and the WideResNet-28-10 model architecture \cite{zagoruyko2017wide}. 
We use the same optimizer configuration and hyperparameter settings reported in other low-cost ensemble studies \cite{wasay2020mothernets, huang2017snapshot, garipov2018loss}.
We compare Sparse Subnetwork Ensembles with published results of several of the top low-cost ensemble methods including: TreeNets, Snapshot Ensembles, Fast Geometric Ensembles, BatchEnsembles, FreeTickets, and MIMO \cite{lee2015m, huang2017snapshot, garipov2018loss, wen2020batchensemble, liu2021freetickets, havasi2021training}.

All methods train the same WideResNet-28-10 model architecture using stochastic gradient descent with Nesterov momentum $\mu=0.9$ and weight decay $\gamma=0.0005$ \cite{sutskever13nesterov}. 
We use a batch size of 128 for training and use random crop, random horizontal flip, and mean standard scaling data augmentations for all approaches \cite{garipov2018loss, havasi2021training, liu2021freetickets, huang2017snapshot}. 
The parent network is trained with step-wise decaying learning rate schedule where an initial learning rate of $\eta_1=0.1$ is used for 50\% of the training budget and decays linearly to $\eta_2=0.001$ at 90\% of the training budget. 
The learning rate is kept constant at $\eta_2=0.001$ for the final 10\% of training. 

Table 5.2 includes results for all compared methods.
The Independent Model is a single WideResNet-28-10 model trained for 200 Epochs. 
The Dropout Model includes dropout layers between convolutional layers in the residual blocks at a rate of 30\% \cite{zagoruyko2017wide}. 
Snapshot Ensembles use a cosine annealing learning rate with an initial learning rate $\eta=0.1$ for a cycle length of 40 epochs \cite{huang2017snapshot}. 
Fast Geometric Ensembles use a pre-training routine for 156 epochs. A curve finding algorithm then runs for 22 epochs with a cycle length of 4, each starting from checkpoints at epoch 120 and 156. 
TreeNets, BatchEnsemble and MIMO are all trained for 250 epochs \cite{lee2015m,wen2020batchensemble,havasi2021training}. 
FreeTickets introduces several configurations for building ensembles. We include their two best configurations for Dynamic Sparse Training (DST, M=3, S=0.8) and Efficient Dynamic Sparse Training (EDST, M=7, S=0.9).



The Sparse Subnetwork ensemble size and training schedule is the same as those used in MotherNets and Fast Geometric Ensembles \cite{wasay2020mothernets, garipov2018loss}. 
Sparse Subnetwork Ensembles train a single parent network for 140 epochs. 
Six children are created with Neural Partitioning and tuned with a one-cycle learning rate for 10 epochs. 
The tuning schedule starts at $\eta_1=0.001$, increases to $\eta_2=0.1$ at 1 epoch and then decays to $\eta_3=1e-7$ using cosine annealing for the final 9 epochs. 
We additionally explored the effect of different partition sizes for child networks in the ensemble where $np = {1/2, 1/3, 1/6}$. Table 5.4 reports these results where we see a very minor degradation in accuracy moving from children with 50\% sparsity to 66\% sparsity.

We report the mean accuracy (Acc), negative log likelihood (NLL), and expected calibration error (ECE) over 10 runs on both CIFAR-10 and CIFAR-100 along with their corrupted variants \cite{nado2022uncertainty, hendrycks2019benchmarking}. 
We report the total number of floating point operations (FLOPs) and epochs used for training each method. 
We organize our tables into two groups based on training cost. 
The first group consists of low-cost training methods that take approximately as long as a single network would take to train. 
The second group of methods use either significantly more epochs or compute per epoch to achieve comparable performance. 
MIMO and BatchEnsemble both make use of batch repetition to train on more data while keeping the number of epochs low. 
FreeTickets (DST and EDST) use very sparse networks to keep FLOP counts low while using many more training epochs.

\begin{table}
\caption{Results for Sparse Subnetwork Ensembles on CIFAR-10 with different child network partition sizes. The parent network is trained for 140 epochs. 6 children are then created through neural partitioning and each is trained for an additional 10 epochs. Total Params counts all parameters in the ensemble.}
\footnotesize
\begin{tabularx}{\columnwidth}{c c c @{\hskip 0.2in}c @{\hskip 0.3in}c @{\hskip 0.3in}c @{\hskip 0.2in}c}
\toprule
Partition Size & Child Acc. $\uparrow$ & Ensemble Acc. $\uparrow$ & NLL $\downarrow$  & ECE $\downarrow$  & FLOPs $\downarrow$ & Total Params $\downarrow$ \\
\midrule
1/2 & 95.5 & 96.5 & 0.114 & 0.005 & 0.85x & 109.5M \\
1/3 & 95.3 & 96.4 & 0.115 & 0.004 & 0.80x & 73.0M \\
1/6 & 92.9 & 95.1 & 0.143 & 0.013 & 0.75x & 36.5M \\
\bottomrule
\end{tabularx}
\end{table}


Sparse Subnetwork Ensembles outperform all low-cost ensemble methods and are competitive with methods that train for significantly longer. 
Our approach produces well calibrated, robust and diverse ensembles with excellent performance on in distribution and out of distribution datasets.

\section{Discussion}

Sparse Subnetwork Ensembles prove to be a highly effective method for constructing diverse, accurate, and efficient ensembles of child networks.
We do this by first training a large parent network.
We then create child networks by sampling and pruning random subnetworks in the parent to produce a collection of sparse child networks with unique connective structures.
We then briefly train each child network for a small number of epochs where they quickly converge to new local optima.

Despite inheriting parameters from the same parent, child networks quickly converge to diverse solutions thanks to their inherited parameters and unique network topologies.
We further explore various ways to encourage diversity among child networks through neural partitioning and one-cycle tuning.
Neural partitioning is used to generate sets of child networks without any parametric overlap and one-cycle tuning uses a phasic learning rate schedule where large learning rates encourage children to move further apart in weight space before converging to local solutions.

The Sparse Subnetwork Ensemble algorithm is flexible and robust.
We conduct several ablation studies where we construct ensembles with various network architectures, pruning strategies, and training configurations.
We also visualize the loss landscape of a network throughout the lifecycle of this algorithm, where we see convergence to wide and flat optima that suggest regions of robust generalization.

This work showcases an important concept in Subnetwork Ensembles where it is found to be more efficient to first train a large parent network to a good solution before decomposing it into smaller child networks.
The large parent network has better optimization dynamics when trained from scratch, while the final state contains large amounts of unrealized capacity.
Sparse Subnetwork children leverage that redundancy to produce accurate and diverse representations with little additional computation.

With the benchmark experiments introduced, Sparse Subnetwork Ensembles outperform other low-cost ensembling methods on benchmark image classification datasets in both low training budget and high training budget regimes.
Our approach achieves comparable accuracy to ensembles that are trained on 4 to 5 times more data. 
Sparse Subnetwork Ensembles not only improve upon the training-time/accuracy trade-off, but also significantly reduces memory and computational cost.


\chapter{Stochastic Subnetwork Ensembles}
\label{chap:stochastic}
\section{Introduction}

A critical component of Sparse Subnetwork Ensembles is the creation and tuning of child networks. The standard approach explored in the previous chapter involves pruning the whole subnetwork in one-shot, followed by a tuning phase. Meanwhile, it's long been known that iterative pruning, which consists of alternating pruning/tuning cycles, is one of the most effective techniques to improve performance of extremely sparse networks \cite{blalock2020state}.

The key insight to iterative pruning is that removing too many parameters at once can lead to a drastic performance collapse as subnetworks get stuck in lower performing regions of the optimization landscape.
Pruning a small number of parameters over several epochs allows the model to better adapt to the changing network structure. However, these iterative methods can still result in subnetworks that are prone to overfitting local optimization regions.

We introduce a novel regularization technique for fine tuning these subnetworks by leveraging stochasticity for the network structure. Rather than using fixed subnetworks and discrete pruning operations, we instead represent subnetworks with probability matrices that determine how likely it is that a parameter is retained on any given forward pass.
The probability matrices are then adjusted during training through the use of dynamic annealing schedules which gradually evolves the network from a stochastic network towards a deterministic subnetwork structure over several epochs. The probabilistic inclusion of extra parameters early in the tuning process allows for gradient information to bleed through into the target subnetwork, encouraging robust adaptation and avoiding the drastic performance collapse observed with one-shot pruning methods.
This approach to representing subnetworks with stochastic masks is highly flexible and offers a lot of fine grained control over the tuning process.

We conduct a large scale ablation study to explore the dynamics of several annealing hyperparameters and their effects on convergence. 
These include the initial stochasticity, number of annealing epochs, amount of sparsity, learning rate schedules, and parameter selection strategies.
Our experiments demonstrate significant improvements over both one-shot and iterative pruning methods with especially large improvements in highly sparse subnetworks (95-98\%).
We then explore how this technique can be leveraged to tune child networks in Sparse Subnetwork Ensembles, where we observe improved accuracy and generalization on benchmark classification tasks.

\section{Implementation}

In pruning literature, sparse network structures are generally static and represented with binary bit masks. We propose a model of representing neural subnetworks with probabilistic masks, where each parameter is assigned a score that determines how likely it is that the parameter will be retained on any given forward pass. This introduces stochasticity into the subnetwork sampling process, which can act as a form of implicit regularization analagous to a reverse dropout, where parameters that would have been pruned have a chance to activate. This technique encourages exploration of a larger space of subnetworks during the fine-tuning phase, preventing it from becoming overly reliant on a single fixed topological configuration, resulting in more robust and generalized subnetworks.

Consider a weight matrix $W \in \mathbb{R}^{m \times n}$ representing the weights of a particular layer in a neural network. We introduce a probability matrix $P \in \mathbb{R}^{m \times n}$ containing scores that represent the probability that a parameter $W_{ij}$ will be will be masked on any given forward pass. The subnetwork mask $M \in \{0,1\}^{m \times n}$ is determined with a Bernoulli realization of the probability matrix $P$. 

The ideas of neural partitioning naturally extend to probabilistic subnetwork representation where the opposed probability matrix $P' = 1 - P$ can be used to create a pair of subnetworks. More partitions can be created so long as the probabilities for each parameter sum to one.
\begin{equation}
(P, P') = \left(
\begin{bmatrix}
0.6683 & 0.3923 & 0.9669 \\
0.8343 & 0.6885 & 0.8940 \\
0.2613 & 0.2249 & 0.9972
\end{bmatrix}
\begin{matrix} \\ \\, \end{matrix}
\begin{bmatrix}
0.3317 & 0.6077 & 0.0331 \\
0.1657 & 0.3115 & 0.1060 \\
0.7387 & 0.7751 & 0.0028
\end{bmatrix}
\right)
\end{equation}

We then anneal these probability values over some number of epochs such that subnetworks are slowly revealed throughout fine-tuning. This is done by introducing an annealing schedule, where the probability values for each parameter slowly move towards $0$ or $1$ depending on the target subnetwork sparsity. At the beginning of the training process, a high level of stochasticity is desirable as it encourages exploration of the weight space which may help to prevent overfitting and search for more effective subnetwork configurations. As training progresses, the stochasticity should gradually decrease to allow for stable convergence. The annealing function is arbitrary, with some popular examples being the linear, cosine, and exponential decay.
\begin{align}
y_{lin}(t) &= \tau_{max} - (\tau_{max} - \tau_{min})  \left( \frac{t}{T_{max}} \right) \\
y_{cos}(t) &= \tau_{min} + \frac{1}{2}(\tau_{max} - \tau_{min}) \left( 1 + cos \left( \frac{t}{T_{max}}\right) \right) \\
y_{exp}(t) &= \tau_{min} + (\tau_{max} - \tau_{min}) e^{- \frac{t}{T_{max}} k}
\end{align}

\subsection{Random Annealing}

The probability matrix can be generated in arbitrary ways. For example, a uniform distribution $P \sim U([0,1])^{m \times n}$ can be used to randomly assign probabilities to each parameter. All parameters with a value less than the sparsity target will then anneal towards 0 while parameters with a probability value greater than the sparsity target will anneal towards 1. This implementation with a uniform distribution results in a mean parameter activation of approximately 50\% at the beginning of tuning regardless of the target subnetwork sparsity.

\vspace{0.1in}

\begin{center}
\begin{tikzpicture}
\begin{axis}[height=5cm,
width=9.5cm,
axis lines=left,
samples=100,
smooth,
ymin=0,ymax=1.25,
xmin=0,xmax=1.25,
xtick={0,0.7,1}, ytick=\empty,
xticklabels={0, $S$, 1}]
\addplot[densely dashed, thick,black] coordinates {(0.7,0) (0.7,1)};
\addplot[very thick,black] coordinates {(1,0) (1,1)};
\addplot[very thick,black] coordinates {(0,0) (0,1)};
\addplot[very thick,black] coordinates {(0,1) (1,1)};

\addplot[->, thick,black] coordinates {(0.65,0.5) (0.05,0.5)};
\addplot[->, thick,black] coordinates {(0.75,0.5) (0.95,0.5)};


\end{axis}
\end{tikzpicture}
\end{center}


Mixtures of Gaussians can be used to assign interesting distributions that offer more fine grained control over network structure. For example, assume that a binary matrix $X \in \{0, 1\}^{m \times n}$ is randomly generated and used to index into a probability matrix $P \in \mathbb{R}^{m \times n}$. Using this index matrix $X$, we can sample from Gaussian distributions with different means and variances.
\begin{equation}
P = \begin{cases}
P_{ij} \sim \mathcal{N}(\mu_1, \sigma_1^2), \ \text{if} \ X_{ij} = 0\\
P_{ij} \sim \mathcal{N}(\mu_2, \sigma_2^2), \ \text{if} \ X_{ij} = 1
\end{cases}
\end{equation}

\pgfmathdeclarefunction{gauss}{2}{%
  \pgfmathparse{1/(#2*sqrt(2*pi))*exp(-((x-#1)^2)/(2*#2^2))}%
}

\vspace{0.1in}


\begin{center}
\begin{tikzpicture}
\begin{axis}[
  no markers, domain=0:1.25, samples=100,
  axis lines=left,
  every axis y label/.style={at=(current axis.above origin),anchor=south},
  every axis x label/.style={at=(current axis.right of origin),anchor=west},
  height=5cm, width=9.5cm,
  xtick={0,1}, ytick=\empty,
  enlargelimits=false, clip=true, axis on top,
  ]
  \addplot[thick,black] coordinates {(1,0) (1,3)};
  \addplot [fill=black!20, draw=none, domain=0:1] {gauss(0.25,0.15)} \closedcycle;
  \addplot [very thick,black!50!black] {gauss(0.25,0.15)};
  \addplot [very thick,black!50!black, restrict x to domain=0:1] {gauss(0.75,0.15)};
\end{axis}
\end{tikzpicture}
\end{center}

This approach to constructing multi-modal distributions can enable many distinct formulations of stochastic subnetworks. Future work may find natural applications to multi-task learning, where certain groups of parameters can be strongly correlated with each other while allowing for overlap with other task specific subnetworks. This stochastic overlap may encourage shared portions of the network to learn generalized features, avoiding the problem in typical network-splitting approaches where task specific subnetworks become increasingly narrow and lead to degraded performance. 

\subsection{Temperature Annealing}

Temperature scaling may be applied to binary matrices in order to allow for an even application of stochasticity, reducing some variance relative to random annealing. Assume that a binary matrix $X \in \{0, 1\}^{m \times n}$ is generated at random. A temperature scaling constant $\tau$ is introduced such that the values in $X$ with a 1 are decayed by $\tau$ and the values with 0 are increased by $\tau$. The mask is then determined on every forward pass with an altered probability according to $\tau$. Altering the initial value of $\tau$ allows for a much more controlled approach to noise injection during the early phases of tuning, which can be desirable when the number of total tuning epochs is limited.
%

A variation of temperature scaling, where $\tau$ is only applied to parameters that are not a part of the target subnetwork, acts as a highly effective form of regularization analagous to reverse Dropout. That is, the subnetwork is always active, but parameters that would have been pruned away now have a chance to pop back in during tuning. This formulation ensures that the target subnetwork will be optimized on every training step, which can enable faster convergence. Allowing other parameters to become active allows for gradient information to contribute to optimization of the target subnetwork which can help to encourage avoidance of local minima.

\begin{equation}
P = \begin{bmatrix}
1 & 1 & 0+\tau & 1 \\
0+\tau & 1 & 0+\tau & 0+\tau \\
1 & 0+\tau & 0+\tau & 1 \\
0+\tau & 0+\tau & 1 & 1
\end{bmatrix}
\end{equation}




\begin{algorithm}
\SetAlgoLined
\KwIn{N, the number of ensemble members}
\KwIn{($X_{train}, X_{test}$), \text{the training, validation, and test data}}
\KwIn{($t_{parent}, t_{child}, t_{anneal}$), \text{the number of parent, child, and annealing epochs}}
\KwIn{($\eta_{parent}, \eta_{child}$), the parent and child optimizer hyperparameters}
\KwIn{$\phi(\rho)$, \text{the subnetwork sampling process with target sparsity $\rho$}}
\KwIn{$\Theta(M, \tau, e, t)$, the temperature scaling function applied to a mask $M$, with initial temperature $\tau$, for current epoch $e$, over $t$ total annealing epochs}

\texttt{\\}
$O = sgd(\eta_{parent})$ \\
$F = initialize()$ \\
$F.w = train(F, O, X_{train}, t_{parent})$ \\
\texttt{\\}
$ensemble = []$ \\
\texttt{\\}
\For{i in 1 to $N$}{
    $f_i = initialize()$ \\
    $f_i.w = F.w$ \\
    \texttt{\\}
    \For{$j$ in $f_i.layers$}{
        $M_j \sim \phi(\rho)^{f_{i,j}.w}$ \\
    }
    \texttt{\\}
    $O = sgd(\eta_{child})$ \\
    \texttt{\\}
\For{$e$ in $t_{child}$}{
        $P \leftarrow \Theta(M, \tau, e, t_{anneal})$ \\
    \texttt{\\}
    \For{$(x, y)$ in $X_{train}$}{
        $\hat{f}_i.w \leftarrow f_i.w \circ Bernoulli(P)$ \\
        $f_i.w = train(\hat{f}_i, O, (x,y))$ \\
    }
}
    \texttt{\\}
    $ensemble \leftarrow f_i$ \\
}
\texttt{\\}
\For{$(x, y)$ in $X_{test}$}{
    $outputs = [F(x), c(x) \text{ for c in ensemble}$] \\
    $predictions = softmax(mean(outputs))$ \\
    $loss = nll(predictions, y)$ \\
    $accuracy = acc(predictions, y)$ \\
}
\caption{Stochastic Subnetwork Ensembles}
\end{algorithm}

\section{Experiments}

\subsection{Ablations}

We begin with an exploration of several stochastic subnetwork annealing configurations with the goal of investigating how different hyperparameters impact the efficiency and quality of subnetwork convergence, compared to established one-shot and iterative pruning techniques.

All ablations use the same ResNet-18 trained for 100 epochs using a standardized optimization configuration \cite{he2015deep}. We use PyTorch's Stochastic Gradient Descent optimizer with an initial learning rate of $0.1$ and Nesterov momentum of $0.9$ \cite{sutskever13nesterov}. After 50 epochs, the learning rate is decayed to $0.01$ and again to $0.001$ for the final 10 epochs.
We use the benchmark CIFAR datasets with standard data augmentations including random crop, random horizontal flip, and mean standard normalization.

\begin{table}
\includegraphics[width=0.48\textwidth]{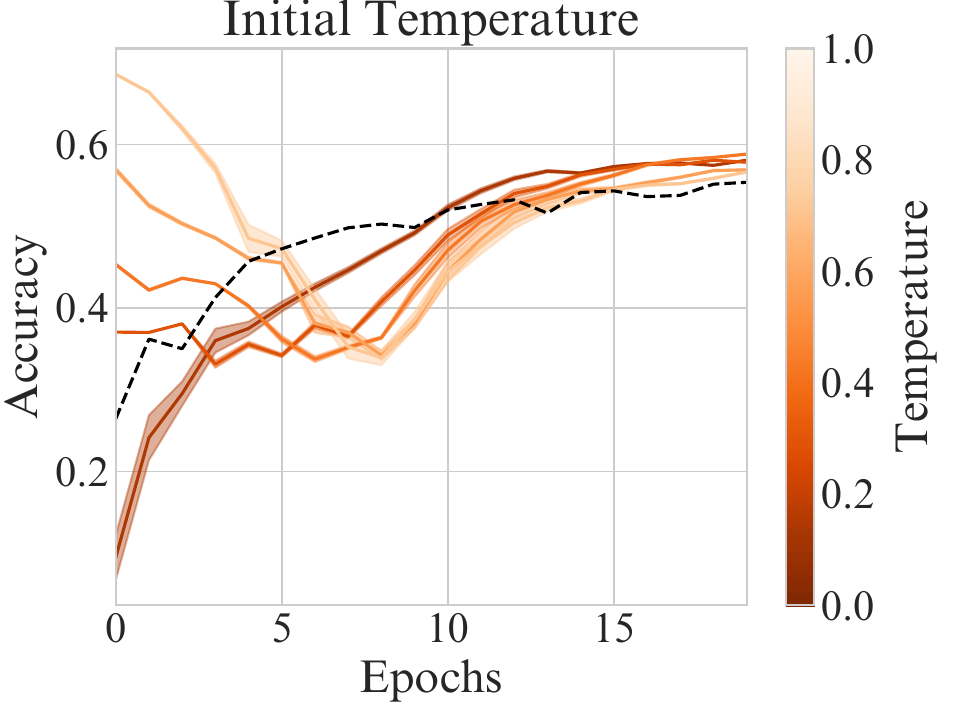}
\hspace{0.1in}
\includegraphics[width=0.48\textwidth]{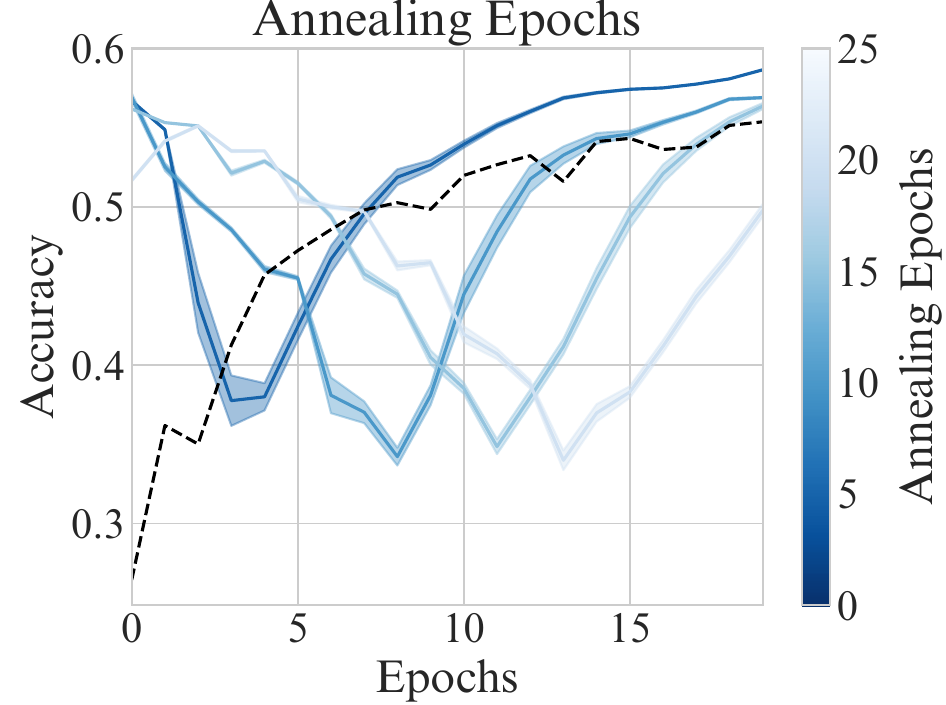}
\captionof{figure}{Ablations exploring how the initial temperature and the number of annealing epochs affect convergence behavior.}
\vspace{0.1in}
\captionof{table}{Comparison between various baseline approaches when used with random/magnitude pruning on CIFAR-100. We report the best accuracy for each method at various levels of target sparsity when tuned with both a constant learning rate and a one-cycle learning rate schedule. Our annealing strategies consistently outperform established one-shot and iterative pruning methods.}
\vspace{0.1in}
\footnotesize
\begin{tabularx}{\columnwidth}{X c c c c c}
\toprule
Random Pruning/Constant LR & 50\% & 70\% & 90\% & 95\% & 98\% \\
\midrule
One-Shot Baseline & 0.640 & 0.634 & 0.553 & 0.490 & 0.401 \\
Iterative Pruning & 0.655 & 0.642 & 0.578 & 0.502 & 0.422 \\
Random Annealing & 0.655 & 0.636 & 0.591 & 0.546 & 0.457 \\
Temperature Annealing & \textbf{0.658} & \textbf{0.643} & \textbf{0.595} & \textbf{0.554} & \textbf{0.463} \\
\bottomrule
\end{tabularx}

\vspace{0.2in}

\begin{tabularx}{\columnwidth}{X c c c c c}
\toprule
Random Pruning/One-Cycle LR & 50\% & 70\% & 90\% & 95\% & 98\% \\
\midrule
One-Shot Baseline & 0.702 & 0.685 & 0.620 & 0.573 & 0.493 \\
Iterative Pruning & 0.702 & 0.692 & 0.639 & 0.573 & 0.508 \\
Random Annealing & 0.703 & 0.692 & 0.646 & 0.600 & 0.527 \\
Temperature Annealing & \textbf{0.706} & \textbf{0.696} & \textbf{0.649} & \textbf{0.602} & \textbf{0.533} \\
\bottomrule
\end{tabularx}

\vspace{0.2in}

\begin{tabularx}{\columnwidth}{X c c c c c}
\toprule
Magnitude Pruning/Constant LR & 50\% & 70\% & 90\% & 95\% & 98\% \\
\midrule
One-Shot Baseline & 0.673 & 0.668 & 0.661 & 0.635 & 0.542 \\
Iterative Pruning & 0.682 &0.680 & 0.660 & 0.633 & 0.564 \\
Temperature Annealing & \textbf{0.685} & \textbf{0.682} & \textbf{0.663} & \textbf{0.635} & \textbf{0.566} \\
\bottomrule
\end{tabularx}

\vspace{0.2in}

\begin{tabularx}{\columnwidth}{X c c c c c}
\toprule
Magnitude Pruning/One-Cycle LR & 50\% & 70\% & 90\% & 95\% & 98\% \\
\midrule
One-Shot Baseline & 0.709 & 0.701 & 0.697 & 0.675 & 0.604 \\
Iterative Pruning & 0.711 & 0.711 & 0.700 & 0.675 & 0.606 \\
Temperature Annealing & \textbf{0.714} & \textbf{0.714} & \textbf{0.702} & \textbf{0.678} & \textbf{0.615} \\
\bottomrule
\end{tabularx}
\end{table}

Pruning is done in a layerwise unstructured fashion, after which each subnetwork is tuned for an additional 20 epochs. We experiment with both a constant learning rate of $0.01$ and a one-cycle policy with a max learning rate of $0.1$. We explore the results for each configuration with different levels of final subnetwork sparsity $\rho \in [0.5, 0.7, 0.9, 0.95, 0.98]$. We additionally include results for subnetworks created through L1 unstructured pruning. In this case, parameters with the smallest magnitudes at each layer are pruned.

The one-shot baseline prunes the parent network to the target sparsity before we start tuning. 
Iterative pruning calculates the number of parameters to remove at the beginning of each epoch over some number of pruning epochs $\varepsilon \in [5, 10, 15, 20]$ such that the final target sparsity is hit.
Random Annealing uses a probability matrix that is generated according to a random uniform distribution, where each parameter is assigned a value between 0 and 1. Parameters with a value less than the target sparsity value are linearly annealed to 0 and values greater than the target sparsity are linearly annealed to 1, over some number of annealing epochs $\varepsilon \in [5, 10, 15, 20]$. 
Temperature Annealing uses a randomly generated binary bitmask according to the target sparsity. All parameters with a value of 0 are modified to an initial temperature value of $\tau \in [0.2, 0.4, 0.6, 0.8, 1.0]$. Those values are then cosine annealed to 0 over some number of annealing epochs $\varepsilon \in [5, 10, 15, 20]$. 

Table 6.1 includes the accuracies for the best configurations of each method on CIFAR-100. We see consistent improvement with both of our stochastic annealing methods over the baseline one-shot and iterative pruning techniques across all sparsities and with both a constant learning rate and a one-cycle learning rate policy. As networks become more sparse, the benefits from our annealing approaches become more significant, with a 6\% and 4\% improvement at 98\% sparsity over the one-shot baseline with a constant and one-cycle rate schedule respectively. We also observed improved performance with magnitude pruning, however the differences were smaller at 2\% and 1\% improvement.



\begin{figure}[t]
    \includegraphics[width=0.48\textwidth]{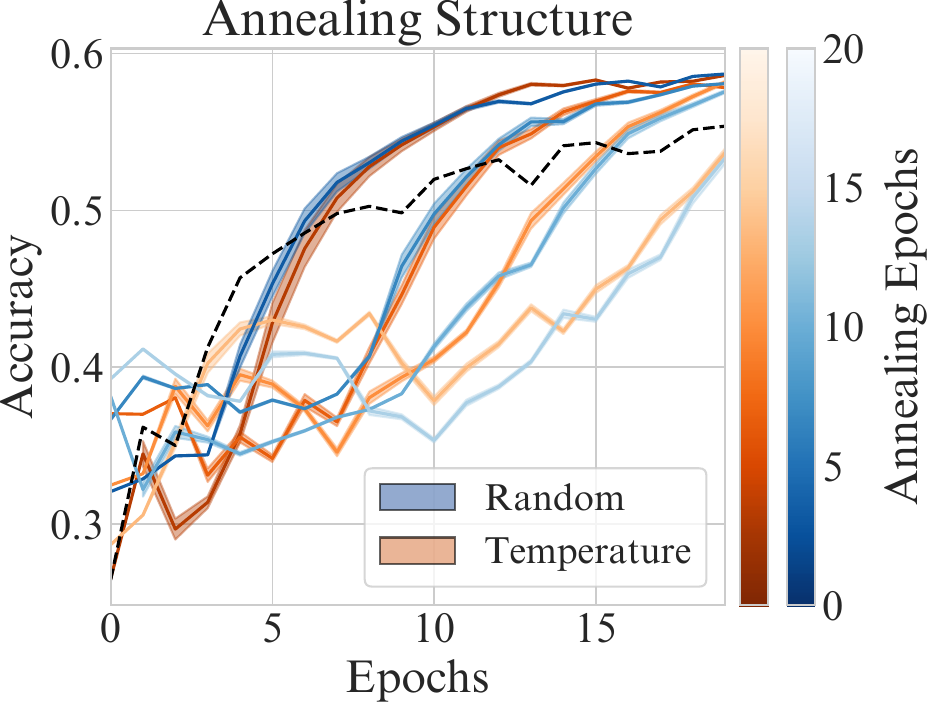}
    \hspace{0.1in}
    \includegraphics[width=0.48\textwidth]{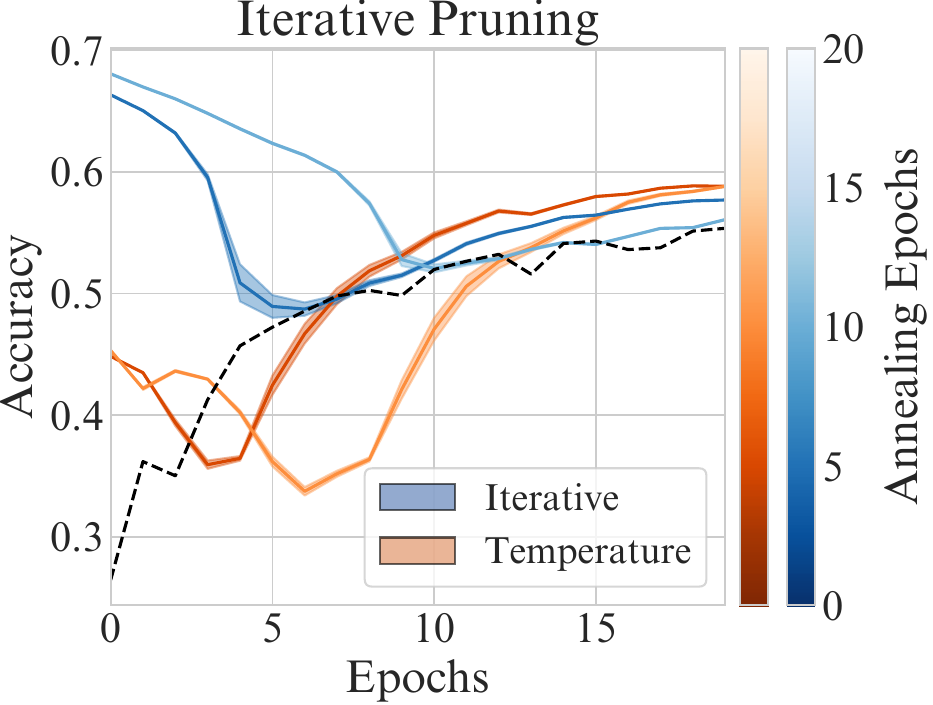}
    \caption{Ablations over the number of annealing epochs for random annealing vs temperature annealing and temperature annealing vs iterative pruning.}
\end{figure}

Figure 6.1 includes an exploration of the initial temperature and the number of annealing epochs.
The hyperparameter with the most significant impact on all pruning methodologies was the number of epochs that were pruned or annealed over. We saw best results for all subnetwork sparsities with $e=5$ or $e=10$ annealing epochs. It's important for the final subnetwork topology to be established for a sufficient number of epochs to allow for optimal convergence behavior. This pattern holds for both constant and one-cycle learning rate schedules. 
The initial temperature has a smaller impact on performance than the number of annealing epochs. A small value of $\tau$ means that the target subnetwork is always active and other parameters have a small chance to turn on, while a high value of $\tau$ means that the target subnetwork is always active and other parameters have a high chance to turn on. We saw best results when $\tau$ was in the $0.4$ to $0.6$ range.
This corresponds to a higher state of entropy regarding network structure which results in a stronger regularization effect for those initial training examples.



\begin{figure}[t]
    \includegraphics[width=0.48\textwidth]{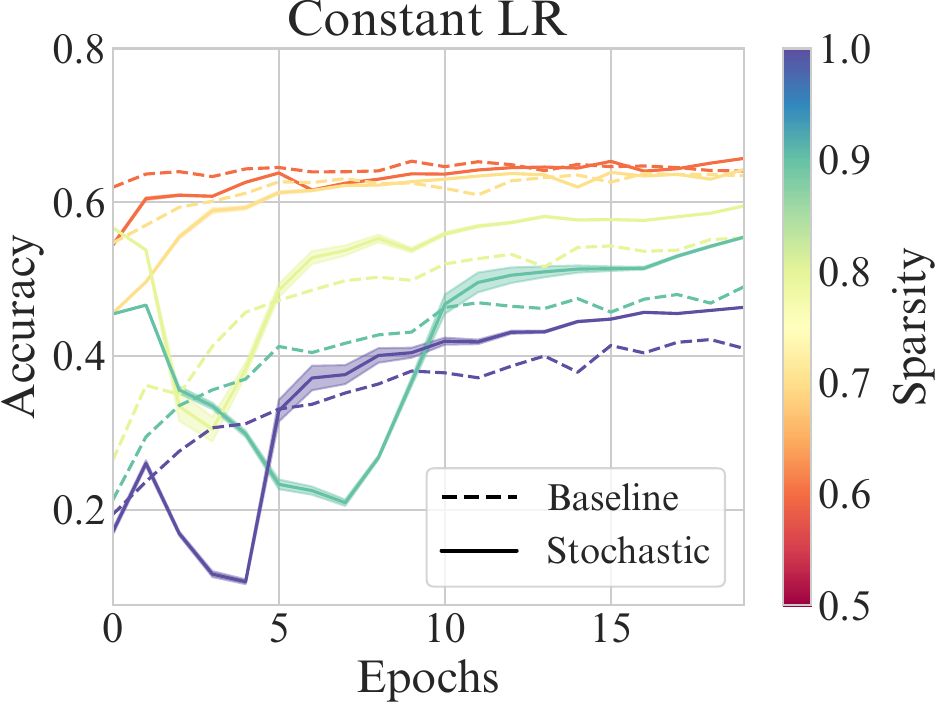}
    \hspace{0.1in}
    \includegraphics[width=0.48\textwidth]{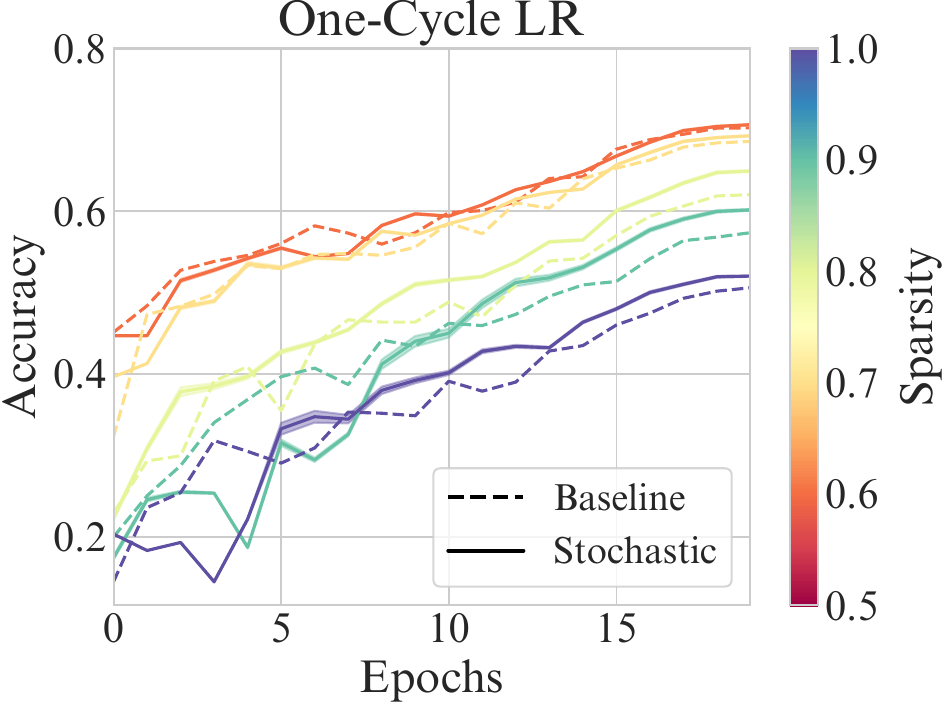}
    \caption{The rightmost graph displays the best performing annealing configuration for each sparsity and plots it against the one-shot pruning baseline. Stochastic annealing outperformed the other methods for each sparsity, with more drastic improvements appearing in extremely sparse networks.}
\end{figure}

Figure 6.2 includes an exploration of random annealing vs temperature annealing and iterative pruning vs temperature annealing. 
Despite the additional variance associated with random annealing, we found that the performance was very good and nearly approached that of temperature annealing.
Temperature annealing consistently outperformed iterative pruning. The graph displays results for annealing epochs $\varepsilon \in [5,10]$ and a target subnetwork sparsity of $0.9$.
While the early accuracy of temperature annealing appears worse, it quickly surpasses iterative pruning as the annealing phase ends.

Figure 6.3 displays the test trajectories for the best models over all sparsities with both a constant learning rate schedule and a one-cycle learning rate schedule.
We compare the results of temperature annealing with that of the one-shot baseline. Temperature Annealing results in better performance for both learning rate schedules at all levels of sparsity that we tested $\rho \in [0.5, 0.7, 0.9, 0.95, 0.98]$. 
The benefits of temperature annealing are most noticeable at higher levels of sparsity with the constant learning rate schedule. 
The one-cycle schedule does result in much better generalization and a reduced accuracy gap between temperature annealing and the baseline. 

\subsection{Benchmarks}

We next evaluate the efficacy of Stochastic Subnetwork Annealing in the context Sparse Subnetwork Ensembles. We include results for these ensembles on corrupted versions of the CIFAR datasets as well, in order to test robustness on out-of-distribution images. These corrupted datasets are generated by adding 20 different kinds of image corruptions (Gaussian noise, snow, blur, pixelation, etc.) at five different levels of severity to the original test sets. The total number of images in each of these additional sets is 1,000,000.

We take the training configuration, ensemble size and parameter settings directly from the large scale benchmark experiment described in the Sparse Subnetwork Ensemble chapter.

All methods compared use WideResNet-28-10 and Stochastic Gradient Descent with Nesterov momentum $\mu=0.9$ and weight decay $\gamma=0.0005$ \cite{sutskever13nesterov}. We use a batch size of 128 for training and use random crop, random horizontal flip, and mean standard normalization data augmentations for all approaches \cite{garipov2018loss, havasi2021training, liu2021freetickets, huang2017snapshot}. The parent learning rate uses a step-wise decay schedule. An initial learning rate of $\eta_1=0.1$ is used for 50\% of the training budget which decays linearly to $\eta_2=0.001$ at 90\% of the training budget. The learning rate is kept constant at $\eta_2=0.001$ for the final 10\% of training. 

We train a single parent network for 140 epochs. Six children are then created by randomly pruning 50\% of the connections in the parent network. Neural partitioning is implemented where pairs of children are generated with opposite sets of inherited parameters. Each child is tuned with a one-cycle learning rate for 10 epochs. The tuning schedule starts at $\eta_1=0.001$, increases to $\eta_2=0.1$ at 1 epoch and then decays to $\eta_3=1e-7$ using cosine annealing for the final 9 epochs.

We implement Stochastic Annealing when tuning the child networks by using the procedure illustrated in Section 6.3.1 for Temperature Annealing. We initialize the binary subnetwork masks with with a target sparsity of $0.5$. We then modify the masks such that the parameters with a value of zero are modified to have an initial temperature of $\tau=0.5$. That value is decayed to $\tau=0$ over 3 annealing epochs using a cosine decay.
We compare the results of these Stochastic Subnetwork Ensembles tuned with Temperature Annealing to the same low-cost methods described in the Chapter 5 (Sparse Subnetwork Ensembles).

\begin{table}[t]
\scriptsize
\caption{Results for ensembles of WideResNet-28-10 models on both CIFAR-10 and CIFAR-100. Methods with $^*$ denote results obtained from \cite{havasi2021training, liu2021freetickets}. Best low-cost ensemble results are {\bf bold}. cAcc, cNLL, and cECE correspond to corrupted test sets.}
\begin{tabularx}{\textwidth}{X l l l l l l c c}
\toprule
Methods (CIFAR-10/WRN-28-10) & Acc $\uparrow$ & NLL $\downarrow$ & ECE $\downarrow$ & cAcc $\uparrow$ & cNLL $\downarrow$ & cECE $\downarrow$ & FLOPs $\downarrow$ & Epochs $\downarrow$ \\
\midrule
Independent Model$^*$ & 96.0 & 0.159 & 0.023 & 76.1 & 1.050 & 0.153 & 3.6e17 & 200 \\
Monte Carlo Dropout$^*$ & 95.9 & 0.160 & 0.024 & 68.8 & 1.270 & 0.166 & 1.00x & 200 \\
Snapshot (M=5) & 96.3 & 0.131 & 0.015 & 76.0 & 1.060 & 0.121 & 1.00x & 200 \\
Fast Geometric (M=12) & 96.3 & 0.126 & 0.015 & 75.4 & 1.157 & 0.122 & 1.00x & 200 \\
Sparse Subnetworks (M=6) & 96.5 & 0.113 & 0.005 & 76.2 & 0.972 & 0.081 & 0.85x & 200 \\
Stochastic Annealing (M=6) & \textbf{96.7} & \textbf{0.110} & \textbf{0.005} & \textbf{76.3} & \textbf{0.968} & \textbf{0.079} & \textbf{0.85x} & \textbf{200} \\
\midrule
TreeNet (M=3)$^*$ & 95.9 & 0.258 & 0.018 & 75.5 & 0.969 & 0.137 & 1.52x & 250 \\
BatchEnsemble (M=4)$^*$ & 96.2 & 0.143 & 0.021 & 77.5 & 1.020 & 0.129 & 4.40x & 250 \\
Multi-Input Multi-Output (M=3)$^*$ & 96.4 & 0.123 & 0.010 & 76.6 & 0.927 & 0.112 & 4.00x & 250 \\
FreeTickets (EDST) (M=7)$^*$ & 96.4 & 0.127 & 0.012 & 76.7 & 0.880 & 0.100 & 0.57x & 850 \\
FreeTickets (DST) (M=3)$^*$ & 96.4 & 0.124 & 0.011 & 77.6 & 0.840 & 0.090 & 1.01x & 750 \\
Dense Ensemble (M=4)$^*$ & 96.6 & 0.114 & 0.010 & 77.9 & 0.810 & 0.087 & 1.00x & 800 \\
\bottomrule
\end{tabularx}

\vspace{0.2in}

\begin{tabularx}{\textwidth}{X l l l l l l c c }
\toprule
Methods (CIFAR-100/WRN-28-10) & Acc $\uparrow$ & NLL $\downarrow$ & ECE $\downarrow$ & cAcc $\uparrow$ & cNLL $\downarrow$ & cECE $\downarrow$ & FLOPs $\downarrow$ & Epochs $\downarrow$ \\
\midrule
Independent Model$^*$ & 79.8 & 0.875 & 0.086 & 51.4 & 2.700 & 0.239 & 3.6e17 & 200 \\
Monte Carlo Dropout$^*$ & 79.6 & 0.830 & 0.050 & 42.6 & 2.900 & 0.202 & 1.00x & 200 \\
Snapshot (M=5) & 82.1 & 0.661 & 0.040 & 52.2 & 2.595 & 0.145 & 1.00x & 200 \\
Fast Geometric (M=12) & 82.3 & 0.653 & 0.038 & 51.7 & 2.638 & 0.137 & 1.00x & 200 \\
Sparse Subnetworks (M=6) & 82.7 & 0.634 & 0.013 & 52.7 & 2.487 & 0.131 & 0.85x & 200 \\
Stochastic Annealing (M=6) & \textbf{83.1} & \textbf{0.633} & \textbf{0.010} & \textbf{52.8} & \textbf{2.440} & \textbf{0.131} & \textbf{0.85x} & \textbf{200} \\
\midrule
TreeNet (M=3)$^*$ & 80.8 & 0.777 & 0.047 & 53.5 & 2.295 & 0.176 & 1.52x & 250 \\
BatchEnsemble (M=4)$^*$ & 81.5 & 0.740 & 0.056 & 54.1 & 2.490 & 0.191 & 4.40x & 250 \\
Multi-Input Multi-Output (M=3)$^*$ & 82.0 & 0.690 & 0.022 & 53.7 & 2.284 & 0.129 & 4.00x & 250 \\
FreeTickets (EDST) (M=7)$^*$ & 82.6 & 0.653 & 0.036 & 52.7 & 2.410 & 0.170 & 0.57x & 850 \\
FreeTickets (DST) (M=3)$^*$ & 82.8 & 0.633 & 0.026 & 54.3 & 2.280 & 0.140 & 1.01x & 750 \\
Dense Ensemble (M=4)$^*$ & 82.7 & 0.666 & 0.021 & 54.1 & 2.270 & 0.138 & 1.00x & 800 \\
\bottomrule
\end{tabularx}
\end{table}

We report the mean accuracy (Acc), negative log likelihood (NLL), and expected calibration error (ECE) over 10 runs on both CIFAR-10 and CIFAR-100 along with their corrupted variants \cite{nado2022uncertainty, hendrycks2019benchmarking}. We also report the total number of floating point operations (FLOPs) and epochs used for training each method. We organize our tables into two groups based on training cost. The first group consists of low-cost training methods that take approximately as long as a single network would take to train. The second group of methods use either significantly more epochs or compute per epoch to achieve comparable performance.

\section{Discussion}

Stochastic Subnetwork Annealing offers a novel approach to tuning pruned models by representing subnetworks with probabilistic masks. Rather than discretely removing parameters, we instead create probability matrices that alter the chance for parameters to be retained on any given forward pass.
We then anneal those probability values towards a deterministic binary mask over several epochs.
We introduce several approaches for constructing stochastic subnetworks.
Random annealing samples from a uniform distribution to assign random probabilities for every parameter, while temperature annealing applies an even amount of stochasticity to all parameters not in the target subnetwork.

The efficacy of Stochastic Subnetwork Annealing is affirmed by recent insights revealed in optimization research investigating the impact of warmup epochs \cite{gilmer2021loss, ma2021adequacy, gotmare2018closer}. 
Early epochs are critical during optimization as they set the foundational trajectory for the rest of the training process.
Stochastic Subnetwork Annealing provides effective regularization during the early epochs of subnetwork tuning to promote training stability and encourage robust adaptation.

Our experiments display marked improvement over the established one-shot and iterative pruning benchmarks for subnetworks at various levels of sparsity. This technique is especially effective for very sparse models up to 98\%. We conduct an extensive ablation study to explore the dynamics of this technique with regard to different hyperparameters, pruning methodologies, and learning rate schedules.
We additionally implement this technique in the context of Sparse Subnetwork Ensembles where we report significantly better performance against benchmark ensemble methods.


\chapter{Diversity Analysis}
\label{chap:diversity}
\section{Introduction}

Diversity is known to be an important consideration in the construction of robust neural network ensembles \cite{ueda1996generalization, brown2005diversity, kuncheva2003measures}. A collection of well trained models will generalize better if they are diverse in the patterns they respond to and the predictions they make. Diversity is especially important for low-cost ensemble methods because members often share network structure in order to avoid training several independent models from scratch. Diversity is traditionally analyzed by measuring differences between the outputs of models. However, this gives little insight into how knowledge representations differ between ensemble members. 


Meanwhile, neural network interpretability methods continue to grow in popularity and importance in machine learning research \cite{zhang2021interpretability}. 
This field is shifting the preconceived notions that deep neural networks act as black box models. By utilizing feature visualization and saliency attribution techniques, it becomes possible to better explore \textit{how} and \textit{why} neural networks make decisions by visualizing the types of patterns that specific neurons within the network respond to.
These methods are especially useful for evaluating low-cost ensemble methods that share network structure, because we can meaningfully compare feature representations of the same neuron in two different networks and visualize how they evolve and change over time.



We explore these ideas by conducting a large scale analysis of both traditional diversity metrics and this new interpretable approach in Sparse Subnetwork Ensembles.
Our experiments reveal that the unique network topology of Sparse Subnetwork children encourages each ensemble member to learn and respond to unique patterns and feature representations.
This insight could be highly effective for building more robust ensemble techniques in the future as our experiments offer a new perspective for analyzing diversity.
The application of interpretability techniques can be invaluable for better understanding how to measure and promote diversity in low-cost ensembles.

\section{Output Diversity}

\begin{table}[t]
\caption{Average prediction disagreement ratio and KL divergence between ensemble members for several benchmark methods on CIFAR-10 with WideResNet-28x10. Results reported from \cite{liu2021freetickets, havasi2021training, whitaker2022prune}}
\footnotesize
\begin{tabularx}{\columnwidth}{X c c c}
\toprule
Methods & $d_{PDR}$ $\uparrow$ & $d_{KL}$ $\uparrow$ & Acc $\uparrow$ \\
\midrule
Treenet & 0.010 & 0.010 & 95.9\\
BatchEnsemble & 0.014 & 0.020 & 96.2 \\
EDST Ensemble & 0.026 & 0.057 & 96.4 \\
MIMO & 0.032 & 0.081 & 96.4 \\
Dense Ensemble & 0.032 & 0.086 & 96.6 \\
Snapshot Ensemble & 0.034 & 0.068 & 96.3 \\
DST Ensemble & 0.035 & 0.095 & 96.4 \\
Sparse Subnetwork Ensemble & 0.036 & 0.090 & 96.5 \\
\bottomrule
\end{tabularx}
\end{table}

Several metrics have been used to quantify diversity in classification ensembles, the most popular of which are the pairwise output correlation, Kullback-Leibler divergence, and prediction disagreement ratio \cite{kuncheva2003measures}.
Pairwise Output Correlation measures the average correlation between the outputs for each pair of models in the ensemble (Eqn. 7.1).
Kullback-Leibler Divergence measures the relative entropy between two probability distributions (Eqn. 7.2). It approximately describes how different one probability distribution is from one another.
Prediction Disagreement Ratio measures the pairwise ratio of disagreements between the predicted classes of models (Eqn. 7.3).
\begin{align}
    d_{corr}(f_1,f_2) &= \frac{1}{N} \sum_{i=1}^{N} \sum_{c=1}^C  \frac{\text{cov}(f_1(x_i)_c, f_2(x_i)_c)}{\sigma_{f_1(x_i)_c} \sigma_{f_2(x_i)_c}} \\
    d_{KL}(f_1, f_2) &= \frac{1}{N} \sum_{i=1}^N \sum_{c=1}^C f_1(x_i)_c \log \left( \frac{f_1(x_i)_c}{f_2(x_i)_c} \right)\\
    d_{PDR}(f_1, f_2) &= \frac{1}{N} \sum_{i=1}^N argmax \ (f_1(x_i)) \neq argmax \ (f_2(x_i))
\end{align}

\noindent where $N$ is the number of test samples, $x_i$ in an input sample, $f_i(x_i)$ is the output of a model, $argmax(f_i(x_i))$ is the predicted class label, $cov()$ is the covariance function, and $\sigma_{f(x_i)}$ is the standard deviation of the output vector.

Table 7.1 displays average prediction disagreement ratios and KL divergences for several low-cost ensembles on CIFAR-10 with WideResNet-28x10. These are the same ensembles evaluated in the large scale benchmark experiments described in Chapters 5 and 6.
Figure 7.1 displays the pairwise correlation values for three variations of Sparse Subnetwork Ensembles.
These include children created with random pruning (R), children created with neural partitioning (NP), and children tuned with bootstrap aggregation (BAG). While bootstrap aggregation does result in more diverse children, the accuracy is lower than methods where the children are trained on the whole dataset.

\begin{figure}[t]
\includegraphics[width=\textwidth]{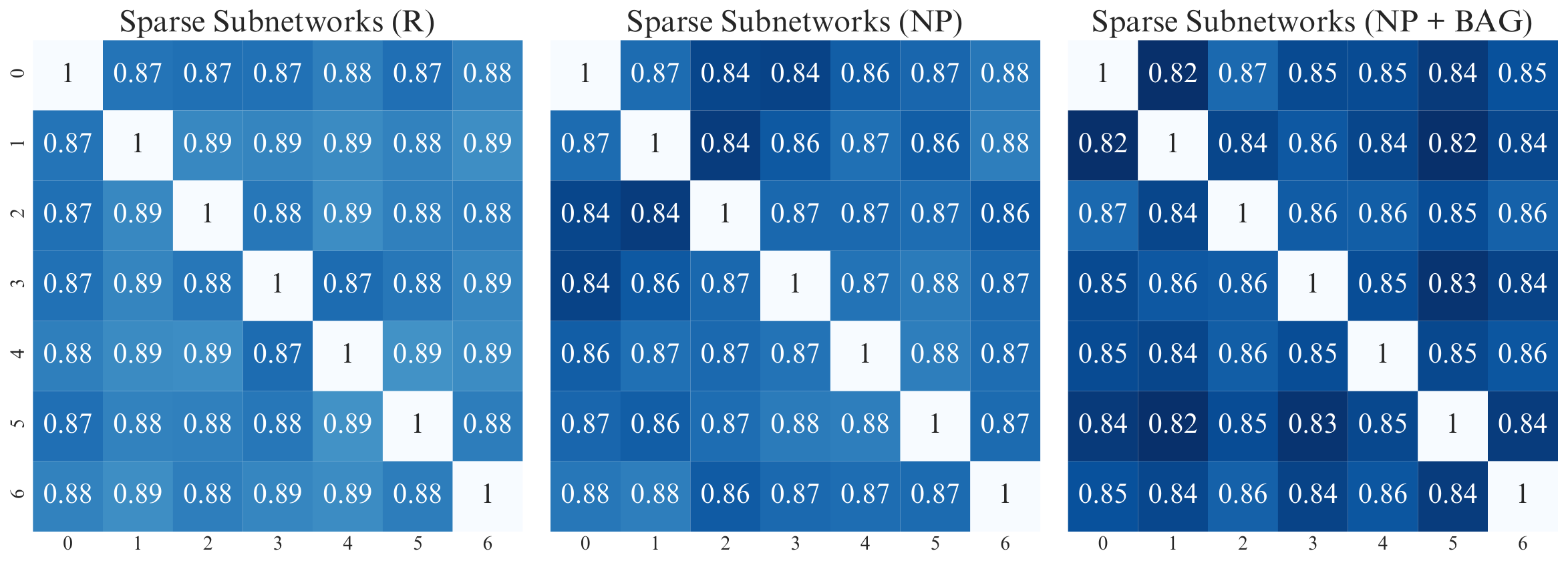}

\vspace{0.1in}

\small
\begin{tabularx}{\columnwidth}{X c c}
\toprule
~ & Ensemble Accuracy $\uparrow$ & Mean Correlation $\downarrow$  \\
\midrule
Sparse Subnetwork Ensemble (R) & 0.840 & 0.893 $\pm$ 0.041 \\
Sparse Subnetwork Ensemble (NP) & 0.844 & 0.884 $\pm$ 0.045 \\
Sparse Subnetwork Ensemble (NP + BAG) & 0.839 & 0.868 $\pm$ 0.051 \\
\bottomrule
\end{tabularx}
\caption{Pairwise correlation heatmaps for ResNet-18 on CIFAR-10. (R/NP) denotes random pruning/neural partitioning. (BAG) denotes bootstrap aggregation.}
\end{figure}

\subsection{Prediction Clusters}

We next use t-distributed stochastic neighbor embedding (T-SNE) to visualize the representations of test data predictions into lower dimensional clusters \cite{maaten2008tsne}.
T-SNE builds these cluster visualizations by converting high dimensional outputs into a lower dimensional embedding while preserving the relationships between neighboring data points.
It does this by first calculating a pairwise similarity between points in the high dimensional space using a Gaussian kernel.
Then the algorithm tries to map the high dimensional outputs to the low dimensional space while preserving the pairwise similarity by minimizing the KL divergence between them.

Figure 7.2 displays the results of this algorithm for the parent and child networks in a Sparse Subnetwork Ensemble. The parent network is a WideResNet-28x10 trained on CIFAR-10 using the training procedure described in the benchmark experiments in Chapters 5 and 6.
These cluster visualizations can reveal interesting insights regarding diversity with regard to the types of classes and predictions that are clustered together.
The mispredictions reveal significant diversity among child networks as these cluster visualizations illustrate their distinct weaknesses.

\begin{figure}[t]
\centering
    \includegraphics[width=\columnwidth]{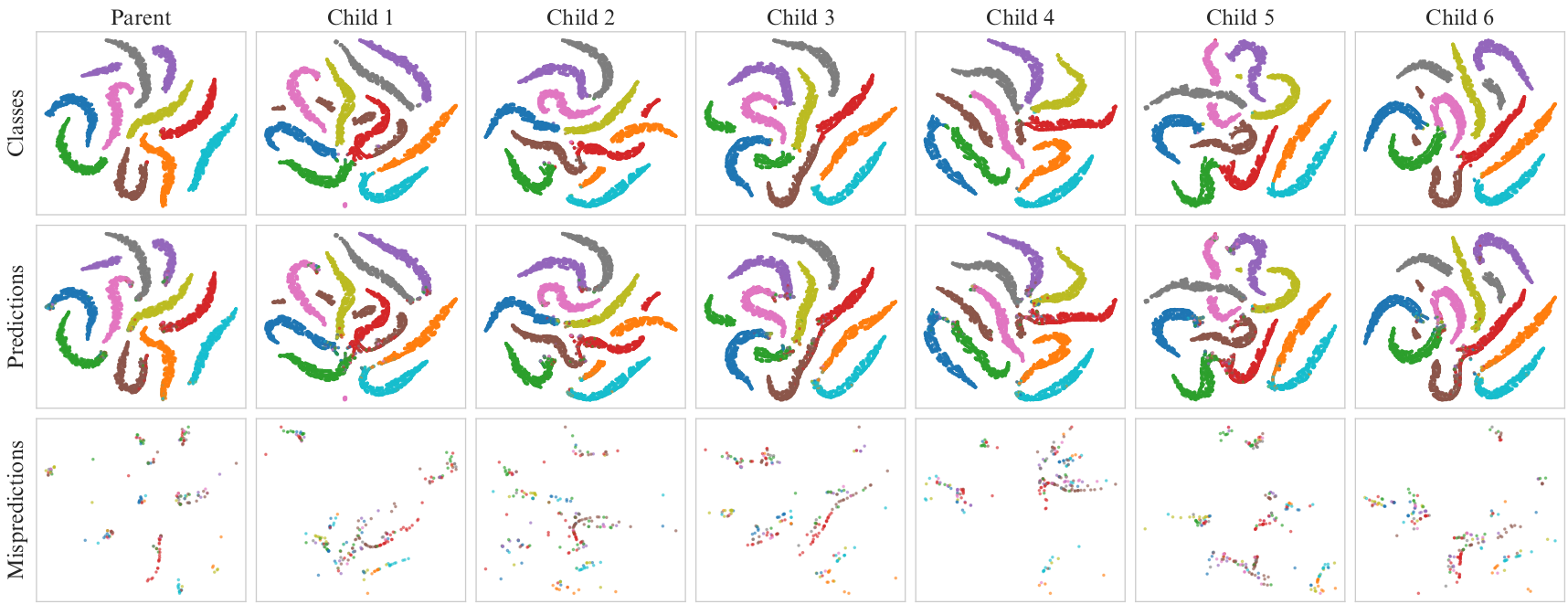}
    \caption{T-SNE cluster representations of CIFAR-10 and WideResNet-28x10. The first row is the representation of classes in CIFAR-10 that each of our models produce. The second row displays the results of our model predictions and the final row displays the incorrect predictions.}
\end{figure}

\section{Interpretable Diversity}

Evaluating diversity through outputs has notable limitations as it can paint an incomplete picture of the underlying diversity within an ensemble. 
These metrics may be sensitive to data distributions and depend heavily on individual member accuracy.
Figure 7.1 provides an example how a group of less accurate models with high diversity can perform worse than a group of more accurate models with lower diversity.
Because of this, output diversity metrics can be misleading when comparing the results of different ensemble methods.
These diversity metrics alone are not sufficient to fully analyze the generalization capabilities of different ensemble methods.

We explore how we can extend diversity analysis with neural network interpretability methods. 
These methods can be used to quantify the diversity between models in the learned feature spaces, rather than solely on output predictions.
By visualizing the types of patterns that ensemble members respond to, we can gain a more comprehensive look at how diversity can play a role in decision making processes.


Feature visualization uses optimization to create images that maximize or minimize activations of specific parts of the network. This is done by first initializing an image with random noise. 
This image serves as the starting point for an optimization problem wherein the objective is to maximize or minimize the activation of target neurons. This image is then fed through a pre-trained network, where stochastic gradient descent is used to iteratively update the pixel values of the image to optimize the activation of the designated neuron.
The resulting images offer critical insights into the types of patterns and features that particular neurons respond to.

Saliency Maps visualize the gradient of a prediction with respect to the input image, highlighting the parts of the input that a network responds strongly to. This is done by first performing a forward pass through the network with a given input image. The saliency is computed by setting all outputs of the non-predicted class to 0 and back propagating the predicted class score back through to the inputs \cite{avanti2017saliency}. This is the gradient $G(X)$ of the output of the predicted class label $Y_c$ with respect to the input image $X$.
%
SmoothGrad was later introduced as an extension that reduces noise in the visualizations by averaging the gradient over several Gaussian perturbed inputs $\gamma \sim \mathcal{N}(0, \sigma^2)$ \cite{smilkov2017smoothgrad}.
\begin{align}
G(X) &= \frac{\partial Y_c}{\partial X} \\
SG(X) &= \frac{1}{N} \sum_{i=1}^N \frac{\partial Y_c}{\partial X + \gamma_i}
\end{align}
%

\subsection{Perceptual Hashing}

Image similarity metrics are generally used to measure image quality as a response to noise, corruption, or compression against a ground truth image \cite{sinha2011perceptually}. 
While these metrics are simple to compute and provide a straightforward quantitative measure, they tend to be highly sensitive and may not correlate well with human perception of image quality.
Sensitivity to pixel level details and small spatial transformations can be problematic when applied to images of feature visualizations that are generated via stochastic processes.

We instead introduce a robust image hash similarity based approach to quantifying diversity. Image hashing algorithms compress images into binary bit strings such that images that are visually similar will result in hashes that are similar. The bits of the image hash are spatially significant, so the similarity between two hashes can be described by the Hamming Distance, which is the number of positions in which the bits of two hashes differ. Images that are more similar will therefore have a lower Hamming Distance than images that are diverse \cite{zauner2010implementation}.


We use several popular image hashing algorithms for measuring the distances between feature visualizations, including: average hash, perceptual hash, difference hash, wavelet hash, and color hash \cite{buchner2021imagehash}. Average hash, difference hash, wavelet hash and perceptual hash all start by resizing a given image to an 8x8 pixel square and converting it to grayscale. The hash is then constructed by assigning a 1 or 0 for each pixel in the block according to some heuristic. Average Hash assigns a 1 if the pixel value is greater than the mean pixel value.
Perceptual Hash does the same except all values are converted to the frequency domain using a discrete cosine transformation first. Wavelet hash uses a discrete wavelet transform instead of a discrete cosine transform. Difference hash is a gradient hash where the difference between pixels values is compared to the mean difference between pixel values. Finally, color hash maintains color information by skipping the grayscale step and instead using the hue, saturation, and value space of an image separately.

\section{Experiments}


We conduct a large scale comparison between child networks in a Snapshot Ensemble and a Sparse Subnetwork Ensemble. We start with an open source Inception network pre-trained on ImageNet \cite{pytorch2022torchvision}. Following the hyperparameters given in the Snapshot Ensemble paper, we continue training of the pre-trained network for two cycles of 40 epochs in length with a cyclic cosine-annealing decay schedule with an initial learning rate of 0.1 and a final learning rate of 1e-5 \cite{huang2017snapshot}.

We use the same pre-trained network as the parent for the Sparse Subnetwork Ensemble as well. Two child networks are created via neural partitioning with an unstructured sparsity target of 50\%. Each child is fine tuned for 40 epochs using a one-cycle learning rate schedule with a max learning rate of 0.1 and a final learning rate of 1e-5. We choose these hyperparameters to match the computational budget defined in Snapshot Ensembles \cite{huang2017snapshot}.

\begin{figure}
\centering
\includegraphics[width=\textwidth]{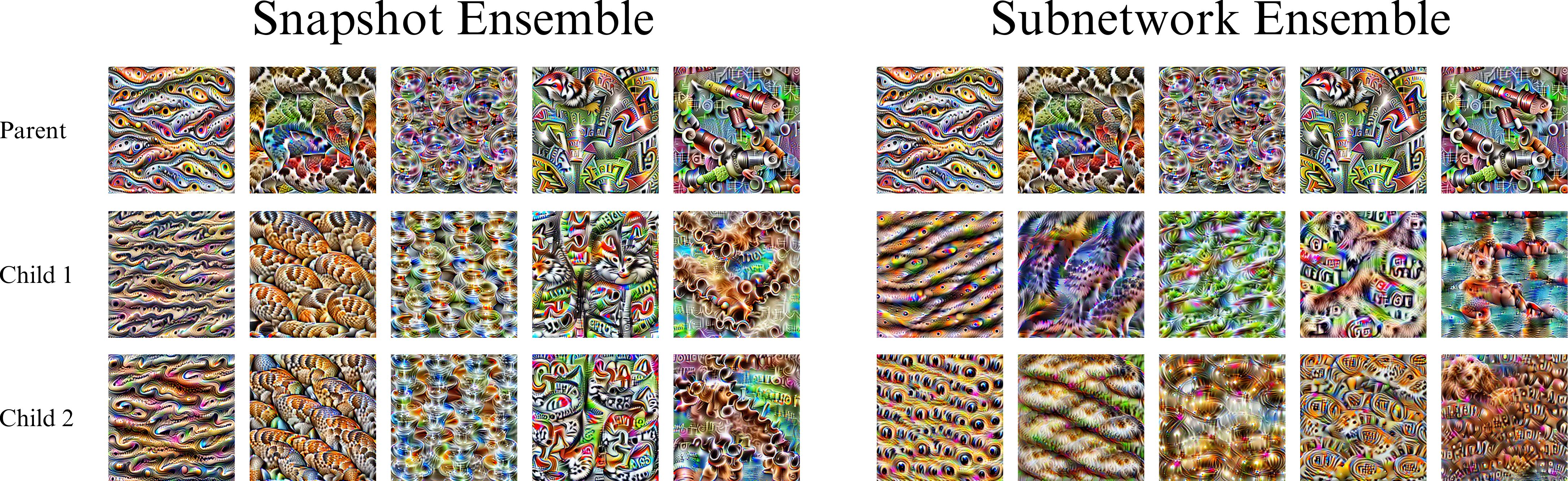}

\vspace{0.1in}

\scriptsize
\begin{tabularx}{\textwidth}{X c c c c c}
\toprule
Method & Average Hash & Perceptual Hash & Difference Hash & Wavelet Hash & Color Hash \\
\midrule
Snapshot Ensemble & $30.85 \pm 0.27$ & $30.77 \pm 0.25$ & $31.55 \pm 0.28$ & $30.94 \pm 0.28$ & $27.56 \pm 0.39$ \\
Sparse Subnetwork Ensemble & $31.16 \pm 0.28$ & $31.19 \pm 0.25$ & $31.60 \pm 0.27$ & $31.17 \pm 0.27$ & $32.74 \pm 0.51$ \\
\bottomrule
\end{tabularx}

\caption{Feature visualizations of a random selection of neurons in child networks from a Snapshot Ensemble and a Sparse Subnetwork Ensemble. Values reported are the mean hash distances between the two child networks over all 1024 visualizations in the final convolutional layer. Higher values represent greater distance and more diversity between representations. The sparsity from Subnetwork Ensembles results in more diverse feature representations than two subsequent checkpoints in Snapshot Ensembles.}
\end{figure}

Using the child networks constructed above, we create feature visualizations of identical neurons in each of the networks \cite{kiat2021lucent, tensorflow2021lucid}. Feature visualizations are optimized using ADAM with a learning rate of 0.05 for 1024 steps. Several random data augmentations are applied at each step, including: jittering by up to 8 pixels, scaling by a factor between 0.95 and 1.05, rotating by an angle between -5 and 5 degrees, and jittering a second time by up to 4 pixels.

We then compute 64 bit image hashes for each visualization (average hash, perceptual hash, difference hash, wavelet hash, and color hash) and report the Hamming Distance between each pair of child network visualization hashes.

We first present feature visualizations from several neurons selected at random from different layers within each network. Figure 7.3 displays some sample visualizations where each column corresponds to the same neuron in each of the different networks. Subnetwork children often focus on different elements present in the parent network's visualization, while Snapshot checkpoints appear to be much more highly correlated with the parent.

\begin{figure}[!ht]
    \includegraphics[width=\textwidth]{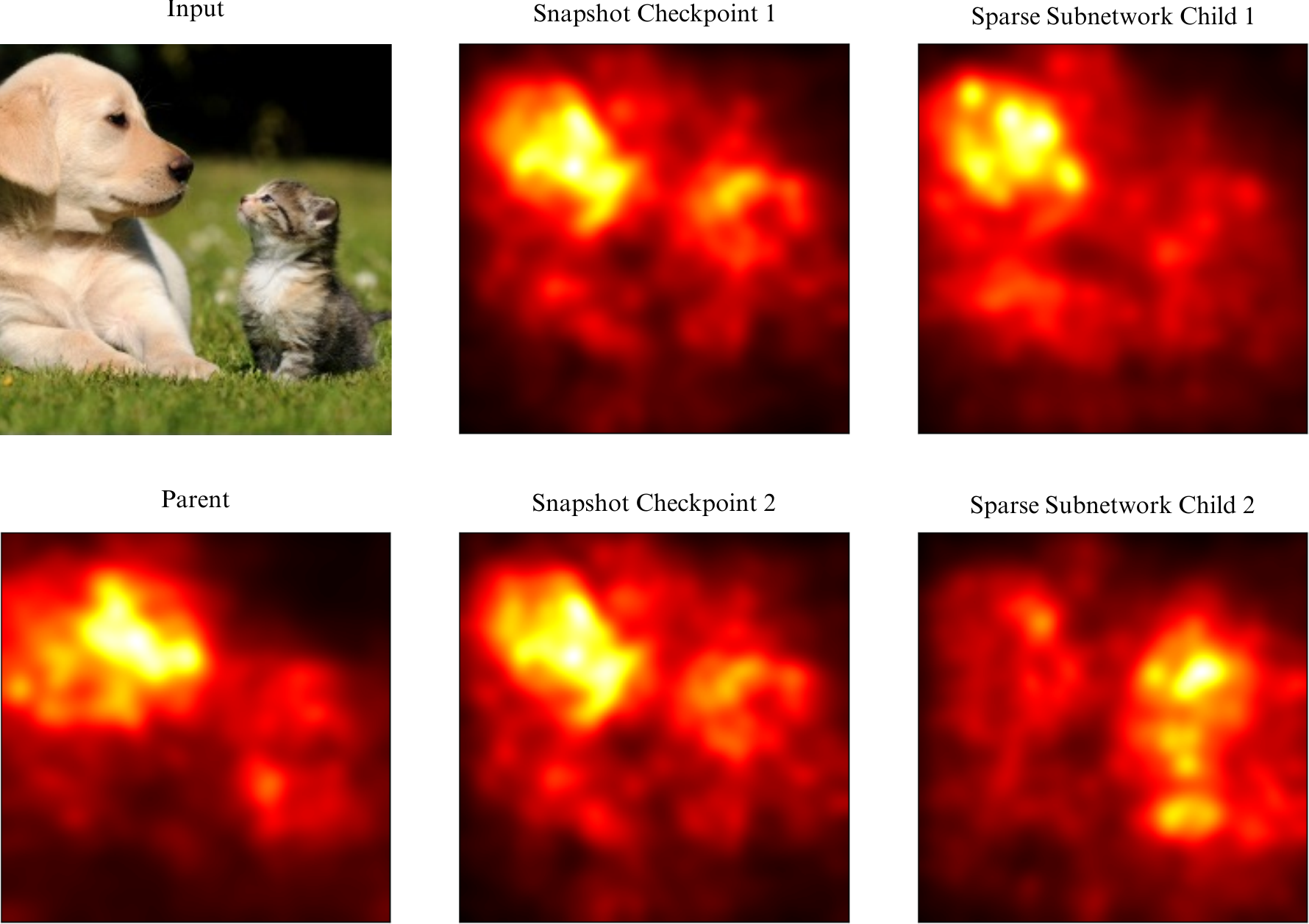}
    
    \vspace{0.2in}
    
    \scriptsize
    \begin{tabularx}{\columnwidth}{X c c c c c }
    \toprule
     & RMSE & Average Hash & Perceptual Hash & Difference Hash & Wavelet Hash \\
    \midrule
    Snapshot Ensemble & $0.06 \pm 0.001$ & $10.01 \pm 0.04$ & $18.12 \pm 0.04$ & $10.00 \pm 0.04$ & $15.13 \pm 0.04$ \\
    Sparse Subnetwork Ensemble & $0.07 \pm 0.001$ & $ 11.13 \pm 0.04$ & $19.95 \pm 0.04$ & $11.03 \pm 0.04$ & $15.97 \pm 0.04$ \\
    \bottomrule
    \end{tabularx}
    \caption{Mean distances between Snapshot Ensemble and Sparse Subnetwork Ensemble saliency maps for all 50,000 samples in the ILSVRC2012 ImageNet validation set. The visualization includes SmoothGrad saliency maps for a shared parent network, two snapshot ensemble checkpoints, and two sparse subnetwork children. The visualization is an example of how saliency maps can be useful with ambiguous input as both snapshot checkpoints focus on the dog while the subnetworks split their attention between the dog and the cat. The diversity encouraged by sparse network structures help to reduce bias present in the parent network.}
\end{figure}

We extend these feature visualizations by analyzing every channel in the final convolutional layer. Thus, for each pair of child networks, 1024 feature visualizations are created for each. All of these visualizations are hashed using the algorithms described above and then the distance between sibling network hashes are measured. These final neurons are especially relevant as they represent the final stage of feature extraction before being fed into the fully-connected classifier layers. They most closely illustrate how each child network represents concepts and the differences over all the neurons in this layer will provide a comprehensive overview on the diversity between network representations. 
The table in Figure 7.3 reports the mean Hamming distances between each pair of child networks over all feature visualization hashes. For every hashing algorithm we see more diversity in the Sparse Subnetwork representations than in the Snapshot Ensemble.

We then explore how each ensemble method interprets input by creating SmoothGrad saliency maps for each of the 50,000 samples in the ImageNet ILSVRC2012 validation set \cite{deng2009imagenet}. Figure 7.4 reports the average Root Mean Square Error, Average Hash, Perceptual Hash, Difference Hash, and Wavelet Hash distance between each pair of child networks over all 50,000 images.
We also include an example visualization of how saliency maps can be useful when dealing with ambiguous inputs. We feed an image containing a puppy and a kitten to each child network in the Snapshot and Subnetwork Ensemble and compute their saliency maps. The parent network and the Snapshot checkpoints focus primarily on the puppy in the input image. The subnetwork children split their attention, where one child focuses on the puppy and the other on the kitten. This diversity can be valuable in building robust ensembles as techniques that encourage diverse representations can help to reduce bias (focus on the puppy) that is inherent in the parent network.

\section{Discussion}

Diversity is an important property of robust neural network ensembles. However, traditional measures of diversity focus only on model outputs and give little insight into how ensemble members represent knowledge differently. We introduce a combination of interpretability methods and perceptual hashing as an effective approach for qualitatively analyzing diversity and measuring the similarity between representations.

Diversity is especially important for low-cost ensemble methods as they tend to share information between members in order to reduce computational cost. In particular, temporal ensembles and evolutionary ensembles create members that inherit parameters and network structure from previous iterations, which allows for meaningful comparisons between the same neurons of different networks.

Our experiments explore how feature visualization and saliency maps could be useful in comparing the diversity between members in Subnetwork Ensembles and we compare that diversity to checkpoints taken in Snapshot Ensembles. We find that the sparsity and unique network topology of sparse subnetwork children is highly effective for encouraging diverse feature representations when compared to Snapshot Ensembles. Neural partitioning ensures that the convolutional filters are geometrically opposed which results in convergence to unique and diverse optima. Despite the significant pruning that these children undergo, Sparse Subnetwork Ensembles maintain high classification accuracy.

Interpretability methods are making deep neural networks more accessible and understandable, and we believe the introduction of these methods to ensemble learning can provide better insights into diversity and aid in the construction of more robust low-cost ensembles.


\chapter{Conclusions}
\label{chap:conclusions}
\section{Summary}

We introduce and formalize Subnetwork Ensembling as a general framework for creating ensembles of deep neural networks, where a large parent network is trained and then used to spawn an ensemble of child networks through a perturbative process applied to randomly sampled subnetworks in the parent.
Several methods were explored for constructing and optimizing child networks through noise perturbations, network pruning, and stochastic masking.
We augment these processes with the introduction of Neural Partitioning, which is an effective variance reduction technique where the parameters of parent networks are divided into sets of children with no parametric overlap.
This work is grounded in the landscape of modern deep ensemble learning research where each method was thoroughly evaluated across a number of benchmark tasks and ablation studies.

We frame Noisy Subnetwork Ensembles as a no-cost ensemble approach where a collection of noise perturbed children can reliably improve generalization on the difficult tasks without requiring any additional training. 
We relate some of the principles behind this work to popular ideas in Neuroevolutionary research.
We build on this with Sparse Subnetwork Ensembles where a fine tuning phase is applied to sparse children to significantly improve the predictive capacity of the ensemble.
The unique sparse network topology encourages child networks to converge to unique optima.
We compare Sparse Subnetwork Ensembles to many recent ensemble learning methods and we observe state-of-the-art performance on fixed training budget benchmarks.
We further improve performance with the introduction of Stochastic Subnetwork Ensembles, where subnetworks are represented with probability matrices that govern parameter activation on forward passes.
We slowly anneal the probability values of the stochastic mask over time towards a deterministic target subnetwork. 
We observe significant improvements over established one-shot and iterative pruning baselines and we further improve performance on benchmarks explored with Sparse Subnetwork Ensembles.

We then conduct a large scale analysis of diversity in Sparse Subnetwork Ensembles.
We first explore the diversity of predictions through common diversity metrics used in ensemble learning literature.
We then introduce a novel approach to analyzing feature space diversity by leveraging network interpretability methods. 
We use feature visualization and saliency maps to visualize knowledge representations with ensemble members. 
We introduce perceptual hashing algorithms as a means to quantify the visual differences between feature representations.
We compare the results of these investigations to another popular low-cost ensemble method called Snaphshot Ensembles, where we found that Sparse Subnetwork Ensembles display significantly more representational diversity among sibling networks.
These results reaffirm the hypothesis that topological structure plays a critical role in encouraging diverse and robust learning.

\section{Discussion}

Subnetwork Ensembling proves to be a highly effective technique for improving the training efficiency, parametric utilization, and generalization of deep ensembles while minimizing computational cost.
We achieve this by uniting several important concepts and insights observed in deep learning research.
Utilizing a large overparameterized parent network facilitates better convergence efficiency and training dynamics.
These trained parent networks often contain large amounts of redundancy that we can leverage to construct accurate child networks that train quickly.
Utilizing different subspaces of the parent network encourages diverse representations among child networks as topological transformations affect the underlying optimization landscape.
The introduction of sparsity, through subnetwork sampling, serves as a regularization effect that leads to more robust solutions.
This leads to more diversity among child networks as distinct network topologies  can transform the underlying optimization landscape such that child networks converge to diverse local optima.

This approach also offers several unique advantages over competitive low-cost ensemble methods.
Parent networks can be optimized independently of the child networks which allows for flexibility with network architectures, regularization techniques, and training procedures.
Child networks can be generated and trained quickly, which enables dynamic growth of ensemble sizes and parallel wide scaling.
Child networks can be made to be very sparse with little discernible loss in accuracy which holds great potential for reducing memory and storage costs.

While Subnetwork Ensembles offer numerous advantages for improving training efficiency, generalization, and computational requirements, it is not without its limitations.
Subnetwork Ensembles are fundamentally constrained by the size and capability of the parent network.
This necessitates a parent network that is large enough to ensure that sampled subnetworks can produce accurate solutions.
Furthermore, our research prioritizes the efficient training of these ensembles, which is an increasingly important consideration as the training costs of large scale neural networks are becoming prohibitively expensive and time consuming.
However, it is important to note that there is a large body of work that instead looks to optimize the inference cost of evaluating ensembles.
Our methodology entails separate forward passes for each subnetwork, rendering it less efficient than pseudo-ensemble techniques that incorporate members into a unified model.

One objective of our investigation into Subnetwork Ensembles is to formulate this process under a generalized framework that could provide a foundation for conducting future research.
Emerging network architectures, algorithmic techniques, modalities and problem domains offer new opportunities for implementing these ensembles to improve generalization at minimal cost.
Connections between Subnetwork Ensembles and observed structures in biological neural networks may hold interesting prospects for research that aims to bridge the gaps between machine learning and neuroscience.
Our approach would also be a natural fit for multitask learning, which is an important area of research for artificial general intelligence.
If the trend of increasing model scale continues, the utilization of subnetworks may become increasingly important for training large models. 
Colossal super networks may become so large that traditional optimization methods become infeasible and will instead need to rely on the isolation, distribution, and organization of its subnetworks.

As artificial intelligence continues to become a larger part of the cultural conversation, it's interesting to consider how our research impacts the ways in which we view the world.
Just as Subnetwork Ensembles benefit from the harmonious interaction of its members, so too does the universe operate through interconnected relationships at all scales. From the diverse set of subatomic quanta that form space-time, to the delicate balance of ecosystems that support life, and even to the gravitational relationships between celestial bodies. We consider how communities at all scales thrive when diverse perspectives and properties are integrated.
Subnetwork Ensembles can serve as a compelling metaphorical lens through which to examine the intricate and complex set of relationships that define and sustain the natural world.

\backmatter 



\bibliographystyle{plain}
\bibliography{bibliography} 

\appendix 

\end{document}